\documentclass[final,5p,times,twocolumn,authoryear]{elsarticle}

\usepackage{cite}
\usepackage{amsmath,amssymb,amsfonts}
\usepackage{algorithmic}
\usepackage{graphicx}
\usepackage{textcomp}
\usepackage{xcolor}
\usepackage{multirow}
\usepackage{listings}
\usepackage{url}
\usepackage{verbatim}
\usepackage{float}
\usepackage{caption}
\usepackage{subcaption}
\usepackage{footmisc}
\usepackage{booktabs}
\usepackage{subfig}
\usepackage{xspace}
\usepackage{natbib}

\newcommand{\dName}{CONDIS++\xspace}
\newcommand{\dFullName}{Cotton Growth-Stage Discerning Image Set}   
\newcommand{\modelName}{COTONET\xspace}
\newcommand{\nParameters}{7.6M\xspace}
\newcommand{\FLOPS}{27.8 GFLOPS\xspace}
\newcommand{\mAPeasy}{81.1\%\xspace}
\newcommand{\mAPdifficult}{60.6\%\xspace}

\newcommand{\Height}{3024\xspace} 
\newcommand{\Width}{4032\xspace} 

\journal{Computers and Electronics in Agriculture}
\begin{document}
\begin{frontmatter}

\title{\modelName: A custom cotton detection algorithm based on YOLO11 for stage of growth cotton boll detection}

\author[iri]{Guillem González\corref{cor1}}
\ead{ggonzalez@iri.upc.edu}
\author[iri]{Guillem Alenyà}
\author[iri]{Sergi Foix}

\cortext[cor1]{Corresponding author}

\affiliation[iri]{organization={Institut de Robòtica i Informàtica Industrial (CSIC-UPC)},
                  addressline={Llorens i Artigas 4-6},
                  city={Barcelona},
                  postcode={08028},
                  country={Spain}}

\begin{abstract}
    Cotton harvesting is a critical phase where cotton capsules are physically manipulated and can lead to fibre degradation. To maintain the highest quality, harvesting methods must emulate delicate manual grasping, to preserve cotton's intrinsic properties. Automating this process requires systems capable of recognising cotton capsules across various phenological stages.
    To address this challenge, we propose \modelName, an enhanced custom YOLO11 model tailored with attention mechanisms to improve the detection of difficult instances. The architecture incorporates gradients in non-learnable operations to enhance shape and feature extraction. Key architectural modifications include: the replacement of convolutional blocks with Squeeze-and-Exitation blocks, a redesigned backbone integrating attention mechanisms, and the substitution of standard upsampling operations for Content Aware Reassembly of Features (CARAFE). Additionally, we integrate Simple Attention Modules (SimAM) for primary feature aggregation and Parallel Hybrid Attention Mechanisms (PHAM) for channel-wise, spatial-wise and coordinate-wise attention in the downward neck path. This configuration offers increased flexibility and robustness for interpreting the complexity of cotton crop growth.
    COTONET aligns with small-to-medium YOLO models utilizing \nParameters parameters and \FLOPS, making it suitable for low-resource edge computing and mobile robotics. \modelName outperforms the standard YOLO baselines, achieving a mAP50 of \mAPeasy and a mAP50-95 of \mAPdifficult. 
\end{abstract}


\end{frontmatter}

\section{Introduction} \label{INTRO}

Cotton is a resilient natural fibre that is breathable, absorbent, easy to dye, and biodegradable. Its properties make cotton the most widely used natural fibre in manufacturing~\citep{wang2020cottonproperties}. Its applications span diverse sectors, including fashion, medicine, cosmetics, and the food industry, among others. Traditionally, cotton cultivation involves large-scale harvesting where industrial machinery removes the entire plant once the bolls reach maturity. The white cellulose fibres are then separated via ginning~\citep{negm2020cottonpicking}. While time-efficient, this system necessitates annual replanting, increasing operational costs, and can diminish fibre quality due to the intensive mechanical stresses of the ginning process. Although manual harvesting preserves higher fibre integrity, high labour costs often render it economically unviable~\citep{negm2020cottonpicking}.

To address these challenges, mobile robotic systems guided by computer vision are being developed to detect and collect cotton bolls efficiently. These systems aim to emulate manual grasping through innovative end-effectors without compromising fibre quality. Recent research suggests that perennial cotton cultivation (2-to-4 cycles) can perform as well as or better than annual crops~\citep{zhang2022perennial}. Maintaining plants across multiple cycles offers a more sustainable and cost-effective alternative to traditional annual replanting, significantly reducing the carbon footprint and energy consumption of the orchard.

Precise detection is the critical first step in automated harvesting. From a computer vision perspective, the priority is identifying the transition from immature to fully mature states, as fibres only complete their development at full maturity. Once mature, the boll remains stable without wilting for an extended period~\citep{cathey1979cottondissecation}. This stability is a key advantage for robotic harvesting: if ripeness is uncertain, the boll can safely be left on the plant for a later cycle without degradation.

Despite its importance, research on the detection, segmentation, or localization of cotton fruits remains limited, as most of the literature focuses on disease classification~\citep{omaye2024cross, dewangan2023systematic}. Furthermore, there is a scarcity of publicly available datasets for the detection of cotton bolls~\citep{gonzalez2024}. This gap provides an opportunity to investigate innovative methods for detecting specific growth stages. 

Early studies, such as Verma et al.~\citep{verma2024}, fine-tuned a YOLOv8 model to discern growth stages but achieved a modest $mAP50$ of 0.643. Li et al.~\citep{Li2016cottonsegmentation} proposed a region semantic segmentation method combining SLIC, DBSCAN with Wasserstein distance and random forests, achieving semantic labelling prediction in-field cotton images with higher mean values and reduced standard deviation, compared to state-of-the-art algorithms. In 2017, Li et al.~\citep{Li2017cottonsegmentation} would then propose DeepCotton, a deep learning solution to in-field cotton segmentation by combining a fully convolutional network and a refinement algorithm to remove interference. DeepCotton achieves up to 2.6\% accuracy and 8.1\% IoU gains over state-of-the-art methods in ordinary and multi-plant scenarios.

In our preliminary work~\citep{gonzalez2025improved}, we analysed early versions of the proposed cotton crop dataset. Although we achieved strong metrics on simple detections (0.823), feature extraction for general complex detections remained lower (0.548). That dataset lacked representation of critical early growth phases, specifically initial bract formation and early bloom, which are essential for comprehensive lifecycle monitoring.

Besides this preliminary work, other works present limitation for our case study. Verma et al.~\citep{verma2024} achieve low accuracy with lacking classes for complete growing cycle representation. Li et al.~\citep{Li2016cottonsegmentation} implements instance segmentation, which can be easily transformed to object detection, but it would be too slow and computationally expensive. The same reasoning is applied to Li et al.~\citep{Li2017cottonsegmentation}. 

Recently, Gong et al.~\citep{GONG2025101209} presented a relevant study utilising a YOLOv10-nano model to detect four classes: flowers, partially open bolls, fully open bolls, and defective bolls. They achieved a 1.3 percentage point increase in $mAP50$ over the baseline while reducing model complexity by a 5.8\% in parameter size and 15.2\% in floating point operations. Their dataset gathers 2000 images, where approximately 22350 labels were annotated (from visual estimation in Gong et al. Fig.2~\citep{GONG2025101209}). Given its relevance, the work by Gong et al. serves as the primary benchmark for our proposed solution.

Beyond cotton-specific research, numerous studies in agronomic computer vision have adapted YOLO architectures by incorporating state-of-the-art modules to enhance performance on specialized datasets. 

For instance, Li et al.~\citep{Li2025pineappleYOLO} modified YOLOv7-tiny for pineapple detection in open fields by introducing the Convolutional Block Attention Module (CBAM) for improved feature extraction and the Content-Aware Reassembly of Features (CARAFE) module to expand the receptive field during upsampling. By redefining penalty criteria with the Scylla Intersection over Union (SIoU) loss function, they achieved a 5.8\% boost in $mAP_{50}$ without significant computational overhead. Similarly, in apple orchard applications, Liu et al.~\citep{liu2024fasteryoloapapples} optimized the YOLOv8 architecture by replacing standard convolutions with Partial Depth-wise (PDWConv) and Depthwise Separable (DWSConv) convolutions. By utilizing Efficient Intersection over Union (EIoU), they maintained performance comparable to the YOLOv8n baseline while drastically reducing parameters from 3 million to 0.66 million.

Advancements have also been reported in viticulture; Wang et al.~\citep{wang2024grapedetection} integrated dual-channel feature extraction attention (DCFE) and dynamic snake convolutions (DS-Conv) into YOLOv5s. These modifications yielded increases of 2.02\% in $mAP_{50-95}$ and 2.5\% in $mAP_{50}$, providing necessary robustness for the unstructured environments of vineyards. In tomato cultivation, Deng et al.~\citep{deng2025tomato} developed SE-YOLO from a YOLO11 baseline, incorporating ADown for downsampling and a Sobel Edge Detection and Feature Fusion (SEDFF) module to integrate edge information. Furthermore, their SPStem module facilitated the detection of small or occluded fruits, resulting in an overall 3.2\% improvement in $mAP_{50}$. Zhai et al.~\citep{zhai2024greentomato} also addressed environmental complexities with TEAVit, a model tailored for green tomato detection that leverages texture feature extraction (TEAM), contextual semantics (CAM) and addresses background complexities and occlusions (EFM) to exceed 90\% across all primary metrics over their custom tomato dataset.

Building upon these advancements and identifying the limitations in the current state of the art, notably the work of Gong et al., this study pursues the following objectives:

\begin{enumerate}
    \item [(1)] The development of a publicly available cotton fruit dataset featuring six classes that represent distinct phenological stages, captured under diverse perspectives and lighting conditions.
    \item [(2)] The proposal of a custom YOLO architecture, \modelName: a lightweight, attention-centric model that integrates state-of-the-art convolutional modules to outperform standard YOLO variants in complex agricultural scenes.
\end{enumerate}

\section{Dataset Conceptualization} \label{DS}

The task we want to solve is cotton detection in different growth stages, in a complex greenhouse environment, versatile to changes in weather conditions, lighting, visual noise and occlusion, for later crop counting and monitoring. To achieve this, we need to test a series of generic, state-of-the-art object detection models and evaluate them to determine which one performs better. 

However, there are very few works on cotton boll detection, and thus, there are few public datasets of bolls in different growth stages. Most studies of the cotton plant focus on diseases that the plant can suffer from, and because of that, we decided to create our dataset, taking advantage of the images from our previous study~\citep{gonzalez2024}, adding more images, changing the classes, adjusting the labels, and thus creating a more complete representation of what a cotton boll is.

Before building a dataset to study cotton growth, we first need to know the different growth phases and their morphology. In our previous work, we analysed the different phases of cotton growth to distinguish between mature and immature phases in a binary way~\citep{gonzalez2024}. In addition, we need to know the boundaries at which one growth stage changes to the next one for precise labelling. We will be working with the same variety of cotton, called Intercott-211.

Cotton morphology presents itself in very different shapes: the bolls ready to be harvested may have the fibres more or less spread over their surface, elongated or rounded, separated into three, four or five sections, with fibres stretched by being pulled by other parts of the plant. Immature open bolls have varying degrees of openness, and can be mistaken for closed bolls, if they have just opened, or for mature bolls, if their bracts are not yet fully brown. Closed bolls may show a more developed capsule (visible fibres) or a less developed capsule (fibres hidden by bracts).

\begin{figure}[t]
    \includegraphics[width=\columnwidth]{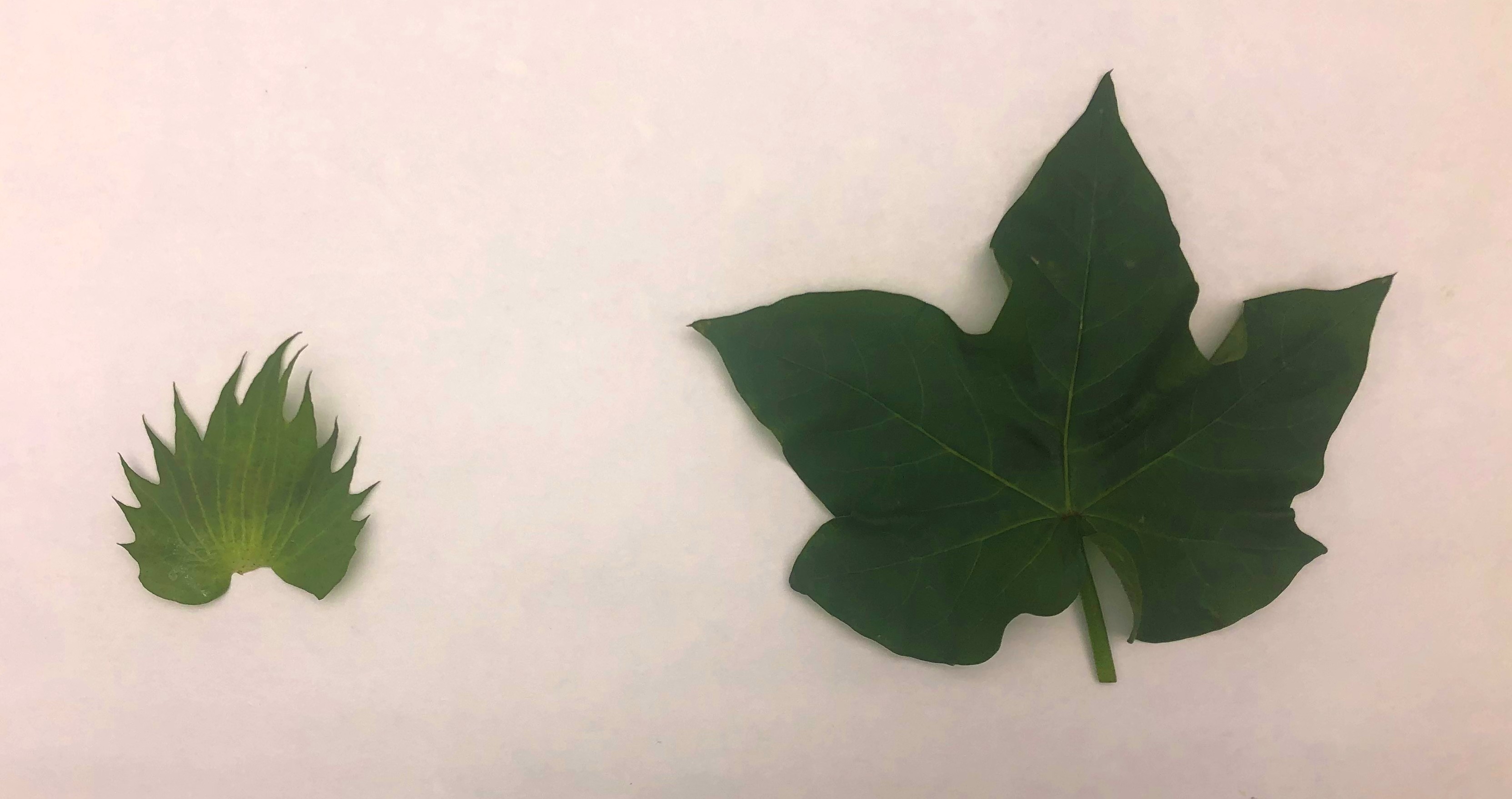}
    \caption{Two types of leaves found in cotton plants. On the left, a sample of a bract. On the right, a sample of a true leaf.}
    \label{fig:True-leaf-VS-bract}
\end{figure}

In addition, when cotton buds start to appear, there also appears a leaf-like organ subtending the future boll, called bracts. Unlike regular leaves, in charge of photosynthesising sunlight, bracts support flower development and later boll growing. When the bolls open up, the bracts and capsule containing the fibres start to dry, and the bracts become rigid. In Fig.~\ref{fig:True-leaf-VS-bract} we can see a side-to-side example. 

As we have seen in the state of the art, Verma et al.~\citep{verma2024} recognise cotton buds, blossoms, squares, unripe open bolls and ripe bolls, collecting a combination of Kaggle datasets and photographs taken with a smartphone. Gong et al.~\citep{GONG2025101209} establish blossoms, partially open bolls, fully open bolls and defective bolls as classes to detect. In our case, we consider it crucial to distinguish between unfertilised flowers and fertilised flowers, since they mark the moment of fertilisation, changing colour from off-white or light yellow to a pinkish, violet colour. This distinction allows maintaining a more detailed analysis of cotton bolls and their growth states. Consequently, we contributed with the conceptualisation of the following classes:

\begin{enumerate}
    \item [0 -] ``Ripe Boll": Cotton ready for harvest, fully developed.
    \item [1 -] ``Open Boll": Immature open cotton.
    \item [2 -] ``Square": Unopened cotton boll, between flower falling and boll full size acquisition.
    \item [3 -] ``Fertilised Flower": Fertilised cotton flower, pink to violet.
    \item [4 -] ``Early Flower": Unfertilised cotton flower, off-white to light yellow colour.
    \item [5 -] ``Bracts": The first little leaves where the flower starts to grow.
\end{enumerate}

Distinguishing between open cotton and mature cotton is crucial; however, it is also key to know when the other classes transition to the next state. For this purpose, we establish the following criteria, with expert knowledge on cotton phenology provided by the personnel at CEBAS-CSIC (Centro de Edafología y Biología Aplicada del Segura):

\begin{enumerate}
    \item ``Bud" to ``Early Flower": The flower cocoon opens, and white to yellowish petals appear and open rapidly.
    \item ``Early Flower" to ``Fertilised Flower": The flower begins to acquire pinkish-violet tones, which means that the flower has been pollinated.
    \item ``Fertilised Flower" to ``Square": The flower falls, and only the leaves are left to envelop the forming cotton boll. A tiny green capsule appears.
    \item ``Square" to ``Open Boll": The boll reaches its maximum size and starts to open up. White fibres start to come out of it, becoming visible.
    \item ``Open Boll" to ``Ripe Boll": The husk that encloses the boll becomes fully open, and the bracts surrounding it dry out, changing from deep green to a brown, ashy colour.
\end{enumerate} 

Once we have built distinguishable classes, we are able to collect a set of images were different examples of cotton bolls, cotton flowers and cotton buds are represented. The labelling process is carried out using the LabelImg~\citep{labelImg} software. From now on, we will refer to the cotton dataset as \dName (\dFullName).

\subsection{Dataset properties}

Images are taken under the same conditions as in~\citep{gonzalez2024}. Pictures were taken with a 15 MP (\Height x \Width) smartphone camera, 50 cm to 1 m away from the plant. The dataset consists of high-definition photographs, as well as general views showing the diversity of growth phases of a plant at the same instant in time. Imaging was done from June to July, between 10 am and 12 pm. Cotton plants are located inside a 500 m$^2$ polycarbonate greenhouse. The temperature ranged from 20°C at night to 32°C during the day, with a relative humidity of 50\% during the day and 85\% at night. There was no shading net at the time of imaging.

\begin{figure*}[t]
\centering

\newcommand{\imgw}{0.19\textwidth}
\newcommand{\interspaceb}{\hspace{2mm}}

\begin{subfigure}{\imgw}
    \includegraphics[width=\linewidth, angle=-90]{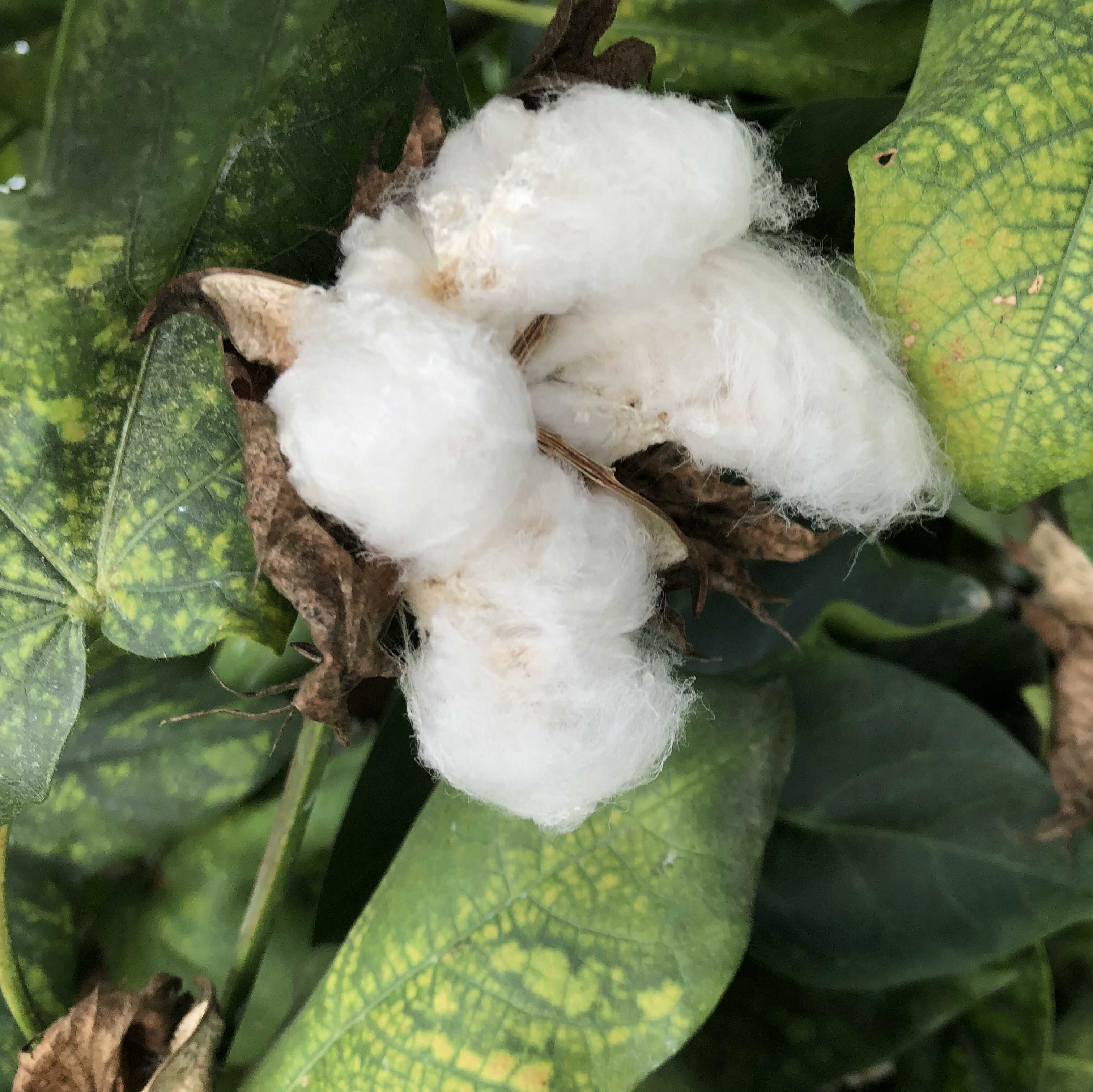}
    \caption{}
\end{subfigure}\interspaceb%\hfill or \hspace{5mm} or \hspace{0.3\textwidth}%
\begin{subfigure}{\imgw}
    \includegraphics[width=\linewidth]{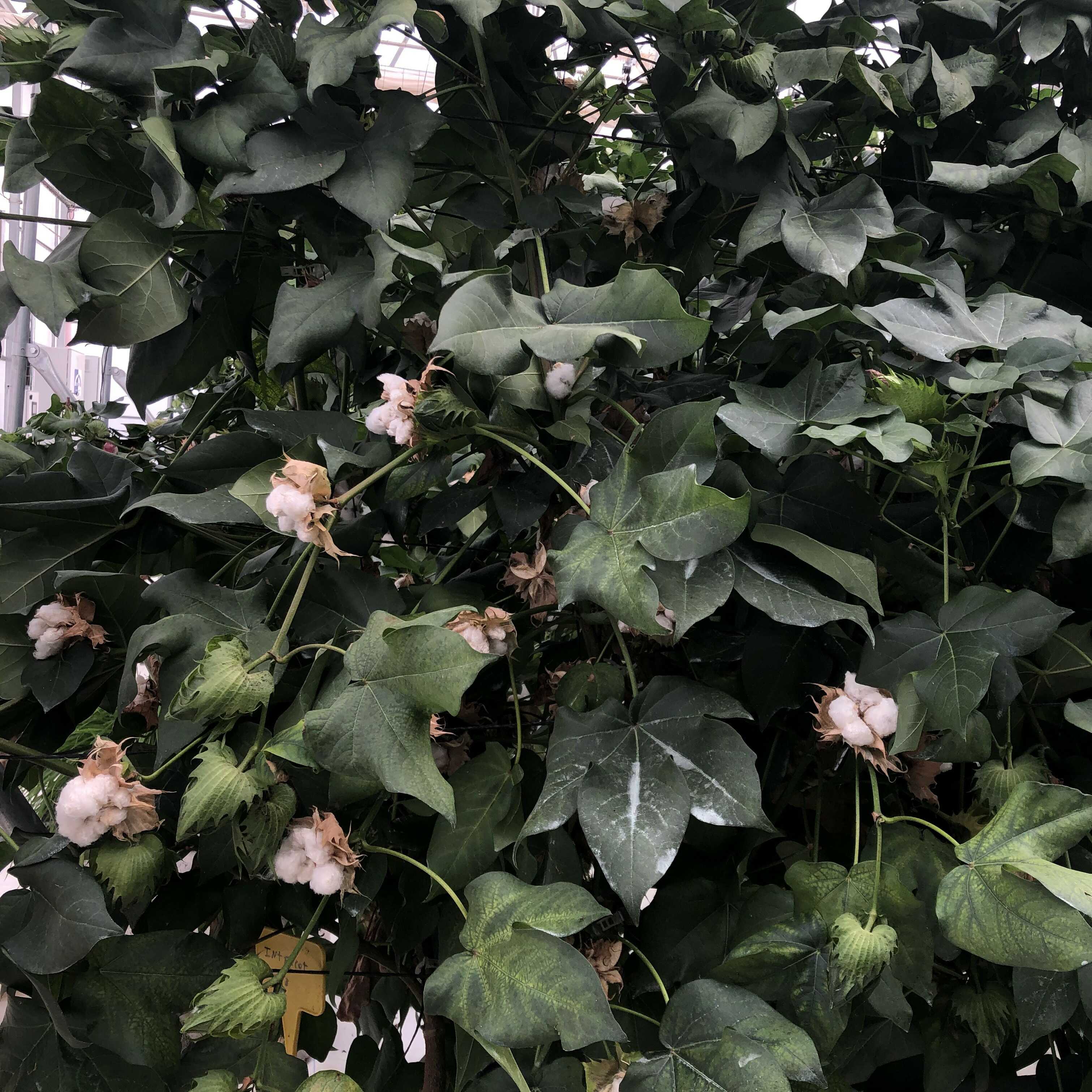}
    \caption{}
\end{subfigure}\interspaceb%
\begin{subfigure}{\imgw}
    \includegraphics[width=\linewidth, angle=-90]{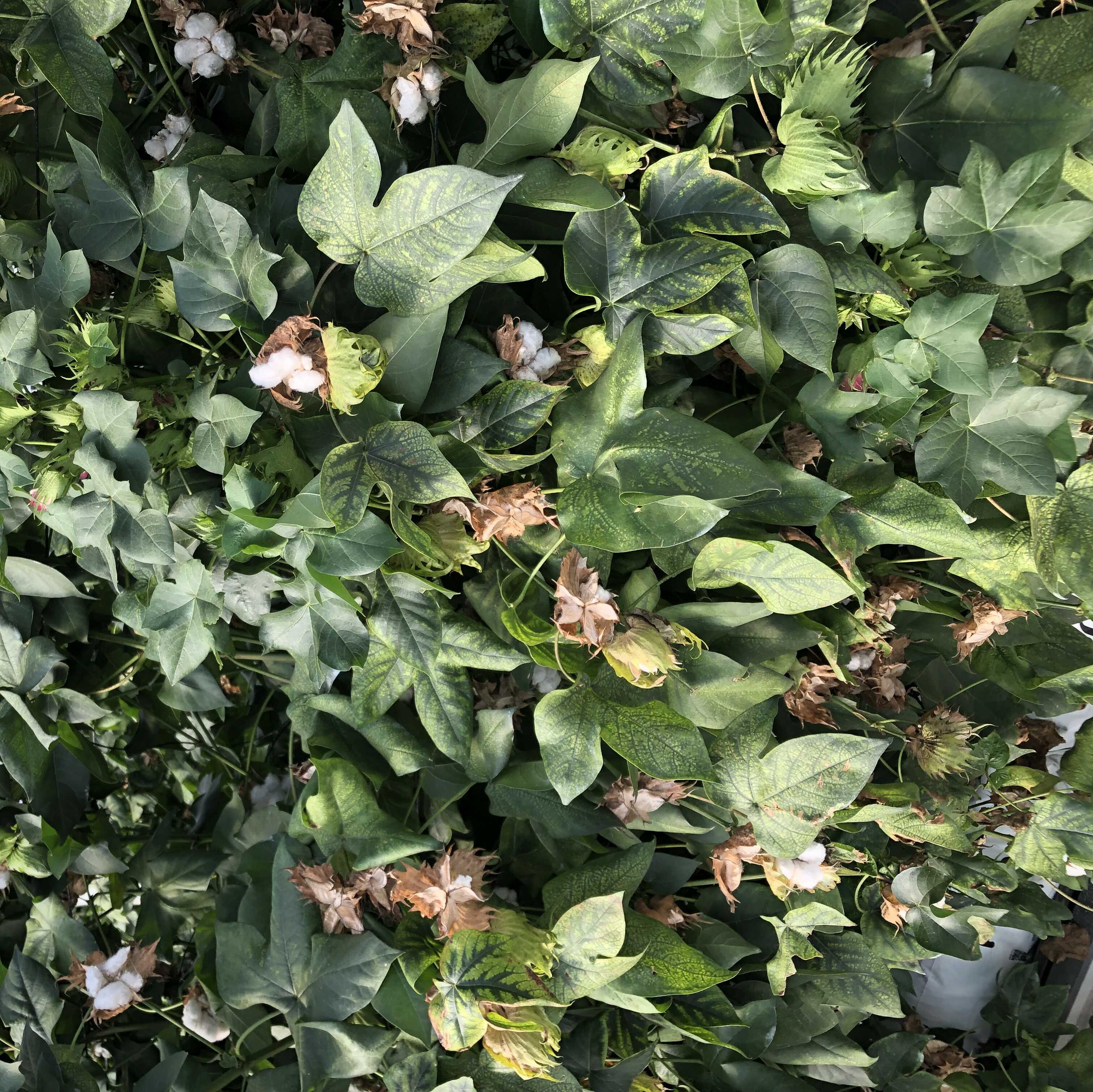}
    \caption{}
\end{subfigure}\interspaceb%
\begin{subfigure}{\imgw}
    \includegraphics[width=\linewidth, angle=-90]{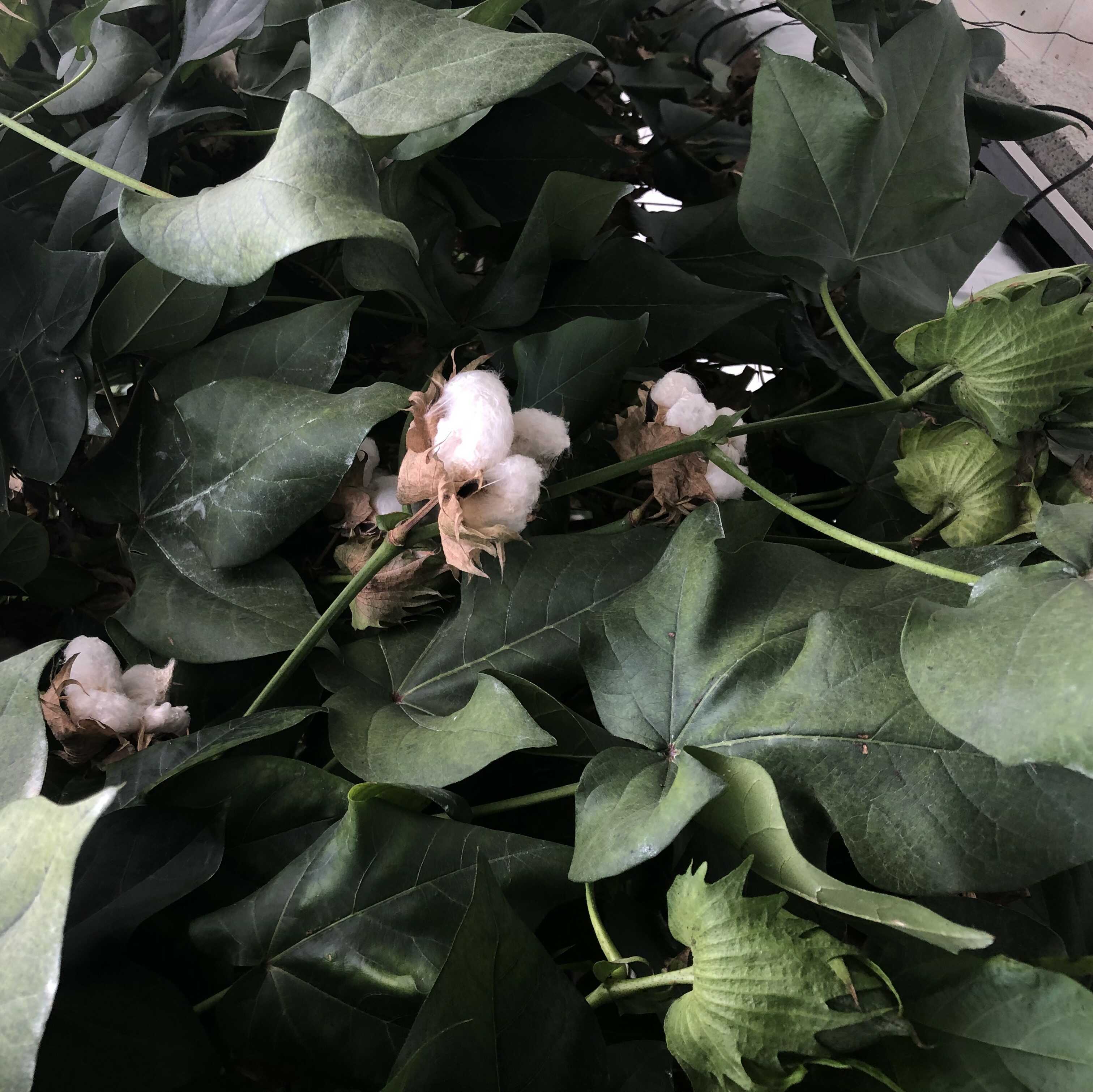}
    \caption{}
\end{subfigure}\interspaceb%
\begin{subfigure}{\imgw}
    \includegraphics[width=\linewidth, angle=-90]{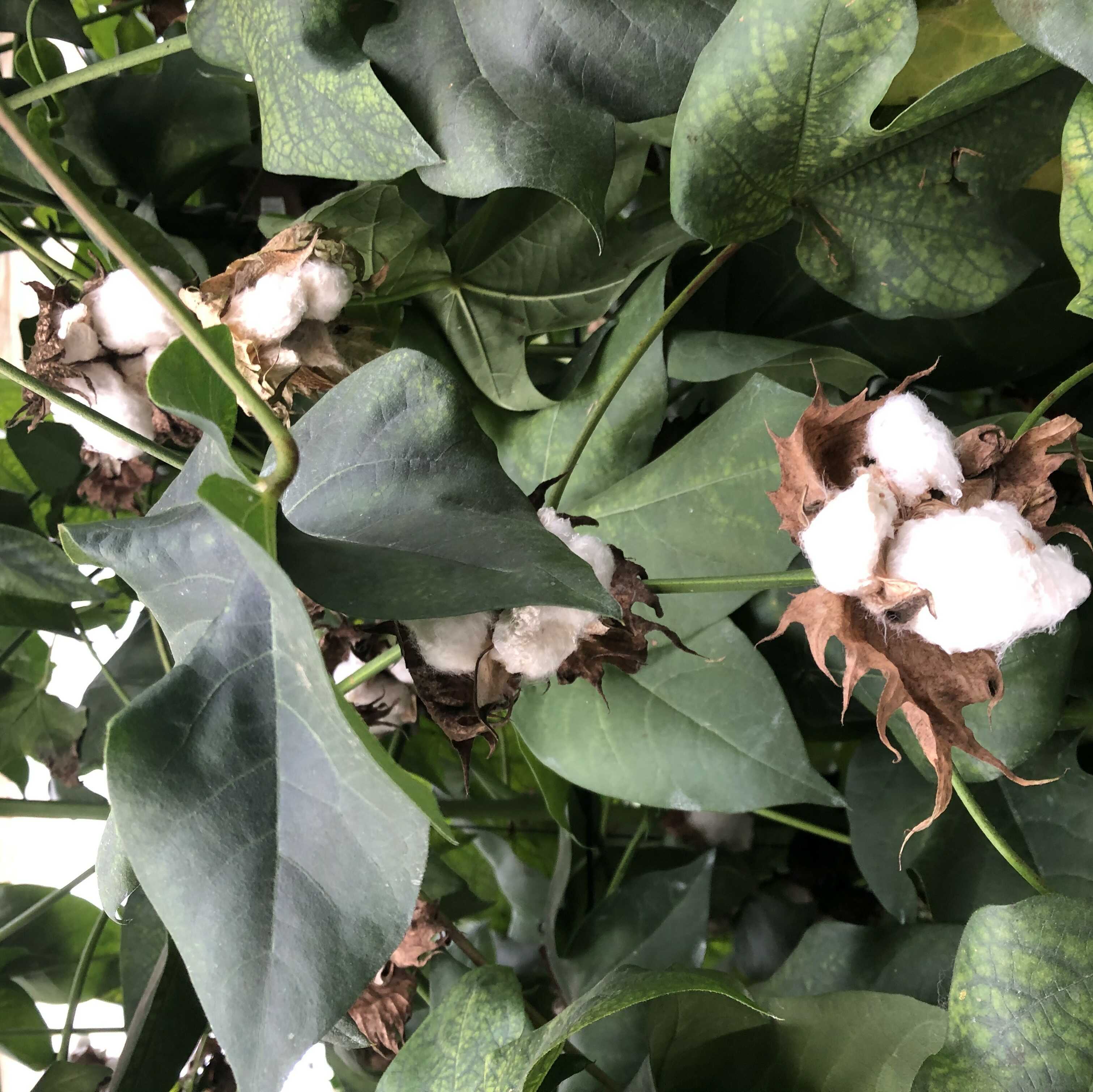}
    \caption{}
\end{subfigure}

\vspace{5pt}

\begin{subfigure}{\imgw}
    \includegraphics[width=\linewidth, angle=-90]{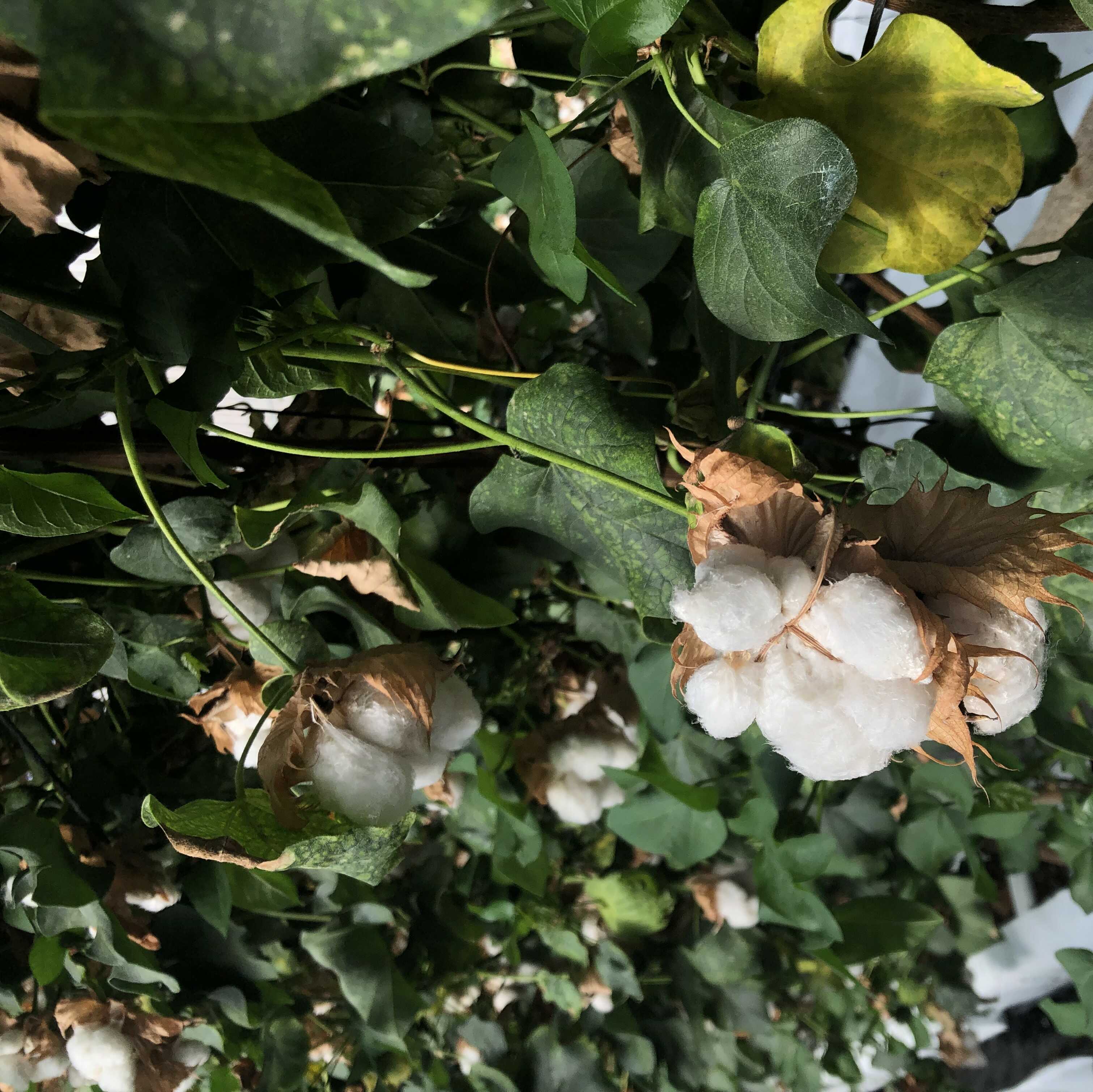}
    \caption{}
\end{subfigure}\interspaceb%
\begin{subfigure}{\imgw}
    \includegraphics[width=\linewidth, angle=-90]{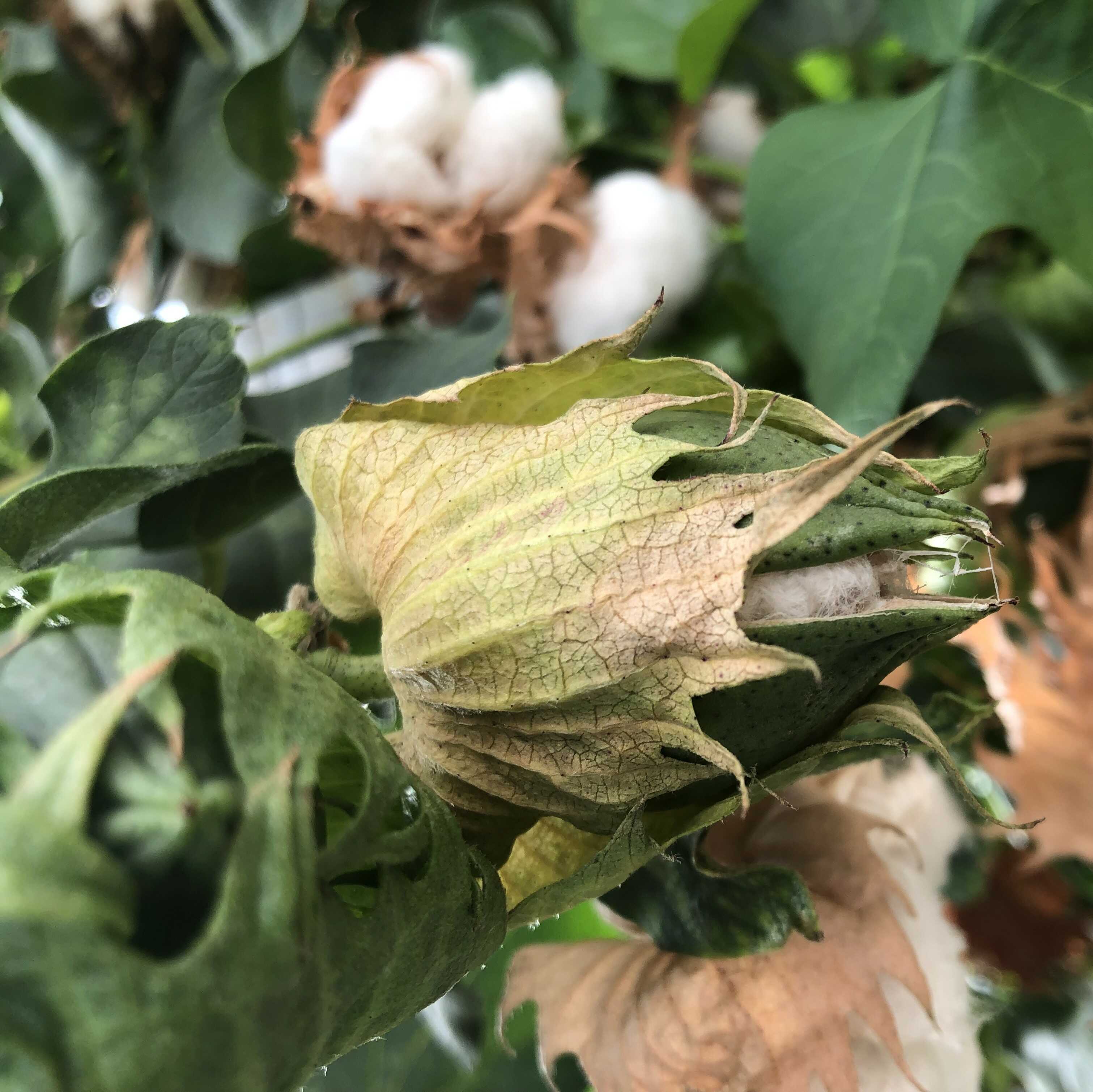}
    \caption{}
\end{subfigure}\interspaceb%
\begin{subfigure}{\imgw}
    \includegraphics[width=\linewidth]{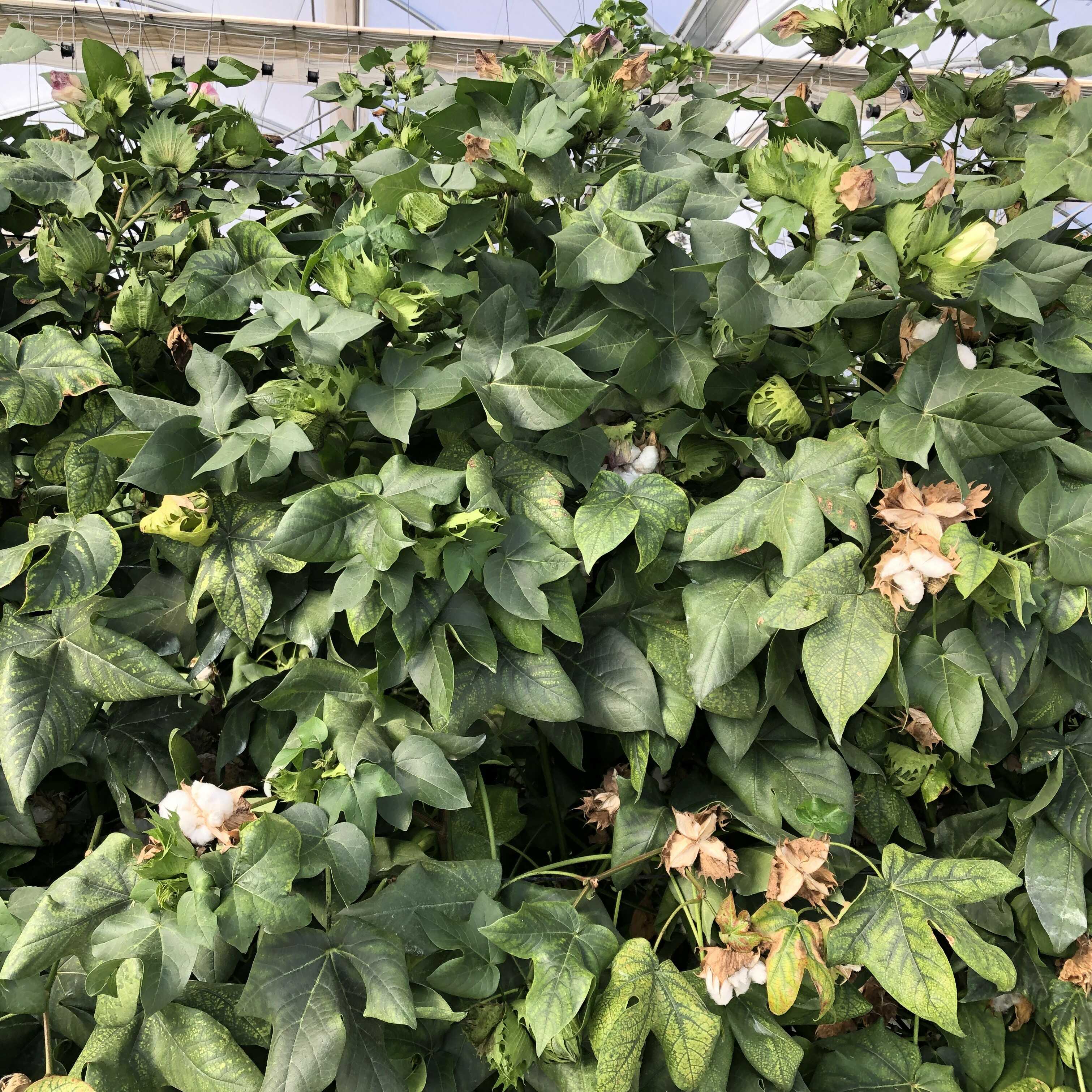}
    \caption{}
\end{subfigure}\interspaceb%
\begin{subfigure}{\imgw}
    \includegraphics[width=\linewidth]{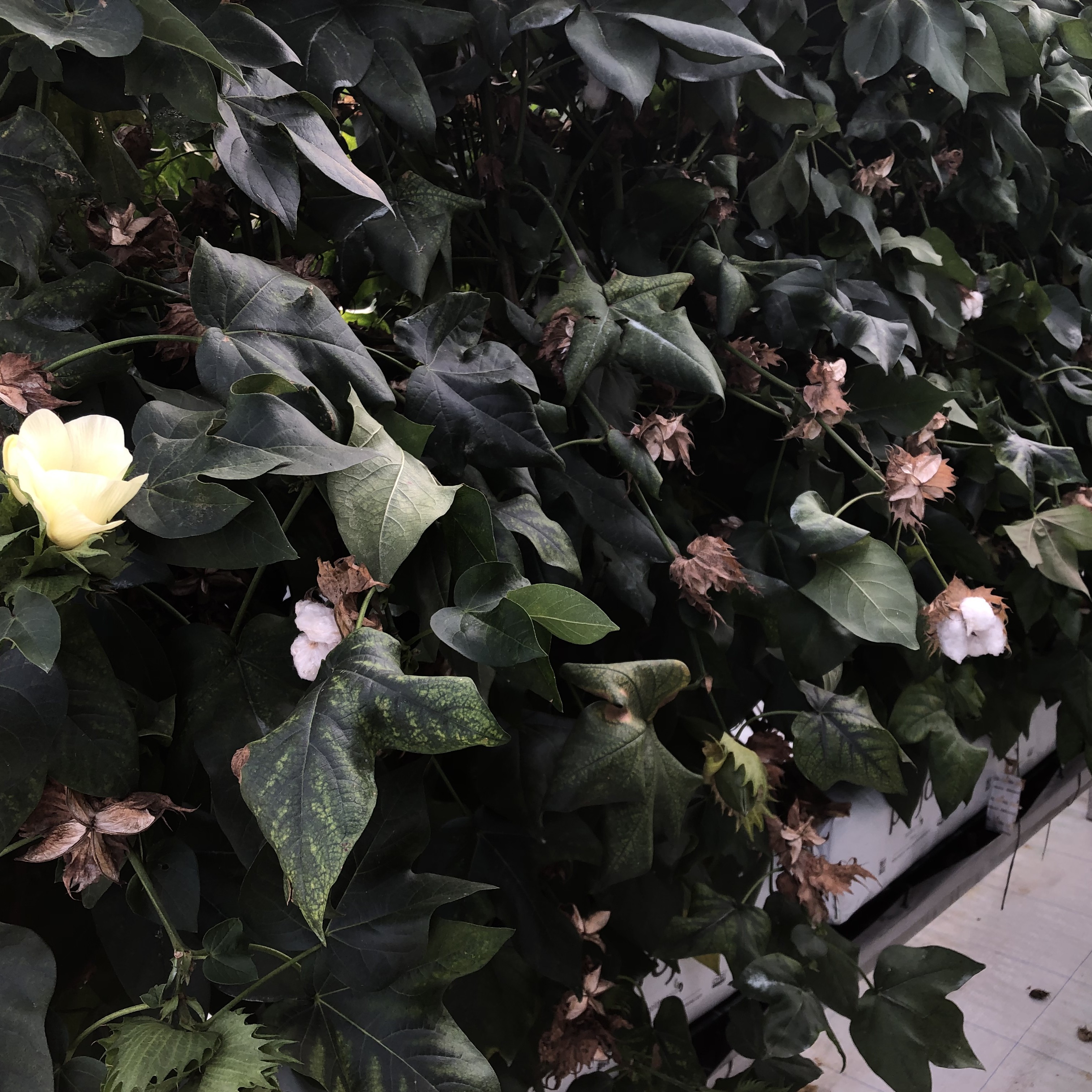}
    \caption{}
\end{subfigure}

\caption{Different types of images found within the dataset: (a) Clear, close view (b) Medium-far view (c) Sunny (d) Cloudy (e) Leaf occlusion (f) Cotton occlusion (g) Blur by focal distance (h) Perpendicular view (i) Angled view.}
\label{fig:dataset_image_types}
\end{figure*}

Fig.~\ref{fig:dataset_image_types} shows the different light conditions, perspectives and visual characteristics of \dName. We wanted to create diverse dataset, that encapsulates the same conditions found in the field. For all \dName's classes, the dataset contains clear, close shots (Fig.~\ref{fig:dataset_image_types} (a, g)), occlusions by other bolls or leaves (Fig.~\ref{fig:dataset_image_types} (e, f)), views perpendicular to the plants (Fig.~\ref{fig:dataset_image_types} (h)), angled shots in different distances (Fig.~\ref{fig:dataset_image_types} (i)), various distances with high level of detail (Fig.~\ref{fig:dataset_image_types} (a, b)), as well as differentiated light conditions (Fig.~\ref{fig:dataset_image_types} (c, d)). The degrees of freedom included in \dName allows the complete representativeness of environmental conditions found in the orchard.

In the end, pictures were taken to capture the maximum representativeness of cotton shapes in different stages, as well as different light and occlusion conditions. We see that obtaining a true-to-life representation of each class, with enough samples to generalise cotton morphology, is a complicated task. For this reason, it is essential to get numerous images for each class, especially those that exhibit a diverse morphology, and thus be able to develop a specialist and abstract view model that overlooks details insignificant to the representation and learns key features of cotton growth phases. 

\subsection{Data augmentation}

Since the number of images for training is small, and taking into account the complexity of the classes we want to detect, new images have to be generated through artificial methods. Many methods are available to manipulate images and produce new pictures for the model to train with, but we will explore only a subset of them. Because the greenhouse presents a variety of lighting and occlusion conditions that affect how cotton bolls appear, the main focus is to increase the dataset size by applying image transformation techniques.

We will choose simple techniques that do not excessively alter the colour properties of the images, since any change outside the distribution of the set would result in lower detection accuracy. Therefore, we will perform simple geometric transformations, as well as slight changes in colour properties. For this reason, the proposed techniques are: rotations (90, 180 and 270 degrees), flippings (horizontally and vertically), random croppings, random occlusions and colour enhancements (random changes in brightness/contrast, hue/saturation values and gamma values). These transformations help the network by providing more views to specialize in the properties of the cotton crop. All enhancement techniques are applied using the Albumentations python library~\citep{albumentations}.

\section{Object detection evaluation concepts} \label{ODM}

For evaluating the performance of the model on \dName, recall and precision metrics will be used:
\begin{align*}
    Recall = \frac{TP}{TP + FN}
\end{align*}
\begin{align*}
    Precision = \frac{TP}{TP + FP} \,,
\end{align*}
where $TP$, $FP$ and $FN$ refer to the number of true positives, false positives, and false negatives detected by the model. $Recall$ ($R$) measures the ability to capture all relevant objects in images. $Precision$ ($P$) quantifies the accuracy of positive predictions made by the model. Precision and recall can also be called sensitivity and specificity, respectively.

Another key concept is the intersection over union ($IoU$), which is the proportion of overlapped area between the predicted bounding box and the ground truth bounding box
\begin{align*}
    IoU = \frac{b\cap b^{gt}}{b\cup b^{gt}} = \frac{TP}{TP+FP+FN} \,,
\end{align*}
where $b$ and $b^{gt}$ are the predicted and ground truth bounding boxes, $(\cap,\cup)$ the intersection and union operators over the area in range $[0,1]$, respectively. Once the IoU has been defined, we can talk about the average precision (AP) and the mean average precision (mAP)
\begin{align*}
    AP = \int_0^1P(R)dR
\end{align*}
\begin{align*}
    mAP = \frac{1}{N}\sum_1^N\int_0^1 P(R)dR \,.
\end{align*}

Given the number of classes $N$. The average precision of every class is calculated with the precision-recall curve for a given intersection over the union threshold. $mAP50$ is the mean average precision for $IoU = 0.5$, related to ``easy detections'', and $mAP50-95$ is the mean average precision for $IoU \in [0.5, 0.95]$, with steps of $0.05$ units, related to performance in average detections, taking into account difficult ones.

The complexity of the model is computed with the number of parameters and floating point operations per second (GFLOPs), and the latency of the prediction is given by the frames per second (FPS) the model can compute. With $L$ a layer of the convolutional model
\begin{align*}
    FPS = \frac{N}{T}
\end{align*}
\begin{align*}
  FLOPs = \sum_{\forall L} 2 \cdot C_{\text{in}} \cdot K_h \cdot K_w \cdot C_{\text{out}} \cdot H_{\text{out}} \cdot W_{\text{out}}
\end{align*}
\begin{align*}
    \# Parameters = \sum_{\forall L} \left( C_{\text{in}} \cdot K_h \cdot K_w \right) \cdot C_{out} \,,
\end{align*}
where $N$, $T$ are the number of processed images and time consumed to predict these $N$ images, respectively, $C_{in}$ the number of channels from input, $C_{out}$ the number of channels from output, ($K_h$,$K_w$) the kernel size, ($H_{out}$,$W_{out}$) the height and width of the output feature map. These formulas apply to each convolutional layer, so computation needs to be done for every layer of the model.

\section{Proposed blocks and bounding box loss function}

Based on current benchmarks for object detection models on COCO 2017~\citep{roboflowComputerVision}, as well as the literature reviewed previously, the YOLO family of models offers the best balance between versatility, ease of use, lightness and mAP50 accuracy. We take YOLO11 as our starting point because of its simple architecture and little usage of attention mechanisms (only C2PSA). It's key to choose the most fitting architecture for our case study, as this step grants the best accuracy. We are testing state-of-the-art blocks and attention mechanisms to enhance feature extraction and refinement, thus obtaining better overall metrics. The objective of this work is to create a lightweight but robust, custom YOLO-like architecture to be deployed in mobile robotic platforms.

State-of-the-art suggests that no specific architecture is suitable for every problem, but instead some combinations of convolutional blocks and attention help model shapes, colours and other properties of crops, enhancing feature extraction, achieving better detection. Our objective is to demonstrate that \dName is the most suitable model for detecting cotton bolls in every stage of growth.

\subsection{Squeeze and Excitation and SEConvBlock}

\begin{figure}[t]
    \includegraphics[width=\columnwidth]{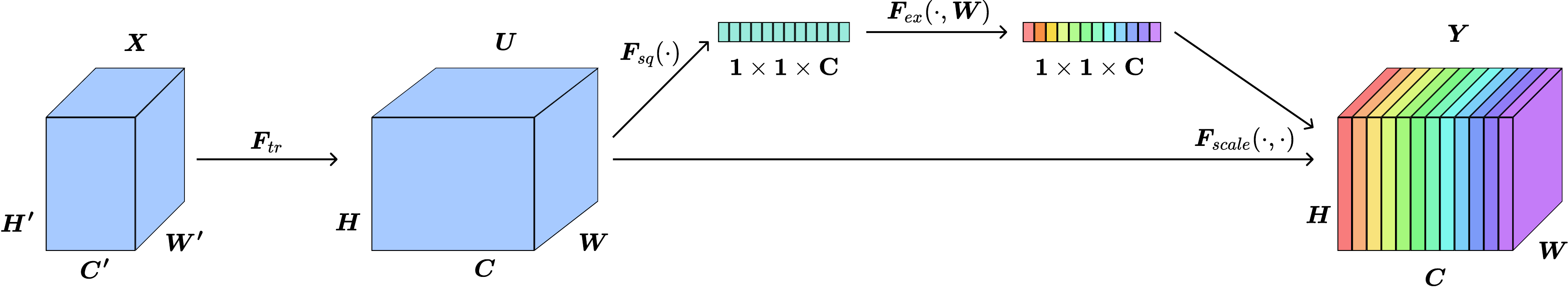}
    \caption{The Squeeze and Excitation (SE) module.}
    \label{fig:SE_block}
\end{figure}

To improve the refinement of the feature maps in the neck, we use Squeeze and Excitation (SE)~\citep{hu2019squeezeandexcitationnetworks} after each convolutional block. SE models interdependencies between channels to recalibrate channel-wise feature maps adaptively. Fig.~\ref{fig:SE_block} shows a visual representation of the SE module.

Because SE can be put after convolutional transformations, we suppose a previous operation to the block, where the transformation takes a given feature map $X \in \mathbb{R}^{H'\times W'\times C'}$ and produces $U = [u_1, u_2,..., u_C] \in \mathbb{R}^{H\times W\times C}$. The \textit{squeeze} operation exploits channel dependencies by applying a channel-wise global average pooling, obtaining a channel descriptor. The result is a statistic $z \in \mathbb{R}^C$ for every channel $C$ in $U$:
\begin{align*}
    z_c = {F}_{sq}\left(u_c\right) = \frac{1}{H \times W} \sum_{i=1}^{H} \sum_{j=1}^{W} u_c(i, j).
\end{align*}

Consequently, the \textit{excitation} operation consists of an adaptive recalibration, enabling the network to learn non-linear, non-mutually exclusive interactions between channels, allowing multiple channels to be emphasised. This is achieved by applying a gating mechanism:
\begin{align*}
    {s} = {F}_{\text{ex}}\left({z}, {W}\right) = \sigma \left( g({z}, {W}) \right) = \sigma \left( {W}_2 \delta ({W}_1 {z}) \right),
\end{align*}
where $\delta,\: \sigma$ refers to the ReLU and Sigmoid activation functions, respectively, $W_1 \in \mathbb{R}^{\frac{C}{r}\times C}$ and $W_2 \in \mathbb{R}^{C \times \frac{C}{r}}$. $(W_1, W_2, \delta, \sigma)$ form two fully connected (FC) layers creating a simple bottleneck. Finally, a re-escalation of the intermediate feature map $U$ is done with the previously obtained activations $s$:
\begin{align*}
    \widetilde{x}_c = {F}_{scale}\left(u_c, s_c\right) = s_c u_c, 
\end{align*}
where $\widetilde{X} = [\widetilde{x}_1, \widetilde{x}_2,...,\widetilde{x}_C]$ and ${F}_{scale}(u_c, s_c)$ is the channel-wise multiplication by the scalar $s_c$.

Squeeze and Excitation is applied at the end of the backbone's convolutional blocks, so it becomes a squeezed and excited convolutional block. We call this union {SEConvBlock}: 
\begin{align*}
    {Y_{SE}} = SE \left( \phi \left(BN\left(f^{k\times k} \left({X}\right)\right) \right) \right) \,,
\end{align*}
where $f^{k\times k}$ the convolutional kernel, $BN$ the batch normalisation operation, $\phi$  the SiLU activation function, and $SE$ the squeeze and excitation operation. This combination is very helpful to add minimal, lightweight attention to convolutional blocks, helping focus on relevant regions after learning representations of past cotton instances.

\subsection{Attention mechanisms}    

Because \dName classes can adopt plenty of shapes, complex to detect and discern from one another, especially when light, optical blurring or other elements modify the scene, attention is key. The model must learn the most general representation of what cotton is, in all of its growth stages. For this reason, we put special emphasis on attention blocks.

YOLO11 takes advantage of the C2PSA block (Cross Stage Partial with Spatial Attention)~\citep{yolo11} at the end of its backbone, focusing on relevant image regions, enhancing attention in small parts and occluded instances, and emphasising spatial relevance in outgoing feature maps. In our case, this block is replaced by a series of Simple Attention Module (SimAM)~\citep{GONG2025101209} after every C3k2 backbone bottleneck. SimAM offers channel-wise and spatial-wise attention without adding gradients and with little floating operations per second, computing the importance of each neuron (value in the feature map). Being a lightweight and fast block makes it very promising for our case study.

Furthermore, Parallel Hybrid Attention Mechanism (PHAM)~\citep{pham_block} blocks are added, prior to the neck's C3k2 bottlenecks, to pass spatially, channel-wise, coordinate-wise improved feature maps. By adding a post-processing step with PHAM blocks, we obtain more informed feature maps, where the relevant regions of the image are enhanced. Although PHAM is heavier in parameter size and floating operations per second than SimAM, PHAM offers local and non-local feature computations for the input feature maps, as well as reducing the weight input of irrelevant background information~\citep{pham_block}.

\begin{figure}[t]
    \includegraphics[width=\columnwidth]{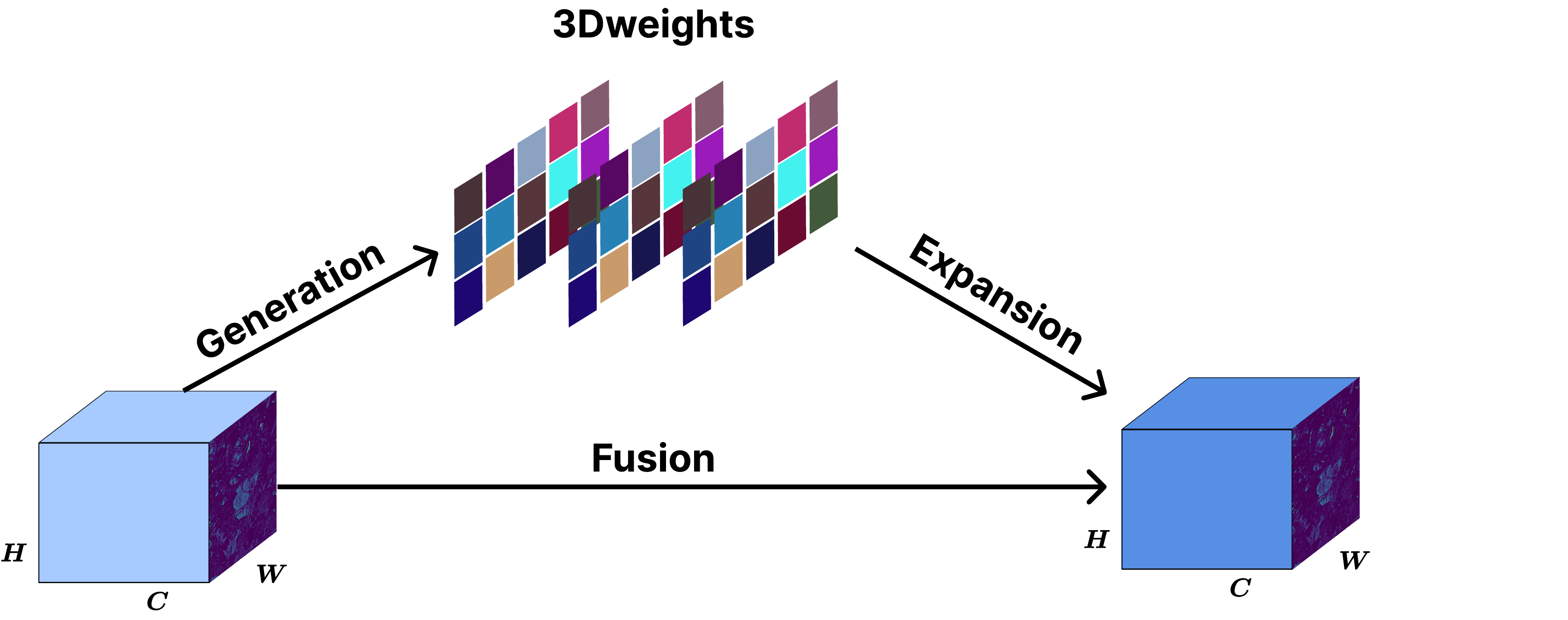}
    \caption{The simple attention (SimAM) module.}
    \label{fig:SimAM_block}
\end{figure}

\textbf{Simple Attention Module}: SimAM (Fig.~\ref{fig:SimAM_block}) is a simple, yet effective attention mechanism that focuses on 3-D attention without adding parameters to the network~\citep{pmlr-v139-yang21o}. This parameter-free attention module assigns 3D weights to every neuron and calculates the relevance of each one through an energy function inspired by spatial suppression~\citep{pmlr-v139-yang21o}. SimAM establishes an energy optimisation problem, where the target neuron $t$ has to be distinguishable, and the surrounding ones suppressed, measuring the linear separability between a neuron and the others based on the following energy function:
\begin{align*}
e_t(w_t, b_t, {y}, x_i) &= \frac{1}{M - 1} \sum_{i=1}^{M - 1} \left(-1 - (w_t x_i + b_t)\right)^2 \\
&\quad + \left(1 - (w_t t + b_t)\right)^2 + \lambda w_t^2 \,,
\end{align*}
obtaining a fast closed-form solution of $w_t$, $b_t$ with the mean and variance over all neurons except $t$ in that channel:
\begin{align*}
    w_t = -\frac{2\left(t-\mu_t\right)}{\left(t-\mu_t\right)^2 + 2\sigma_t^2 + 2\lambda}
\end{align*}
\begin{align*}
    b_t = -\frac{1}{2}\left(t+\mu_t\right)w_t \,.
\end{align*}
\begin{align*}
    \begin{split}
        \mu_t &= \frac{1}{M-1}\sum_{i=1}^{M-1}x_i \\
        \sigma_t^2 &= \frac{1}{M-1}\sum_i^{M-1}\left(x_i-\mu_t\right)^2
    \end{split}
\end{align*}
for $\mu_t$ and $\sigma_t^2$ the mean and variance over all neurons except $t$ in that channel. Assuming that all pixels in a single channel follow the same distribution, we get:

\begin{equation}
    e^*_t = \frac{4\left(\hat{\sigma}^2 + \lambda\right)}{\left(t - \hat{\mu}\right)^2 + 2\hat{\sigma}^2 + 2\lambda} \,,
    \label{eq:SimAMminimalEnergyFunc}
\end{equation}
for $\hat{\mu} = \frac{1}{M-1}\sum_{i=1}^{M-1}x_i$, $\hat{\sigma}^2 = \frac{1}{M-1}\sum_i^{M-1}\left(x_i-\hat{\mu}\right)^2$ the mean and variance over each channel.

Eq. (\ref{eq:SimAMminimalEnergyFunc}) computes the lower energy $e^*_t$ for neuron $t$. The importance of each neuron is computed as ${1}/{e^*_t}$, therefore the output feature map is applied as $X * 1/E$, for $X$ the input feature map, $E$ all $e^*_t$ across channel and spatial dimensions.

\textbf{Parallel Hybrid Attention Mechanism}: PHAM (Fig.~\ref{fig:PHAM_block}) allows refinement of intermediate feature maps with minimal computational cost, applicable to most feed-forward vision models~\citep{pham_block}. In practice, PHAM is added in the backbone for better feature extraction (~\citep{Li2025pineappleYOLO, wang2025insectYOLO}), but we want to test different locations to see where the block is more effective.

\begin{figure}[t]
    \includegraphics[width=\columnwidth]{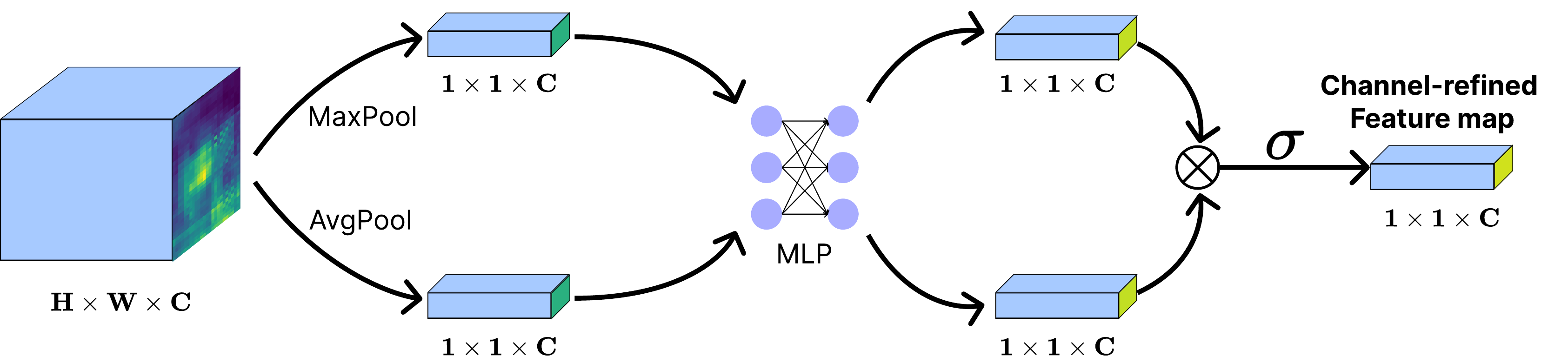}
    \caption{The channel gate from the PHAM block.}
    \label{fig:Channel_gate}
\end{figure}

\begin{figure}[t]
    \includegraphics[width=\columnwidth]{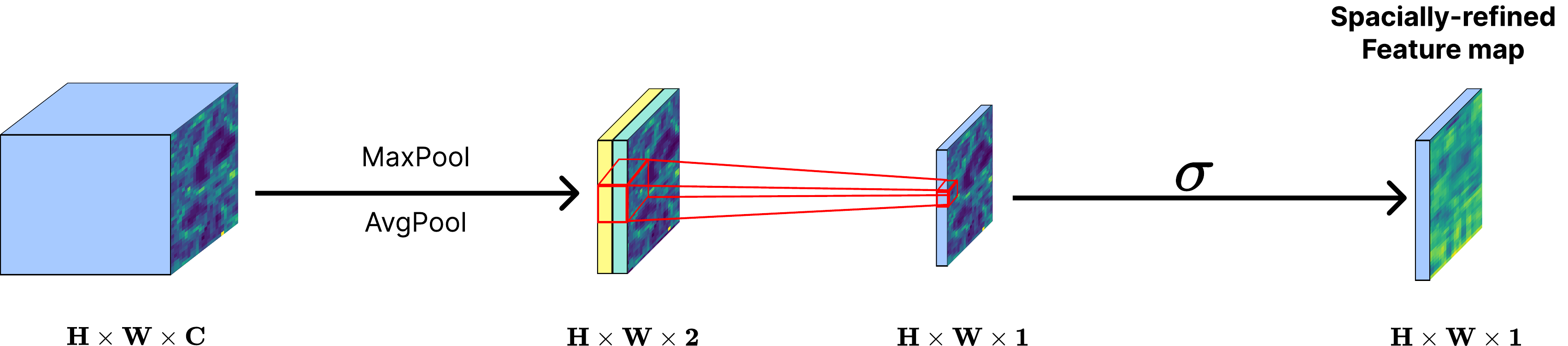}
    \caption{The spatial gate from the PHAM block.}
    \label{fig:Spatial_gate}
\end{figure}

\begin{figure}[t]
    \includegraphics[width=\columnwidth]{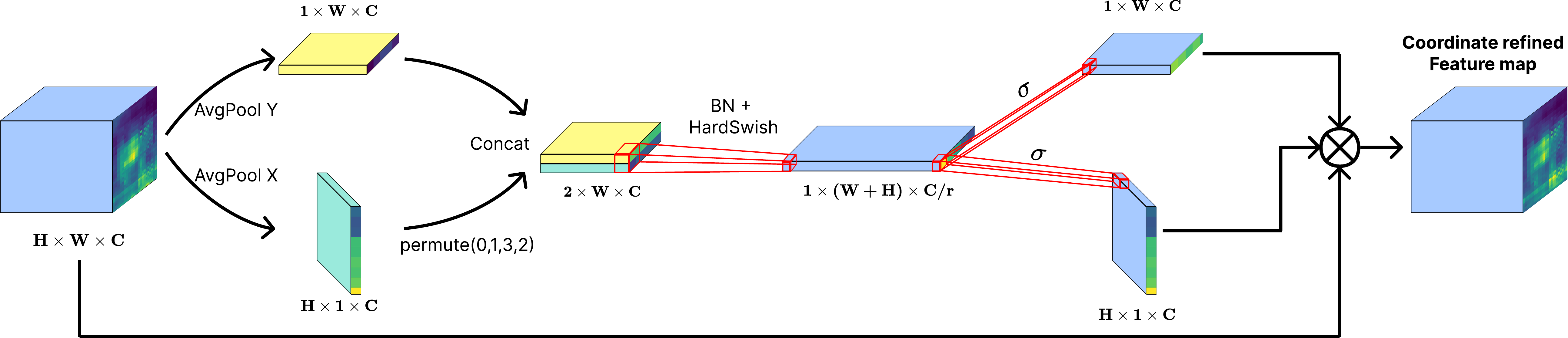}
    \caption{The coordinate gate from the PHAM block.}  \label{fig:Coordinate_gate}
\end{figure}

\begin{figure}[t]
    \includegraphics[width=\columnwidth]{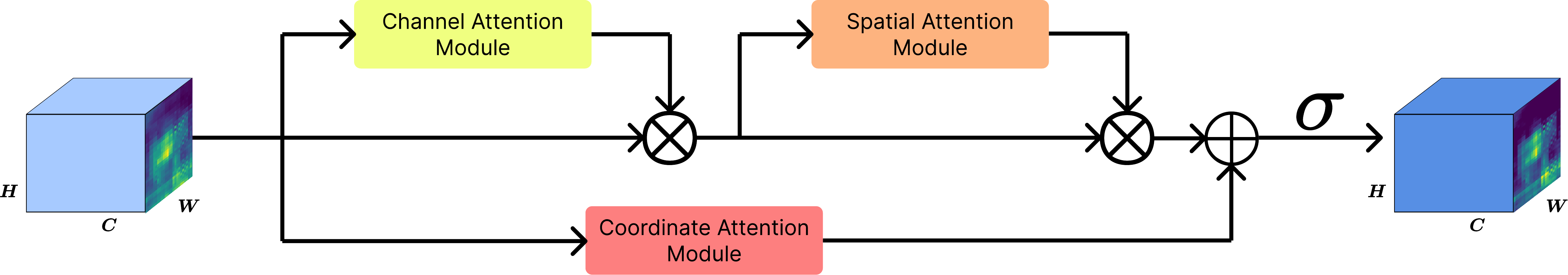}
    \caption{The Parallel Hybrid Attention Module (PHAM).}
    \label{fig:PHAM_block}
\end{figure}

PHAM consists of a Convolutional Block Attention Module (CBAM)~\citep{woo2018CBAMblock} attached to a Coordinate gate. More specifically, PHAM is formed by a Channel gate (Fig.\ref{fig:Channel_gate}) and a Spatial gate (Fig.~\ref{fig:Spatial_gate}), and a Coordinate gate (Fig.~\ref{fig:Coordinate_gate}). Its objective is to refine the input feature map to extract information from relationships channel-wise, spatial-wise and coordinate-wise. In the image, the block enhances what is important (channel-wise), where the relevance is found (spatial-wise) and long-range relations (coordinate-wise).

As is shown in Fig. \ref{fig:Channel_gate}, given the input feature map $X \in \mathbb{R}^{H\times W\times C}$, CAM first computes the global maximum pooling and global average pooling channel descriptors, channel-wise, of $F$, and feeds it to a multilayer perceptron $W$ of two layers and a ReLU activation between layers. After that, one tensor is added to the other element-wise, and the sigmoid function is applied. As a result, we obtain the Channel Attention $M_C$, which is multiplied element-wise to the original feature map, obtaining the channel-refined feature $X_c$.
\begin{align*}
\begin{split}
    M_c({X}) & = \sigma \left( MLP\left( Pool_{avg}\left({X}\right)\right)\right) +
    \left(MLP\left( Pool_{max}\left( {X}\right)\right)\right) \\
    & = \sigma \left({W_1} \left( {W_0} \left( {X_{avg}^c}\right)\right) + {W_1} \left( {W_0} \left( {X_{max}^c}\right)\right) \right)
\end{split}
\end{align*}
\begin{align*}
    {X_c} = M_c({X}) \otimes {X} \,,
\end{align*}
where $\sigma$ is the sigmoid function, ${W_0}$ and ${W_1}$ are the MLP weights, ${W_0}, {W_1} \in \mathbb{R}^{C/r\times C}$. The ReLU activation function is followed by ${W_0}$.

$X_c$ is then passed to SAM, which computes the global maximum and average pooling descriptors, but spatial-wise, and concatenates them, producing a tensor of shape $H\times W\times 2$. It is fed into a convolutional block ($c_{in} = 2, c_{out} = 1, k=7, s=2$) to finally apply a sigmoid function. The produced spatially-focused feature map $M_s(X)$ is then multiplied element-wise by $F_c$ to compute a channel and spatially refined feature map ${X_{cs}}$.
\begin{align*}
\begin{split}
    M_s({X}) & = \sigma \left( f^{7\times 7} \left( \left[ Pool_{avg}({X}); Pool_{max}({X})\right]\right)\right) \\
    & = \sigma \left( f^{7\times 7} \left( [{X_{avg}^s};{X_{max}^s}]\right)\right)
\end{split}
\end{align*}
\begin{align*}
    {X_{cs}} = M_s({X_c}) \otimes {X_c} \,,
\end{align*}
where $f^{7\times 7}$ is the convolutional operation with kernel size ($7\times7$), 
${X_{avg}^s}, {X_{max}^s} \in \mathbb{R}^{1\times H \times W}$. 

At the same time, $X$ is passed to the Coordinate Attention Module. This block breaks down channel attention into two one-dimensional feature encoding processes that combine features across various directions. Average pooling in both width and height dimensions is performed, obtaining $X_w$ and $X_h$ for them to be concatenated, passed through a 2D convolution, batch normalization and a hard swish activation for non-linearity, to obtain a feature map of dimensions $\mathbb{R}^{C/r\times 1\times (W+H)}$. Then two parallel 2D convolutions, plus sigmoid activations, are applied to obtain two feature maps $X_w'$ and $X_h'$. Therefore, $X$, $X_w'$ and $X_h'$ are multiplied, with $X_{coord}$ the output feature map of the coordinate attention module. For $M_{coord}$ the Coordinate Attention Module operation and $\delta$ the ReLU activation function, we have:
\begin{align*}
    y = \phi\left(BN\left(f^{1\times1}\left[ Pool_{avg}^x({X});Pool_{avg}^y({X})\right] \right)\right)
\end{align*}
\begin{align*}
    \begin{split}
    (x_h, x_w) & = \mathrm{split}(y, H, W) \\
    X'_h & = \sigma \left( f^{1\times 1}\left( x_h\right)\right), \\
    X'_w & = \sigma \left( f^{1\times 1}\left( \mathrm{permute}(x_w, 2, 3)\right)\right)
\end{split}
\end{align*}
\begin{align*}
    M_{coord}({X}) = {X_{coord}} = {X} \otimes X'_w \otimes X'_h \,.
\end{align*}

Finally, the PHAM operation is described as
\begin{align*}
    {Y} = \delta \left( {X_{cs}} \oplus {X_{coord}}\right) \,.
\end{align*}

\subsection{Upsampling operator}

YOLO11 uses an upsample operation with a scale factor of 2 and the nearest neighbour as the upsampling algorithm. Although this operation is fast, it does not add learnable parameters to learn, and can lose information from previous convolutions, rattling the relations between gradients from the feature maps. To address this problem, we replace the upsample operation with CARAFE (Content Aware ReAssembly of FEatures)~\citep{wang2019carafecontentawarereassemblyfeatures}. CARAFE enlarges the field of view, aggregating contextual information, as well as adding instance-specific content-aware handling. This solution adds minimal overhead to the overall model, doubling the height and width of the feature maps without reducing the number of channels, fitting perfectly as a substitution for the traditional upsampling operator.

\begin{figure}[t]
    \includegraphics[width=\columnwidth]{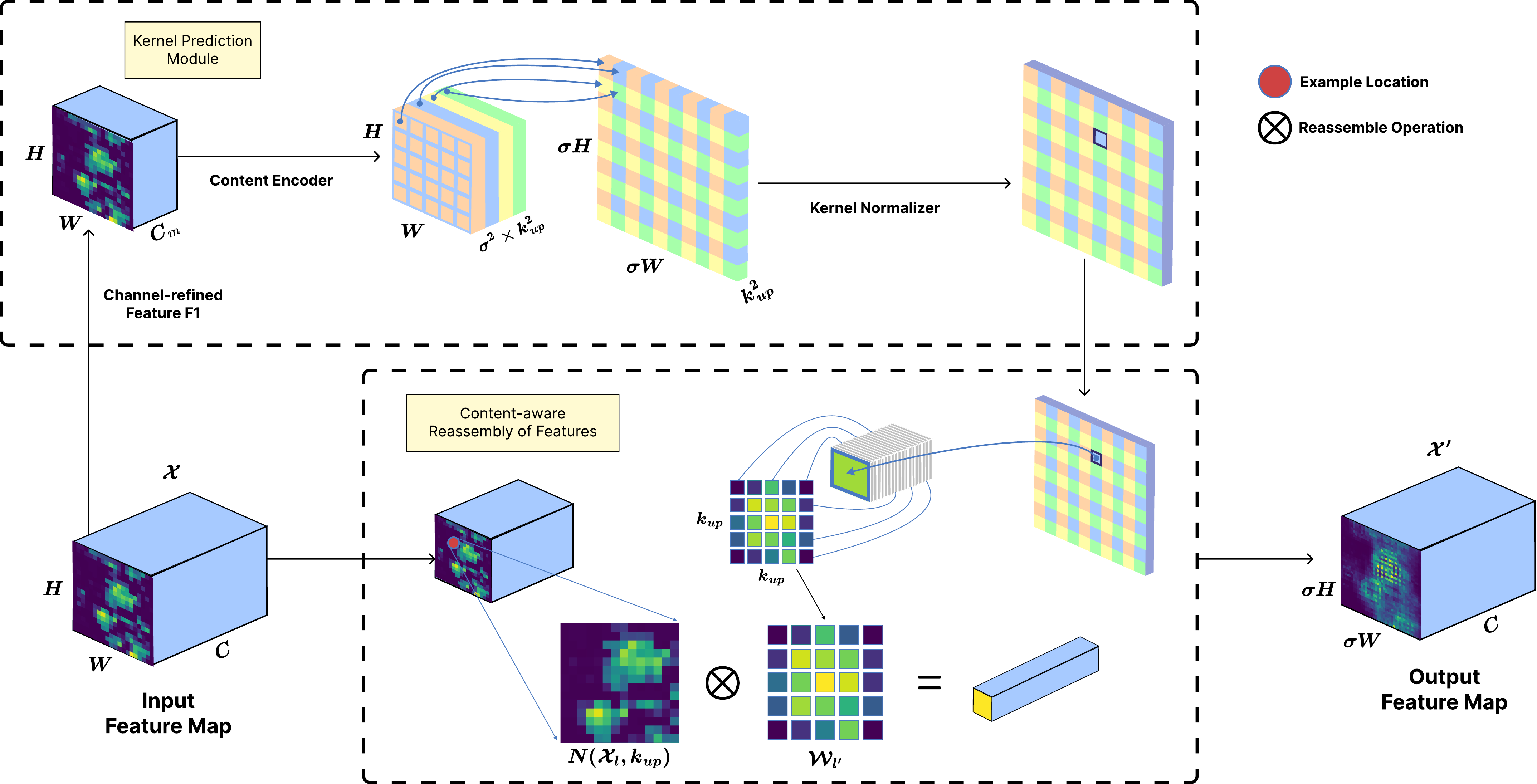}
    \caption{The Content Aware ReAssembly of FEatures (CARAFE) module.}
    \label{fig:CARAFE_block}
\end{figure}

CARAFE (Fig.~\ref{fig:CARAFE_block}) consists of the Kernel Prediction Module (KPM) and the Content-Aware Reorganization Module (CARM). The KPM is formed by a channel compressor, a content encoder and a kernel normalizer.

Given a feature map $X$ of shape $H\times W\times C$ and a factor $\sigma = 2$ in this case, CARAFE will produce another feature map ${X}'$ of shape $\sigma H\times \sigma W\times C$. The location-wise kernel $\mathcal{W}_{l'}$ is computed by the KPM module in Eq. (\ref{eq:content_encoder}), and the output feature map ${X}'$ is computed by the CARM module in Eq. (\ref{eq:kernel_prediction_module}). First, the channel compressor takes a feature map of dimensions $H\times W\times C$ and convolves it to compress the number of channels from $C$ to $C_m$, to make future computations less expensive. $C_m$ is then passed to the content encoder, which applies a convolution layer of kernel size $k_{encoder} = k_{up} - 2$, based on empirical results from the original paper~\citep{wang2019carafecontentawarereassemblyfeatures}. Finally, the kernel normaliser applies a spatial soft-max operation to every $k_{up}\times k_{up}$ reassembly kernel.

\begin{equation}
    \mathcal{W}_l = \psi(N({X}_l, k_{\text{encoder}}))\,.
    \label{eq:content_encoder}
\end{equation}

The CARM module reassembles the features within a local region via $\phi$, which takes the neighbours of some pixel $l = (i,j)$ in the input feature map $X$ about the kernel $k_{up}$, using the location-wise kernel $\mathcal{W}_{l'}$ from the content encoder

\begin{equation}
    X'_l = \phi(N({X}_l, k_{\text{up}}), \mathcal{W}_l) \,,
    \label{eq:kernel_prediction_module}
\end{equation}
where $k$ is an integer that defines a kernel size $(k \times k)$, $N({X}_l, k)$ is the $k\times k$ subregion neighbour to ${X}_l$, $\psi$ is the kernel prediction module operation, and $\phi$ is the weighted sum operator for every pixel in ${X}'_{l}$ using the location-wise kernel $\mathcal{W}_l$ obtained in Eq. (\ref{eq:content_encoder}). With CARAFE as our upsampling operator, we achieve efficiency in memory and latency, in exchange for adding little overhead, enabling gradients (learnability) to upsampling operations.

\subsection{Downsampling operator}

Even though YOLO11 uses regular convolutional blocks for downsampling in the neck, previously to the detection head, this block adds a number of parameters that grow exponentially, and given an input feature map with numerous channels, it can be a huge bottleneck for efficiency and lightness. To solve this problem, we propose the SCDown (Spatial-Channel Decoupled Downsampling) block, previously used in YOLOv10. This block addresses the high computational costs related to parallel convolutional kernels by first reducing the number of channels, from $c_1$ to $c_2$ with a $1\times 1$ kernel, and then applying a separable depth $k\times k$ convolution to decrease the feature map's height and width.

This block is composed of two parts: a point-wise convolution and a depth-wise separable convolution:
\begin{align*}
    Y &= DWConv \left( PConv \left( X \right)\right) \\
    PConv(X) & = \phi \left( BN \left( f^{1\times 1} \left( X\right)\right)\right) \\
    DWConv(X) & =  BN \left( g^{k\times k}_{grups=c_2} \left( X\right)\right) \,,
\end{align*}
where $f, g$ are convolution operations, BN is the batch normalisation operation and $\phi$ is the SiLU activation function.

\subsection{Improvement of bounding box loss function}

YOLO11 uses CIoU (Complete Intersection over Union) loss for bounding boxes, Binary Cross Entropy for class loss and Distribution Focal Loss (DFL) for class imbalance.

For the new box loss, we propose Scylla IoU (SIoU)~\citep{gevorgyan2022scyllaiou} as a substitute for the default CIoU. SIoU takes into consideration four aspects of cost: angle, distance, shape and IoU cost, while CIoU only computes distance, shape and IoU. The core idea of the added angle cost is to inform the model about the direction to where the ground truth box is pointing, so we are not only informing about the distance, but giving the relative distance in $x$ and $y$ coordinate axes from the prediction to the ground truth. Angle cost ($\Lambda$) breaks down as:
\begin{align*}
    \Lambda = 1 - 2*\sin^2{\left( \arcsin{(x)} - \frac{\pi}{4}\right)} \,,
\end{align*}
where
\begin{align*}
    x &= \sin{(\alpha)} =\frac{c_h}{\sigma} \\
    \sigma &= \sqrt{\left(b_{c_x}^{gt} - b_{c_x}\right) + \left(b_{c_y}^{gt} - b_{c_y}\right)} \\
    c_h &= \max{\left(b_{c_y}^{gt}, b_{c_y}\right)} - \min{\left(b_{c_y}^{gt}, b_{c_y}\right)} \,,
\end{align*}
where $c_x$ and $c_y$ are the centre coordinates. The distance cost ($\Delta$) is formulated as:
\begin{align*}
    \Delta = \sum_{t=x,y} \left(1 - e^{-\gamma \rho_t}\right)
\end{align*}
where
\begin{align*}
    \rho_x = \left(\frac{b_{c_x}^{gt} - b_{c_x}}{c_w}\right)^2 \:,\: \rho_y = \left(\frac{b_{c_y}^{gt} - b_{c_y}}{c_w}\right)^2 \:,\: \gamma = 2 - \Lambda \,.
\end{align*}

The shape cost ($\Omega$) is defined as:
\begin{align*}
\Omega &= \sum_{t=w,h} \left(1 - e^{-\omega_t}\right)^\theta \\
\omega_{w} &= \frac{|w-w^{gt}|}{\max\left( w, w^{gt}\right)} \,, \\
\omega_{h} &= \frac{|h-h^{gt}|}{\max\left( h, h^{gt}\right)} \,.
\end{align*}
where $\theta$ defines the relevance of the shape cost to the overall computation. Experimentally, $\theta$ is defined between 2 and 6. Finally, SIoU loss is expressed as:
\begin{align*}
    L_{box} = 1 - IoU + \frac{\Delta + \Omega}{2}
\end{align*}

\section{Proposed YOLO11 modification}

\begin{figure}[t]
    \includegraphics[width=\columnwidth]{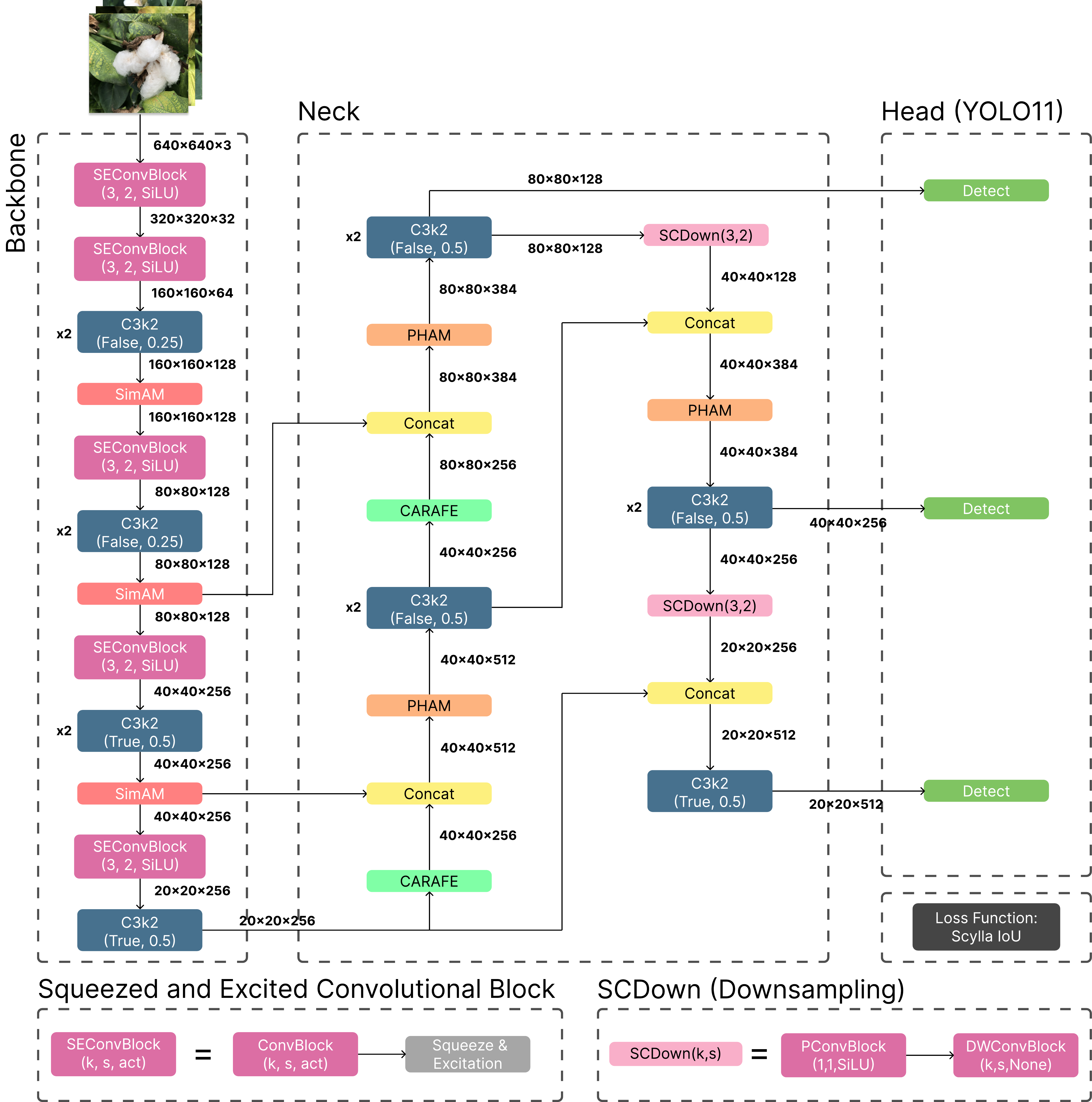}
    \caption{\modelName, the proposed YOLO11 architecture modification.}
    \label{fig:COTONET_architecture}
\end{figure}

Once we have studied the state of the art and the main convolutional blocks, attention mechanisms and loss functions from the state-of-the-art, we propose a modification of YOLO11 that we call \modelName (Fig.~\ref{fig:COTONET_architecture}). The main changes in the network are the addition of Squeeze and Excitation to the backbone's convolutional blocks, SimAM and PHAM for the main attention modules, and SCDown as the down-sampler for the neck's downward phase. The bounding box loss has been modified too, replacing CIoU with SIoU.

\section{Experiments} \label{EXP}

Once the structure of the new model has been established, we evaluate the robustness and reliability of the solution, ensuring that no other combination is more suitable than the one chosen for this case study. Therefore, we must perform ablation studies on each of the blocks and techniques described to ensure that their effectiveness is superior to the original solution and other possible solutions.

To achieve this, we will make modifications to \modelName, performing systematic training and comparing the metrics obtained in terms of accuracy and computational cost. The experiments to be carried out are:

\begin{enumerate}
    \item Comparative analysis between artificial augmentation techniques on \dName. We constructed different datasets from \dName and artificially augmented its training set. We trained \modelName with each of the datasets obtained and compared the metrics obtained (mAP50 and mAP50-95). We selected those techniques that offered greater model learning generalisation capacity on the dataset, i.e., the best validation metrics. 
    \item Comparative analysis, using augmented \dName, of validation metrics between standard YOLO models, ranging from nano to medium size, and those of \modelName.
    \item Ablation studies on the care blocks selected for the final model. Specifically, we performed ablation tests over:
    \begin{enumerate}
        \item Squeeze and Excitation, with the aim of demonstrating that adding it after a convolutional block provides considerable refinement of the feature maps obtained. The metrics of a ConvBlock without attention, a ConvBlock with SE, and a ConvBlock with an Efficient Channel Attention (ECA) module will be compared.
        \item The main attention mechanisms of \modelName (SimAM and PHAM), with the aim of proving that they model the interdependencies of values in feature maps, providing a better focus on regions that are important for subsequent detection. We will compare the absence of attention (no SimAM/PHAM), the use of SimAM, the use of PHAM, and the use of both.
    \end{enumerate}
    \item Comparative analysis by replacing the attention methods used with other state-of-the-art methods, with the aim of verifying that the blocks chosen are the most appropriate for our problem. These are: Convolutional Block Attention Module (CBAM)~\citep{woo2018CBAMblock}, Efficient Multi-Scale Attention (EMA)~\citep{ouyang2023EMAblock}, Multi-scale Aggregation Attention (MAA)~\citep{jin2025cameliaolifera}, Multidimensional Collaborative Attention (MCA)~\citep{mca} and Triplet Attention (TripletAttn)~\citep{tripletattention}.
    \item Bounding box loss function study, comparing the default loss function in YOLO11 (CIoU) to other loss functions in the literature and the proposed bounding box loss function.
\end{enumerate}

\begin{table}[t]
    \centering
    \small
    \setlength{\tabcolsep}{4pt}
    \begin{tabular}{p{0.32\columnwidth} p{0.63\columnwidth}}
    \toprule
    CPU & Intel(R) Xeon(R) Gold 6326 CPU @ 2.90GHz \\
    GPU & 2× NVIDIA A10 \\
    Operating system & Ubuntu 20.04.6 LTS \\
    CUDA version & CUDA 10.1 \\
    \bottomrule
    \end{tabular}
    \caption{Experimental configuration for training and validation processes.}
    \label{tab:hardware_software_specs}
\end{table}

\begin{table}[t]
    \centering
    \small
    \setlength{\tabcolsep}{4pt}
    \begin{tabular}{p{0.45\columnwidth} p{0.50\columnwidth}}
    \toprule
    \textbf{Hyperparameter} & \textbf{Value} \\
    \midrule
    Optimizer & SGD with Nesterov momentum \\
    Learning rate & 0.01 \\
    Momentum & 0.9 \\
    Loss weights (Bbox, Cls, DFL) & (7.5, 0.5, 1.5) \\
    Maximum epochs & 1000 (+ early stopping) \\
    Patience & 50 \\
    Batch size & 16 \\
    Image size & 640 \\
    Cosine learning rate & False \\
    \bottomrule
    \end{tabular}
    \caption{Hyperparameter settings for every model training and validation.}
    \label{tab:experimental_configuration}
\end{table}

\section{Experimental results} \label{RES}

Table~\ref{tab:hardware_software_specs} shows the local server's hardware and software specifications, and table~\ref{tab:experimental_configuration} shows hyperparameter configuration, for all model trainings analysed below. As we mentioned previously in section~\ref{ODM}, the analysis of the following results is carried taking $mAP50$ as our main indicator because it is a solid indicator of how well detects, specially for the case of bolls that are difficult to detect (small, out of focus, deformed, occluded, colour properties out-of-distribution...). $mAP50-95$ is our second indicator for model performance, giving information about how well easy detections are performed.

\subsection{Data augmentation} \label{DA}

To see the effectiveness of different combinations of data augmentation techniques, we start with a series of augmentations that, given the environmental and lighting conditions we are working with, we believe are suitable for achieving greater generalisation of the properties of cotton, and their phenotypical expressions within the greenhouse. Rotations, flips, and changes to the image's colour saturation and hue are subject of testing. 

Using \modelName, we apply data augmentation to \dName individually and with subsets of techniques that we believe work well together. Table~\ref{table:AugmentedDA_benchmark} shows the training results of \modelName with the augmented variants of \dName. The metrics show that data augmentation techniques have a distinct influence on the model's performance over the cotton dataset. Simple geometric transformations show complementary behaviour: rotation gets the highest recall (0.748), indicating a better detection of positive instances, while flipping favours precision (0.795), reflecting a compromise between both metrics.

Strategies based on content modifications, such as random cropping and random occlusion, increase accuracy but reduce recall, probably due to loss of contextual information. In contrast, photometric transformations, especially colour enhancement, maintain a more stable balance and show the most consistent mean average prediction values among individual techniques.

The Rotation + Flipping combination emerges as the optimal configuration, reaching the best values of mAP50 (0.811) and mAP50-95 (0.606). This result suggests that joint spatial diversification improves the generalisation of the model more effectively and robustly than single techniques.

\begin{table*}[t]
    \centering
    \begin{tabular}{lcccc}
        \hline
        \textbf{Augmentation}  & \textbf{Precision} & \textbf{Recall} & \textbf{mAP\textsubscript{50}} & \textbf{mAP\textsubscript{50:95}} \\
        \hline
            Rotation(90º,180º,270º)  & 0.744 & \textbf{0.748} & 0.8 & 0.571 \\
            Flipping(horizontally, vertically)  & 0.795 & 0.731 & 0.802 & 0.554 \\
            Random cropping  & \textbf{0.834} & 0.712 & 0.782 & 0.556 \\
            Colour enhancement  & 0.792 & 0.744 & 0.806 & 0.57 \\
            Histogram equalization  & 0.779 & 0.721 & 0.785 & 0.555 \\
            Random occlusion  & 0.825 & 0.679 & 0.788 & 0.544 \\
        \hline
            {Rotation + Flipping}  & {0.825} & {0.733} & \textbf{0.811} & \textbf{0.606} \\
        \hline
    \end{tabular}
    \caption{Training metrics for \modelName on augmented \dName datasets, given artificially augmented techniques.}
    \label{table:AugmentedDA_benchmark}
\end{table*}

\subsection{Proposed model comparison} \label{modelbench}

To prove the model's effectiveness, we perform a benchmark with all small-to-medium size YOLO models and compare them to \modelName's metrics. Table~\ref{table:yolo_benchmarks} shows the detection metrics for all standard YOLO models from nano to medium sizes, as well as \modelName, with the augmented \dName.

Table~\ref{table:yolo_benchmarks} shows that baseline YOLO models follow a correlation between computational cost and performance. We can see that larger variants tend to achieve better accuracy in prediction, seen in variants 8m, 9m and 10m, for example, which systematically exceed the smaller ``n'' and ``s'' versions. However, this increase in performance also shows the high computational costs of training and executing these models, exceeding limits for deployment in mobile robotic platforms. On the other hand, smaller models stand out for specific metrics, for example YOLOv9s standing out for its recall metric (0.761) and YOLO11m with the higher precision among them all (0.841), suggesting thorough detection and better suppression of false positives, respectively.

\modelName achieves the highest accuracy for easy (mAP50) and overall (mAP50-95) detections, with values of 0.811 and 0.606, respectively, surpassing all baseline models, reducing parameter size (7.6 M) with a light increase in floating point operations per second (27.8). We can come to the conclusion, as seen in the state of the art, that model complexity does not grant better results, but the wise placement of resources and convolutional blocks, together with the attention mechanisms, arranged in such a way as to maximise visual information extraction. \modelName preserves a good trade-off between precision and recall, and taking all metrics into consideration, we can conclude that the proposed architecture offers the best performance-computational cost ratio in comparison to standard YOLO variants. 

\begin{table*}[t]
    \centering
    \begin{tabular}{lcccccc}
        \hline
        \textbf{Model} & \textbf{Model Size (M)} & \textbf{GFLOPs} & \textbf{Precision} & \textbf{Recall} & \textbf{mAP\textsubscript{50}} & \textbf{mAP\textsubscript{50:95}} \\
        \hline
            YOLOv8n  & 3.1  & 8.9 & 0.769 & 0.707 & 0.762 & 0.553 \\
            YOLOv8s  & 11.1 & 28.8 & 0.800 & 0.697 & 0.771 & 0.566 \\
            YOLOv8m  & 25.9 & 79.3 & 0.819 & 0.725 & 0.807 & 0.604 \\
            \hline
            YOLOv9t  & \textbf{2.1} & 8.5    & 0.771 & 0.755 & 0.791 & 0.566 \\
            YOLOv9s  & 7.3  & 27.6 & 0.710 & \textbf{0.761} & 0.784 & 0.585 \\
            YOLOv9m  & 20.2 & 77.9 & 0.804 & 0.720 & 0.795 & 0.603 \\
            \hline
            YOLOv10n & 2.3  & 6.7  & 0.836 & 0.697 & 0.800 & 0.572 \\
            YOLOv10s & 7.2  & 21.6 & 0.796 & 0.732 & 0.791 & 0.578 \\
            YOLOv10m & 15.4 & 59.1 & 0.811 & 0.734 & 0.792 & 0.600 \\
            \hline
            YOLO11n  & 2.6  & \textbf{6.6}  & 0.815 & 0.708 & 0.781 & 0.561 \\
            YOLO11s  & 9.4  & 21.7 & 0.782 & 0.733 & 0.783 & 0.578 \\
            YOLO11m  & 20.1 & 68.5 & \textbf{0.841} & 0.701 & 0.797 & 0.596 \\
            \hline
            YOLO12n  & 2.6  & 6.7 & 0.774 & 0.676 & 0.774 & 0.550 \\
            YOLO12s  & 9.2  & 21.7 & 0.840 & 0.697 & 0.798 & 0.590 \\
            YOLO12m  & 20.2 & 68.1 & 0.796 & 0.738 & 0.800 & 0.593 \\
        \hline
            {\modelName} & {7.6} & {27.8} & {0.825} & {0.733} & \textbf{0.811} & \textbf{0.606} \\
        \hline
    \end{tabular}
    \caption{Training results on augmented \dName for the proposed algorithm and state-of-the-art YOLO baseline models.}
    \label{table:yolo_benchmarks}
\end{table*}

\subsection{Lightweight attention block comparative analysis}

One of the first contributions of this paper is the SEConvBlock. To prove its effectiveness, we compare the same concept (convolution, batch normalization, activation function, lightweight attention module) with other SE-like attention blocks. We use no attention (ConvBlock without attention) as a contrast, and adding ECA (Efficient Channel Attention) to build an ECAConvBlock.

Results to this comparative analysis can be found in table~\ref{table:ConvBlock_attention}. We see that SE offers the highest precision $(+0.016)$ and mAP50-95 $(+0.019)$, while losing some points in recall $(-0.03)$ and mAP50 $(-0.006)$. We come to the conclusion that Squeeze and Excitation offers the best accuracy (mAP50-95) with a minimal parameter overhead. In conclusion, we find that the SEConvBlock offers an improvement over the base convolutional block.

\begin{table*}[t]
    \centering
    \begin{tabular}{lcccccc}
        \hline
        \textbf{Attention} & \textbf{Added parameters} & \textbf{GFLOPs}  & \textbf{Precision} & \textbf{Recall} & \textbf{mAP\textsubscript{50}} & \textbf{mAP\textsubscript{50:95}} \\
        \hline
            No attention & \textbf{0} & \textbf{27.8} & 0.809  &    \textbf{0.763}   &   \textbf{0.817}   &   0.581 \\
            ECA & 3 & \textbf{27.8} & 0.78  &  0.742   &   0.814   &   0.587 \\
        \hline
            {SE} & {19.854} & \textbf{27.8} & \textbf{0.825} & {0.733} & {0.811} & \textbf{0.606} \\
        \hline
    \end{tabular}
    \caption{Lightweight attention operators comparative analysis after backbone's convolutional blocks for \modelName.}
    \label{table:ConvBlock_attention}
\end{table*}

\subsection{Attention block ablation study}

We perform an ablation study to see whether the proposed attention mechanisms are effective or not. In this experiment, we are replacing the attention blocks for the identity operation $I: x \rightarrow x$. After training \modelName for every possible combination (without attention, without SimAM, without PHAM). 

Table~\ref{table:attention_ablation} exposes that the addition of attention mechanisms to the architecture affects the performance of the model. Although attention mechanisms reduce the number of predicted instances (recall of 0.747 without attention), the combination of SimAM and PHAM offers the best sensitivity among tested configurations (precision of 0.825). An in-depth analysis of the table suggests that SimAM is primarily responsible for the precision rise, where only SimAM offers a sensitivity of 0.820. SimAM also lightly increases mean average precision metrics, proposing a more discriminative characteristic representation. PHAM adds model complexity (more parameters and floating point operations per second) but increases mAP50 (0.815), indicating a better detection for lower and less restrictive IoU thresholds.

The SimAM+PHAM combination offers the best global behaviour, with the higher precision (0.825) and higher mAP50-95 (0.606). We can come to the conclusion that both PHAM and SimAM complement each other, where SimAM offers consistency over predictions without computational penalty, PHAM reinforces spatial interdependency modelling. The increase in metrics confirms that integrating both attention mechanisms allows a more robust configuration, taking only a light increment in complexity.

\begin{table*}[t]
    \centering
    \begin{tabular}{lcccccc}
        \hline
        \textbf{Attention} & \textbf{Added parameters} & \textbf{GFLOPs} & \textbf{Precision} & \textbf{Recall} & \textbf{mAP\textsubscript{50}} & \textbf{mAP\textsubscript{50:95}} \\
        \hline
            No attention & \textbf{0} & \textbf{27.5} & 0.751 & \textbf{0.747} & 0.795 & 0.574 \\
            SimAM & \textbf{0} & \textbf{27.5} & 0.82 & 0.728 & 0.803 &  0.58 \\
            PHAM & 126196 & 27.8 &  0.787 & 0.729 & \textbf{0.815} & 0.572 \\
        \hline
            {SimAM + PHAM} & {126196} & \textbf{27.8} & \textbf{0.825} & {0.733} & {0.811} & \textbf{0.606} \\
        \hline
    \end{tabular}
    \caption{Ablation study of attention mechanisms in \modelName.}
    \label{table:attention_ablation}
\end{table*}

\begin{table*}[t]
    \centering
    \begin{tabular}{lcccccc}
        \hline
        \textbf{Attention} & \textbf{Added parameters} & \textbf{GFLOPs} & \textbf{Precision} & \textbf{Recall} & \textbf{mAP\textsubscript{50}} & \textbf{mAP\textsubscript{50:95}} \\
        \hline
            CBAM & 657740 & 28.1 & 0.815 & 0.735 & 0.808 & 0.587 \\
            EMA & 6624 & 28.9 & 0.789 & \textbf{0.781} & \textbf{0.825} & 0.603 \\
            MAA & 2153440 & 42.3 & 0.714 & 0.772 & 0.807 & 0.58 \\
            MCA & \textbf{98} & \textbf{27.6} & 0.739 & 0.757 & 0.798 & 0.57 \\
            TripletAttn & 1200 & 28.4 & \textbf{0.83} & 0.721 & 0.799 & 0.571 \\
        \hline
            {SimAM + PHAM} & {126196} & {27.8} & {0.825} & {0.733} & {0.811} & \textbf{0.606} \\
        \hline
    \end{tabular}
    \caption{Comparative analysis of different attention mechanisms in \modelName.}
    \label{table:attention_comparison}
\end{table*}

\subsection{Bounding box loss function study}

We want to test the different loss functions proposed in the section~\ref{EXP}. Given Complete IoU as bounding box loss function by default, we want to see if Scylla IoU builds a more informed model, that is, it reaches a minimum closer to the global gradient descend, and if any of the other loss functions can provide more information. 

Table~\ref{table:bb_loss_function_comparative} shows the metrics of \modelName trained with \dName with different loss functions of the bounding box. Classical formulations (CIoU and DIoU) exhibit competitive precision values but show limitations in recall and mAP50-95 metrics, suggesting a lack of robustness under high IoU values. GIoU offers a better balance between precision and recall, achieving a higher mAP50 value (0.817) but, once again, lacks accuracy for high IoU thresholds. For MDPIoU and FocalerIoU there's a priority in recall but do not achieve consistent improvements in mAP. These conclusions suggest that an increase in positive detections does not always translate into better spatial accuracy.

\begin{figure*}[h!]
\centering

\newcommand{\imgw}{0.16\textwidth}
\newcommand{\interspacea}{\hspace{1mm}}

\begin{subfigure}{\imgw}\includegraphics[width=\linewidth]{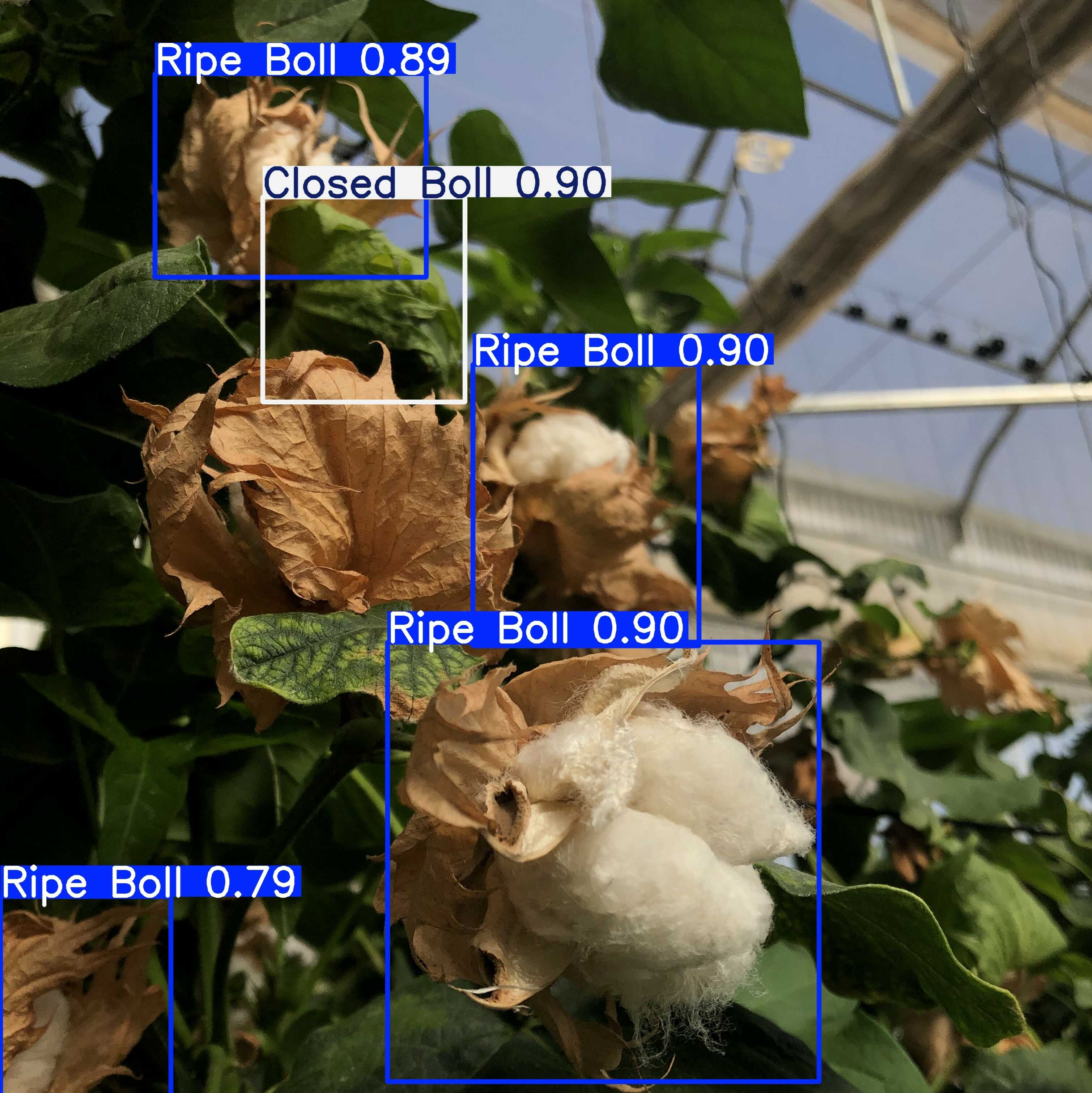}\end{subfigure}\interspacea%
\begin{subfigure}{\imgw}\includegraphics[width=\linewidth]{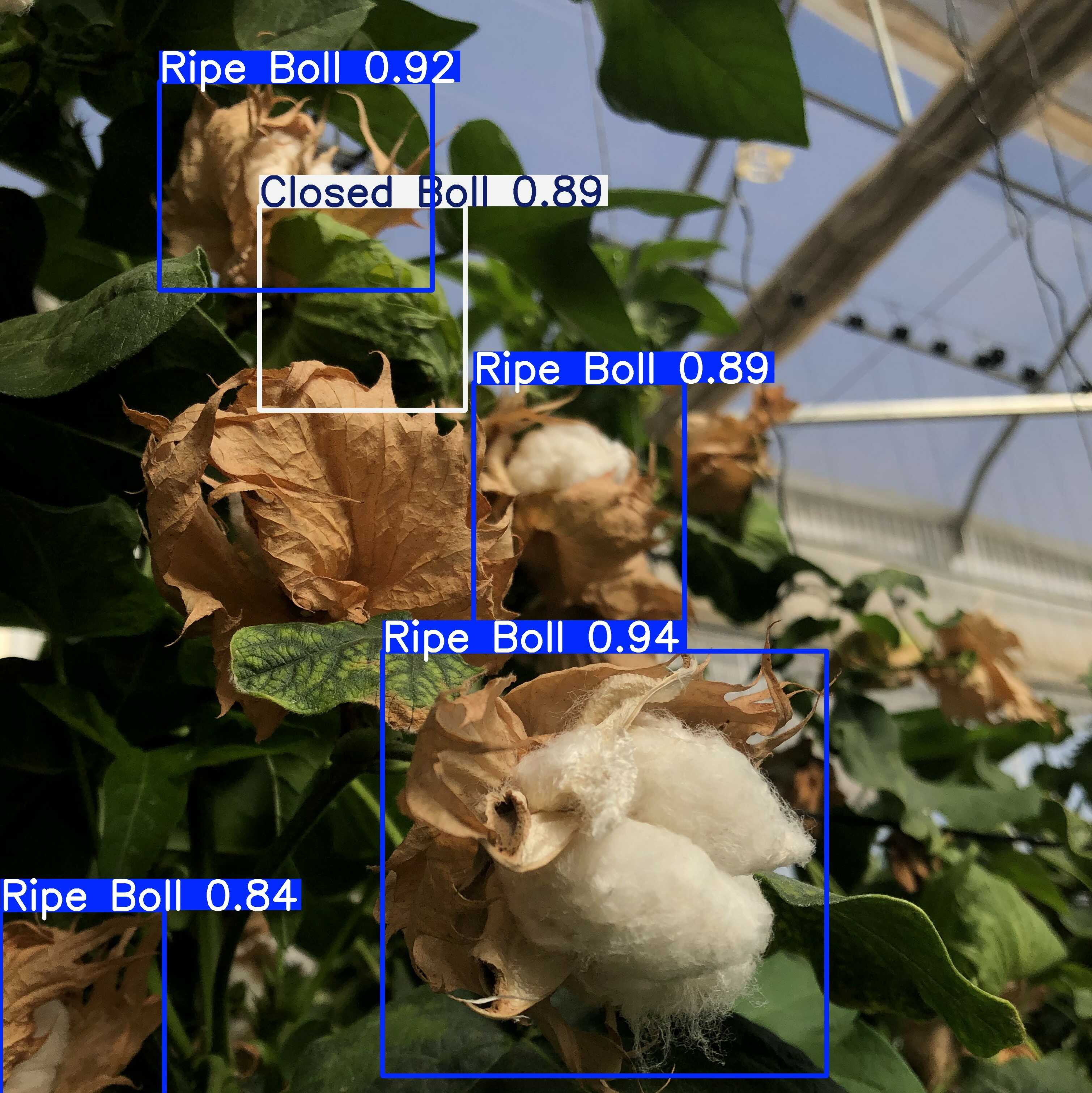}\end{subfigure}\interspacea%
\begin{subfigure}{\imgw}\includegraphics[width=\linewidth]{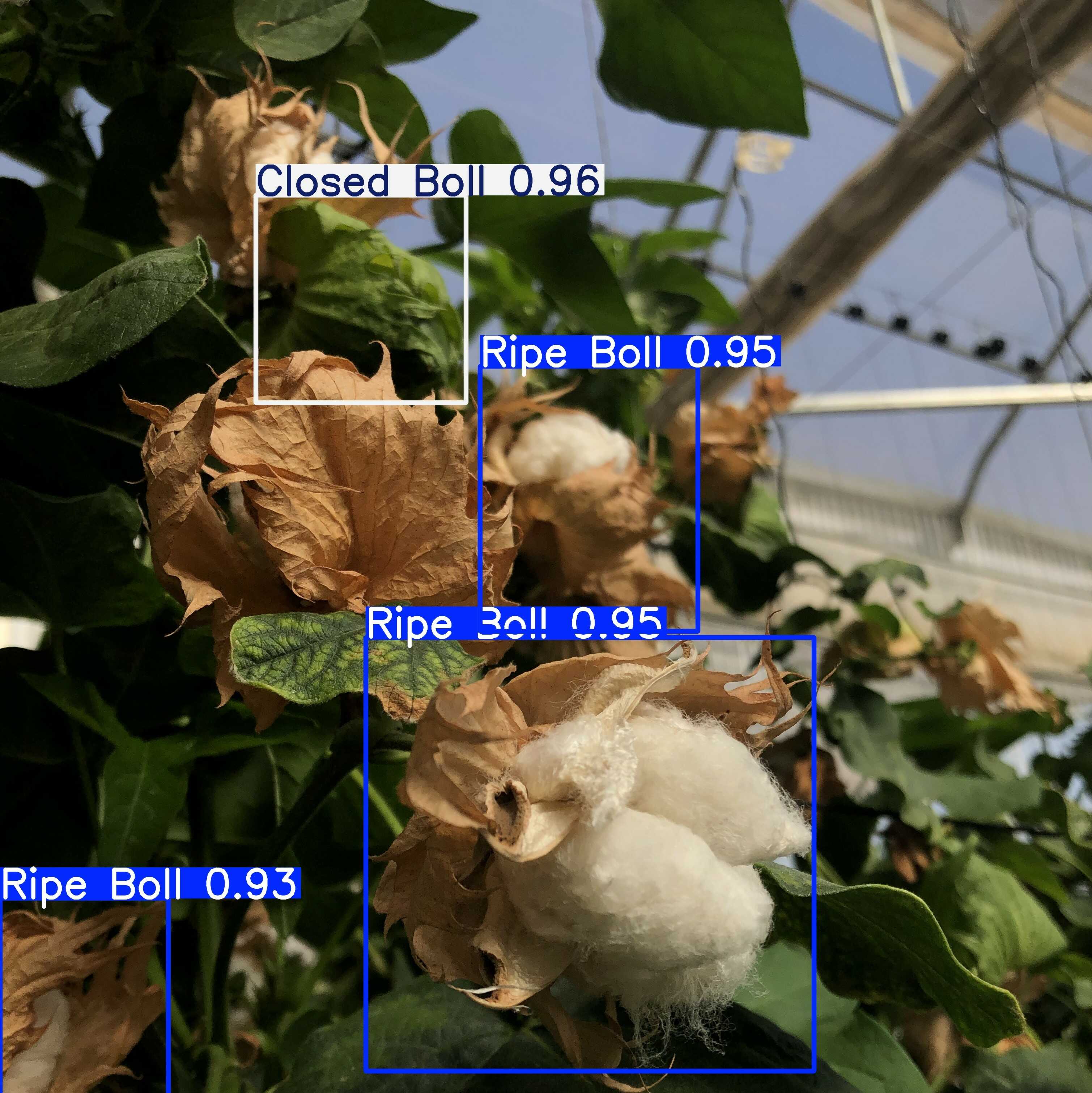}\end{subfigure}\interspacea%
\begin{subfigure}{\imgw}\includegraphics[width=\linewidth]{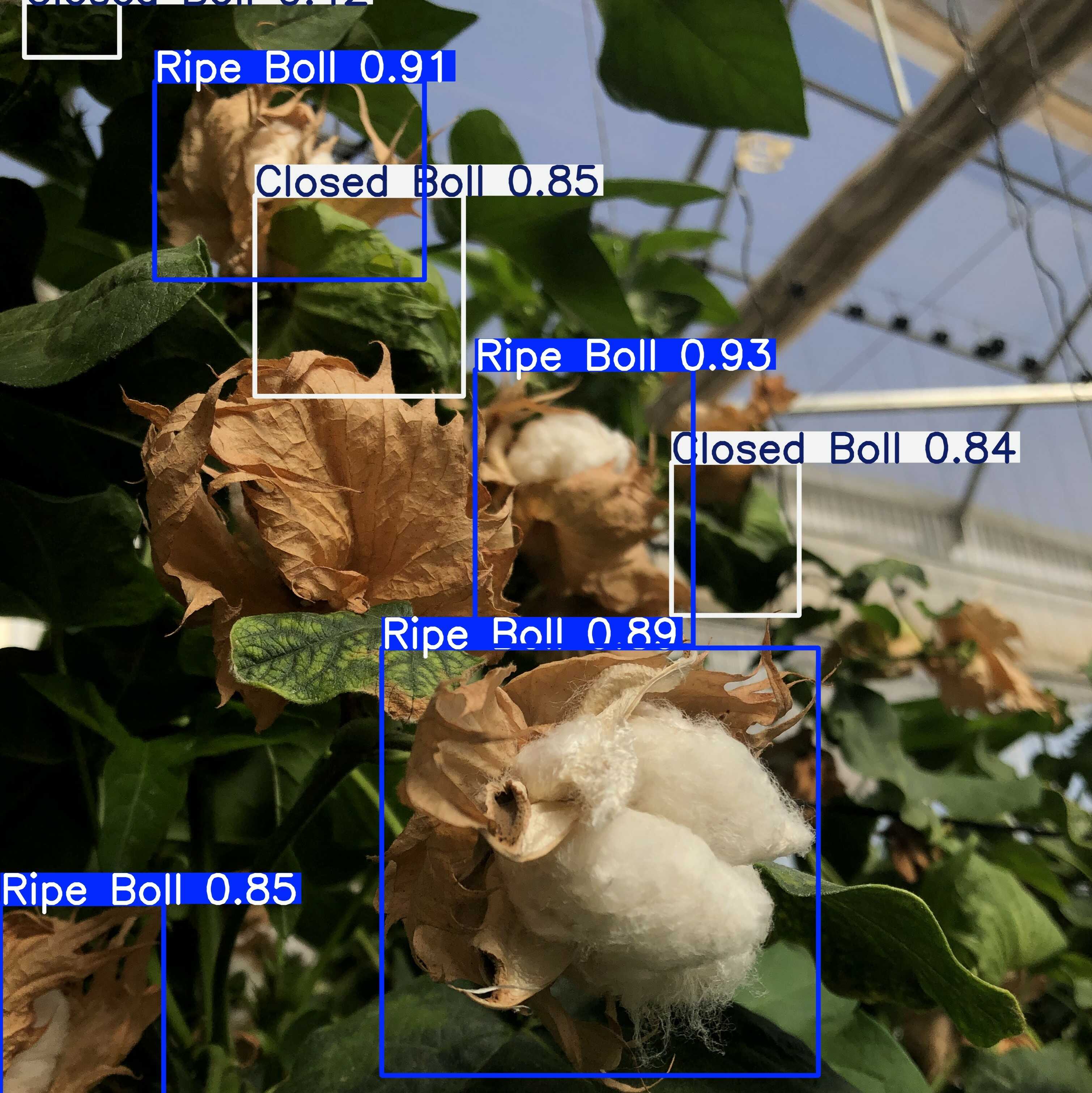}\end{subfigure}\interspacea%
\begin{subfigure}{\imgw}\includegraphics[width=\linewidth]{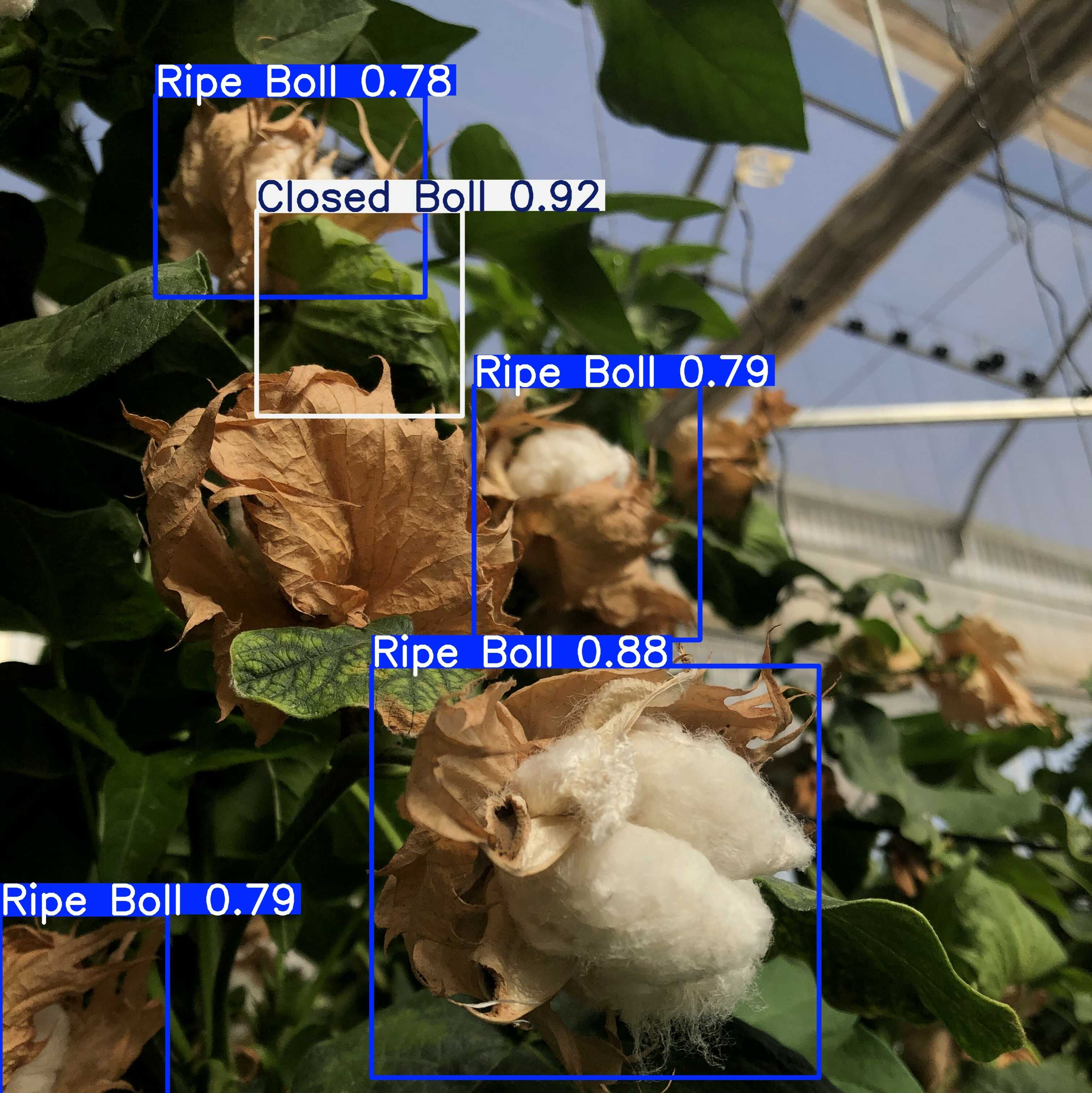}\end{subfigure}\interspacea%
\begin{subfigure}{\imgw}\includegraphics[width=\linewidth]{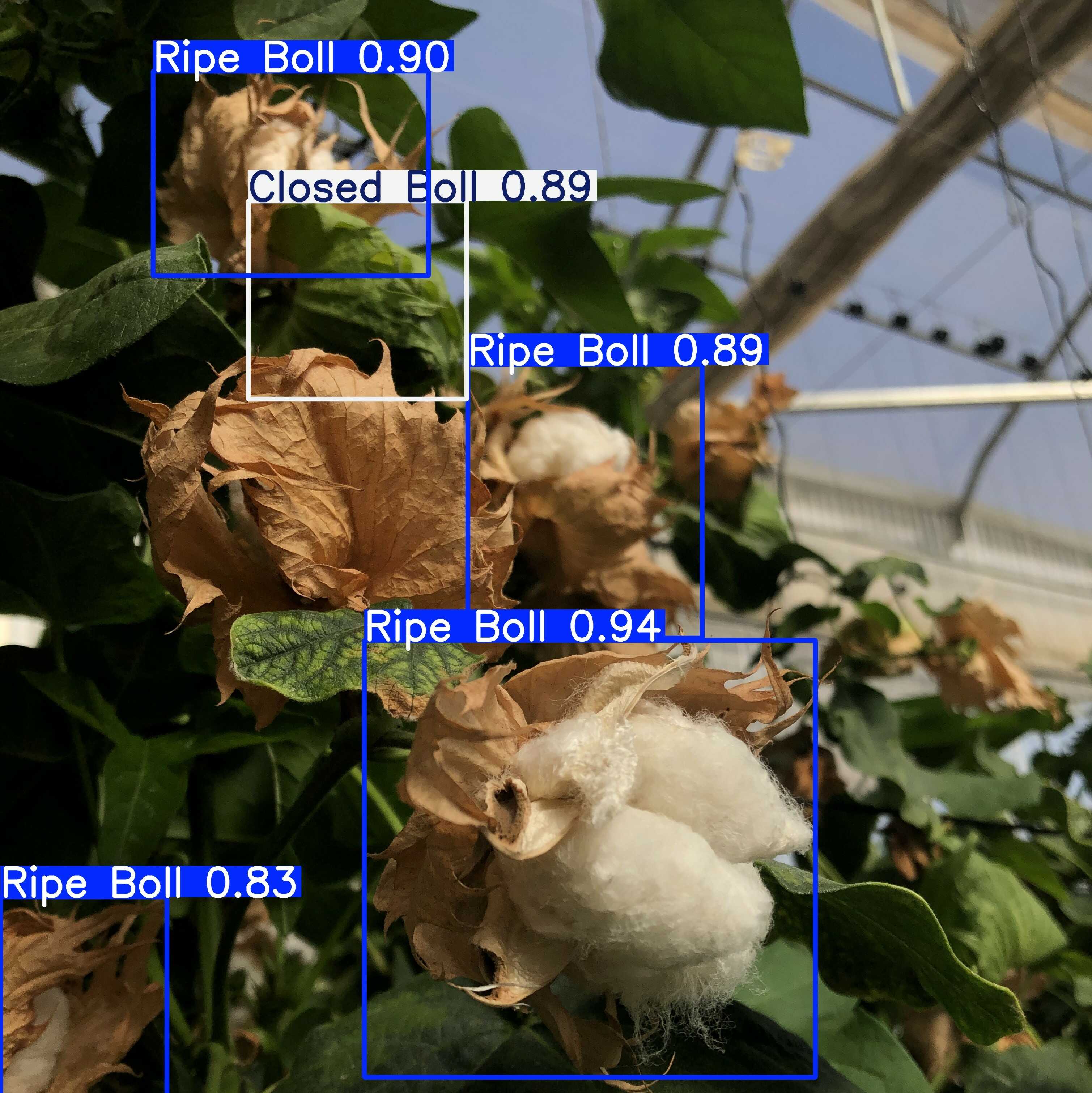}\end{subfigure}

\vspace{1mm}
\begin{subfigure}{\imgw}\includegraphics[width=\linewidth]{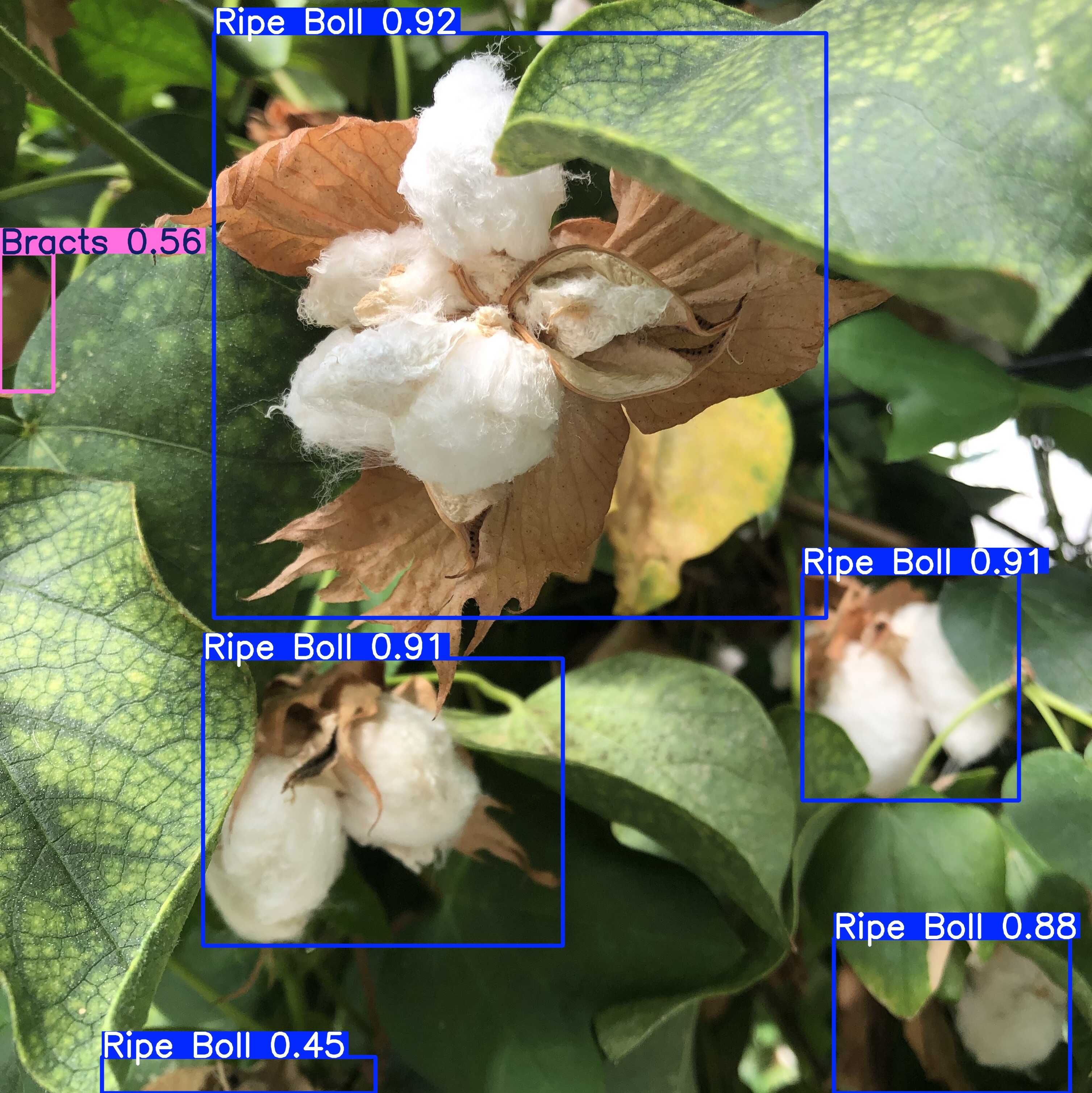}\end{subfigure}\interspacea%
\begin{subfigure}{\imgw}\includegraphics[width=\linewidth]{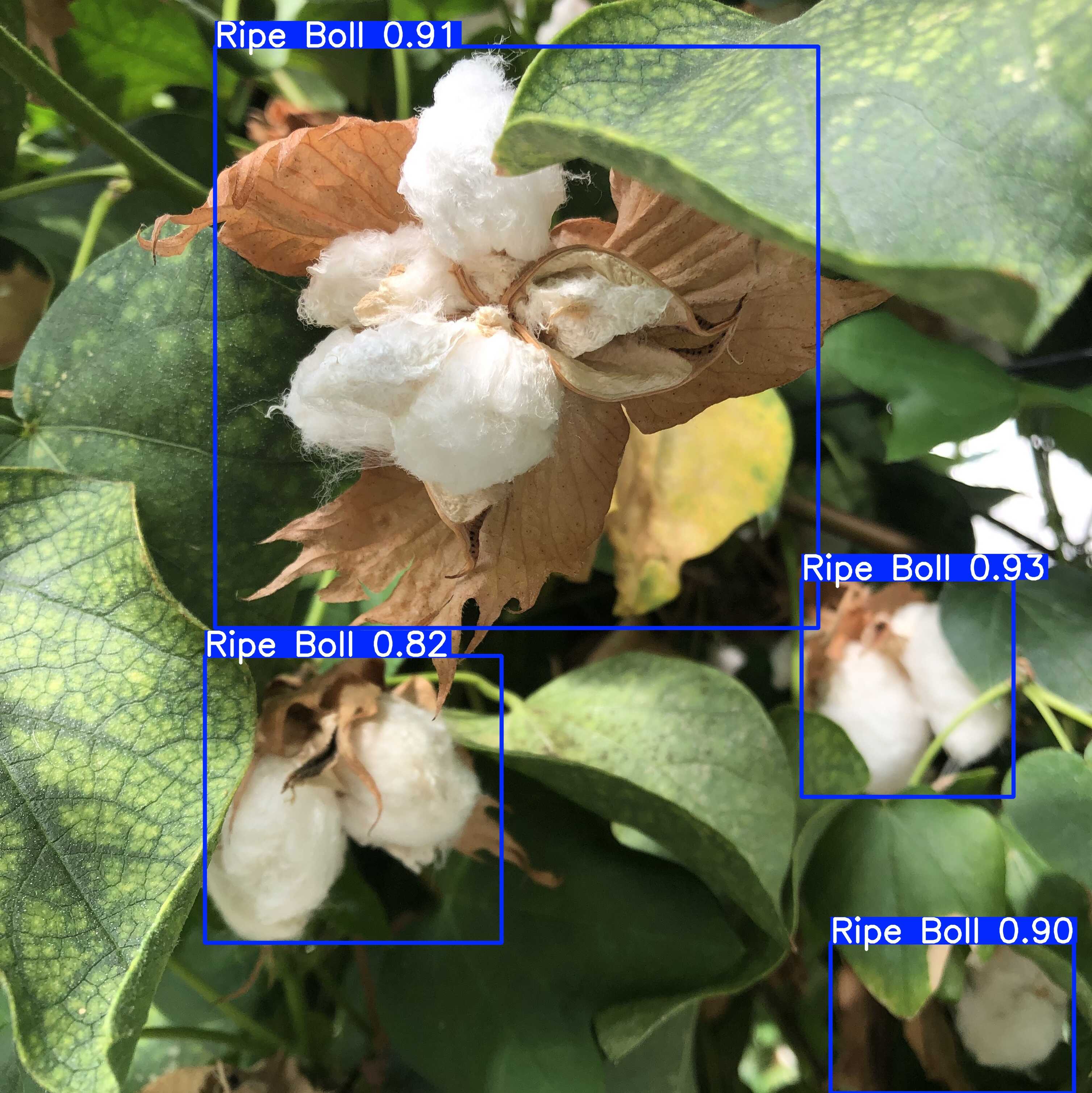}\end{subfigure}\interspacea%
\begin{subfigure}{\imgw}\includegraphics[width=\linewidth]{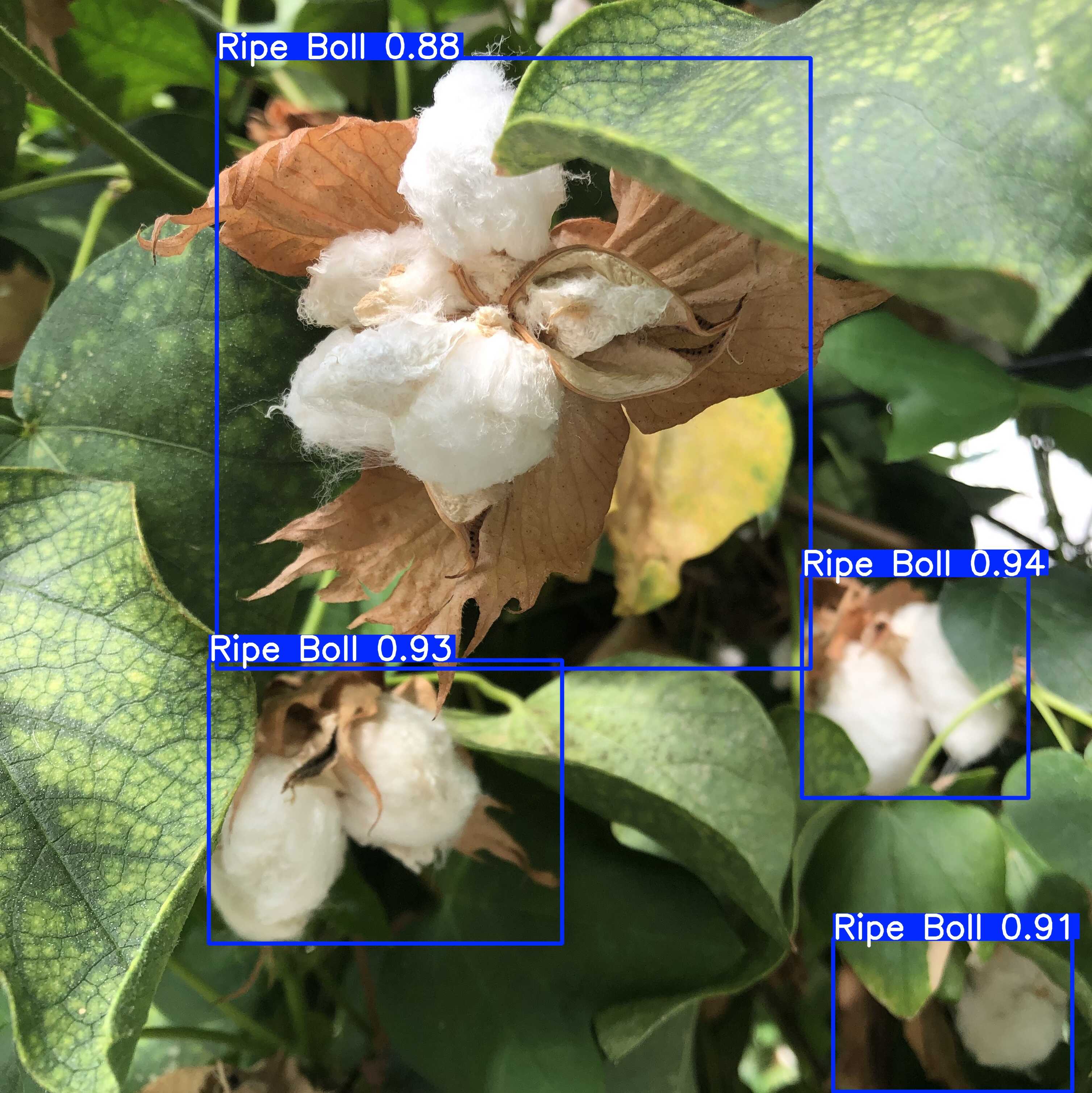}\end{subfigure}\interspacea%
\begin{subfigure}{\imgw}\includegraphics[width=\linewidth]{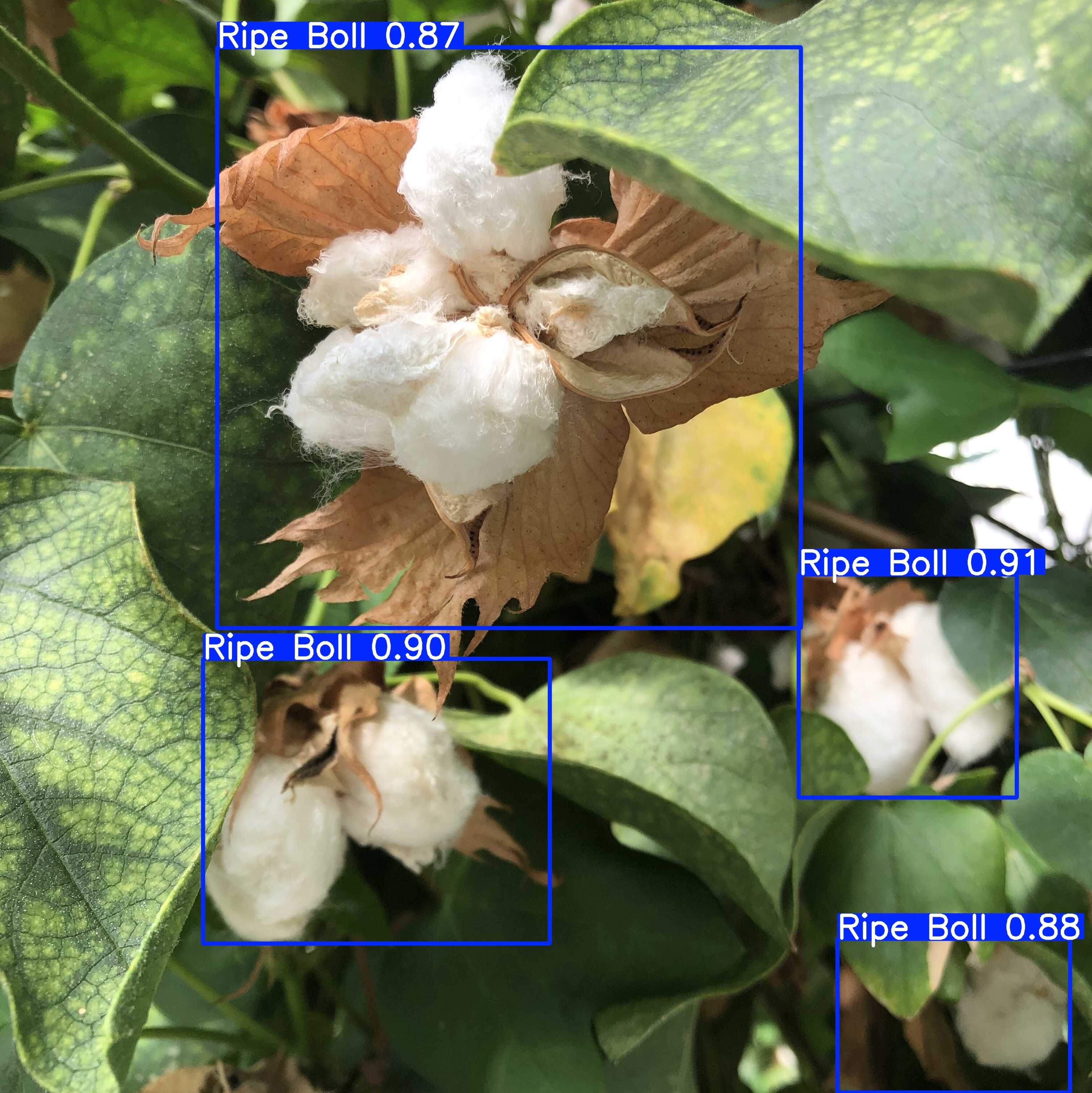}\end{subfigure}\interspacea%
\begin{subfigure}{\imgw}\includegraphics[width=\linewidth]{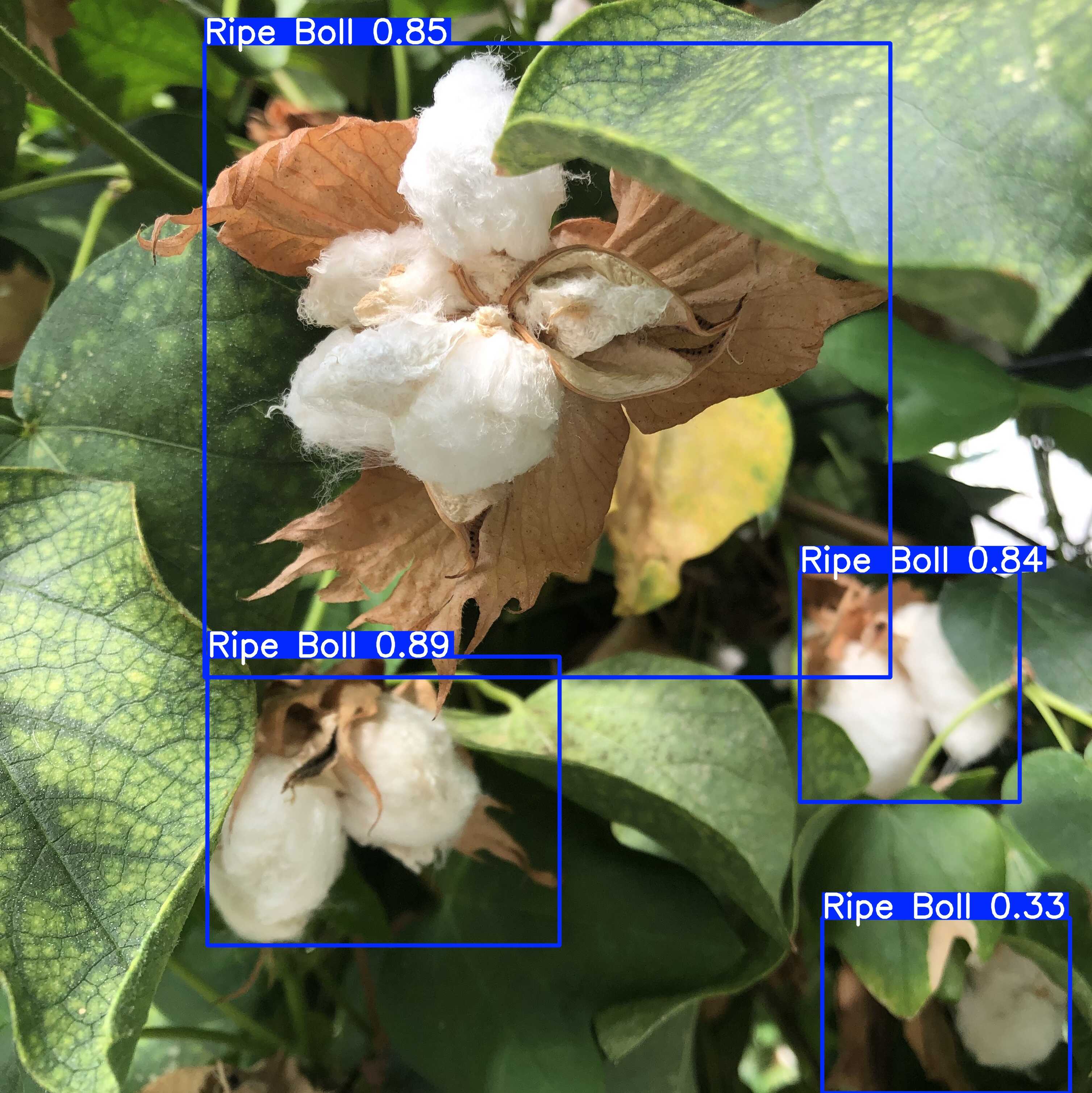}\end{subfigure}\interspacea%
\begin{subfigure}{\imgw}\includegraphics[width=\linewidth]{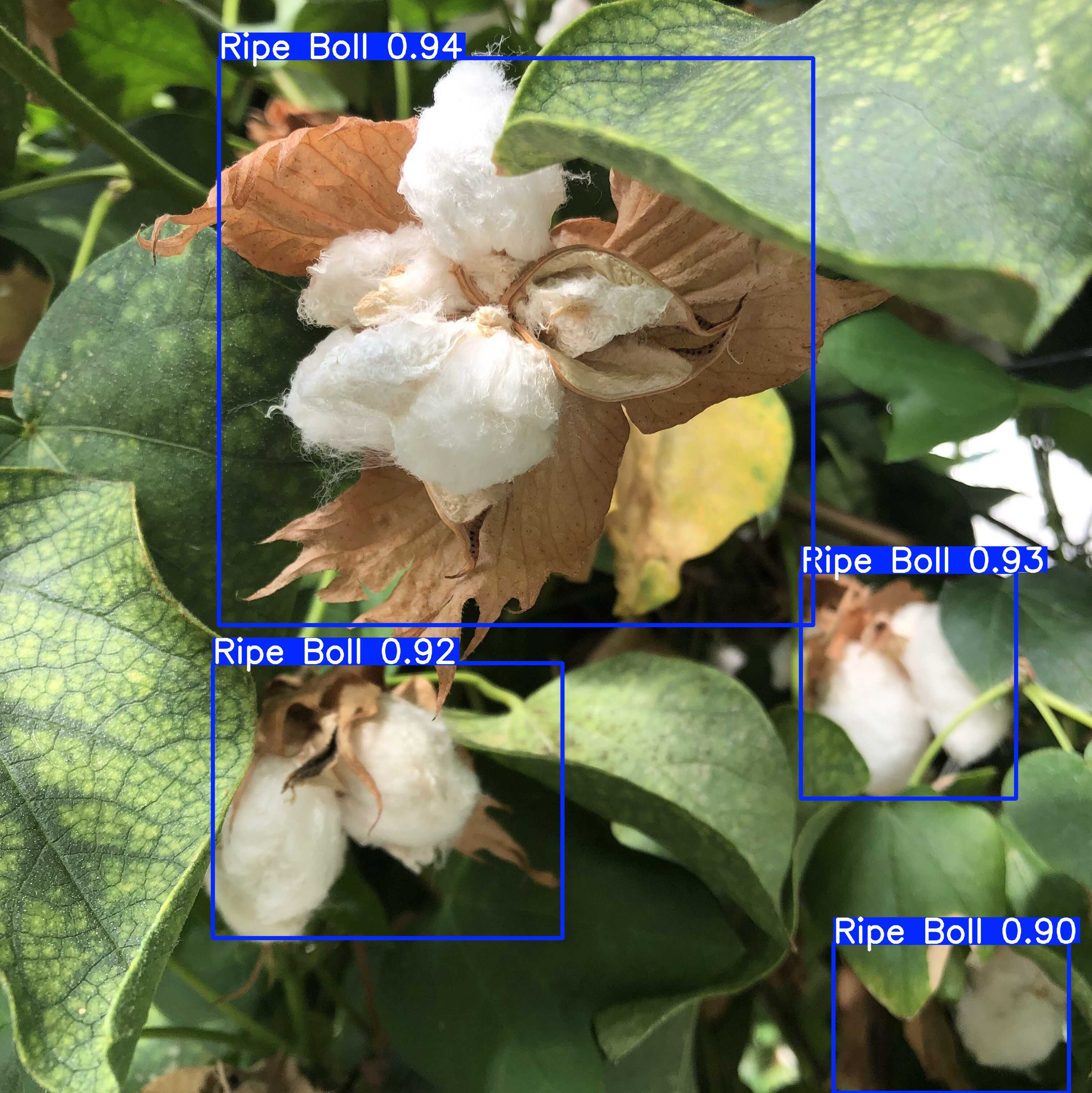}\end{subfigure}

\vspace{1mm}
\begin{subfigure}{\imgw}\includegraphics[width=\linewidth]{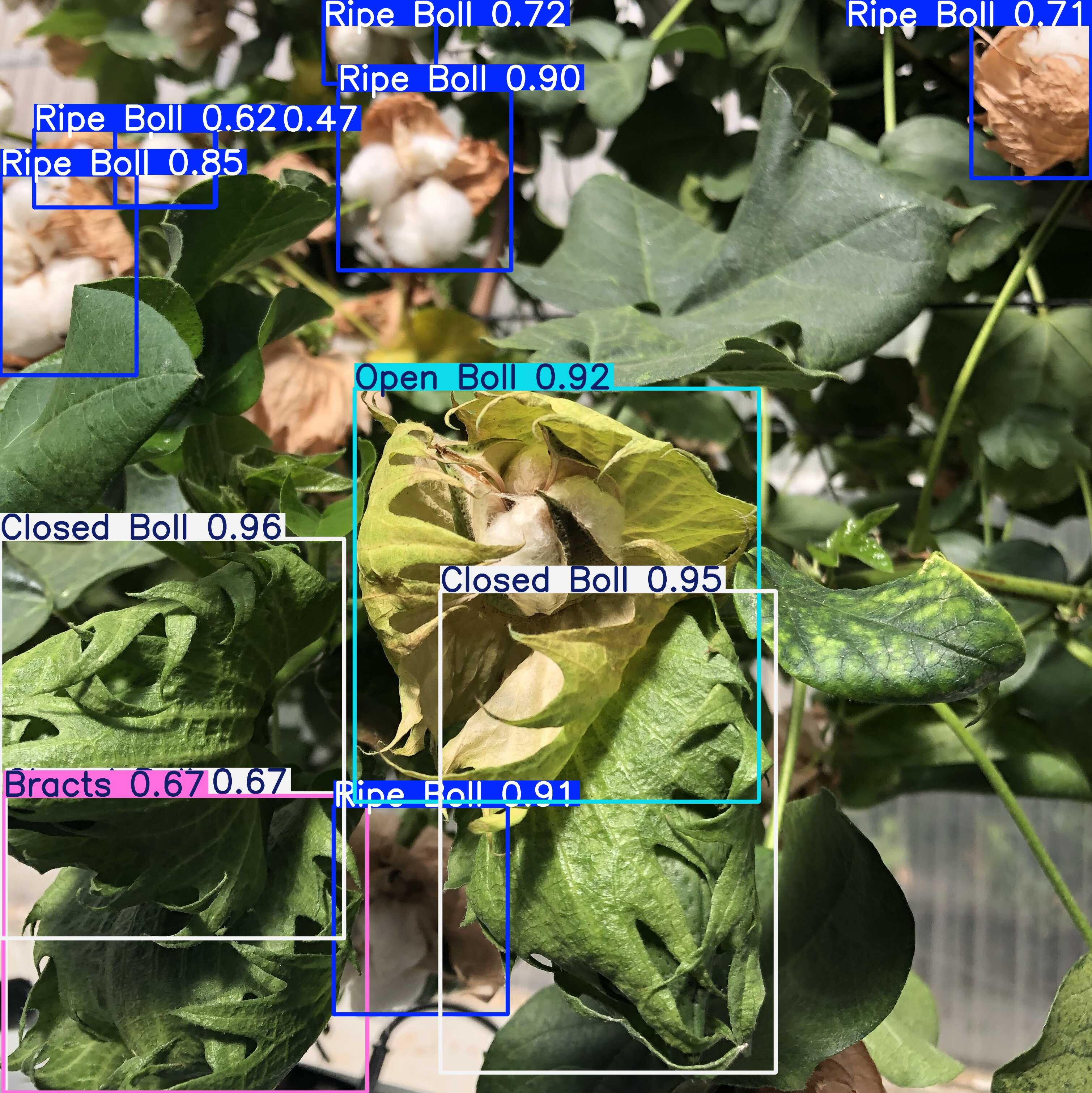}\end{subfigure}\interspacea%
\begin{subfigure}{\imgw}\includegraphics[width=\linewidth]{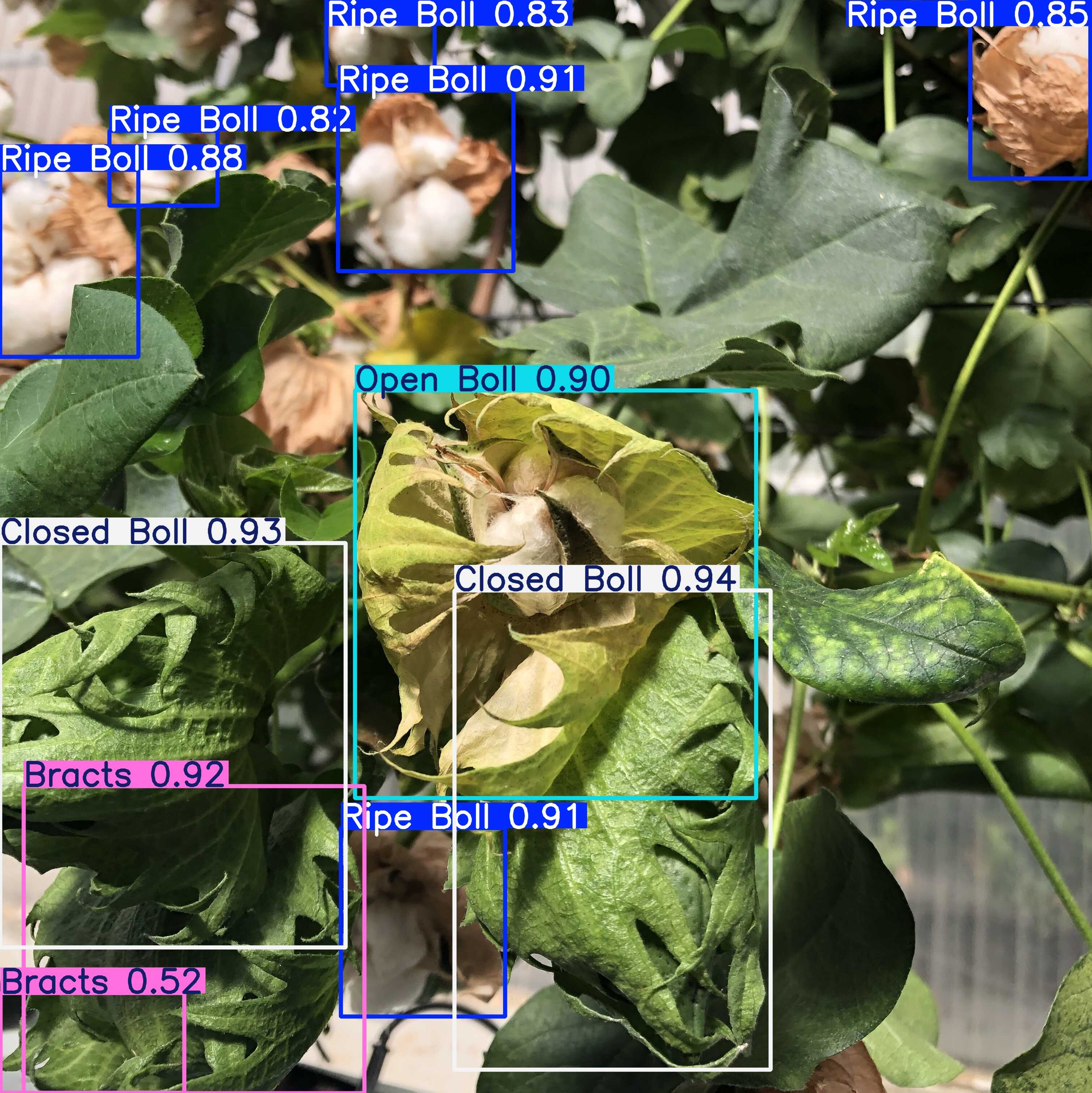}\end{subfigure}\interspacea%
\begin{subfigure}{\imgw}\includegraphics[width=\linewidth]{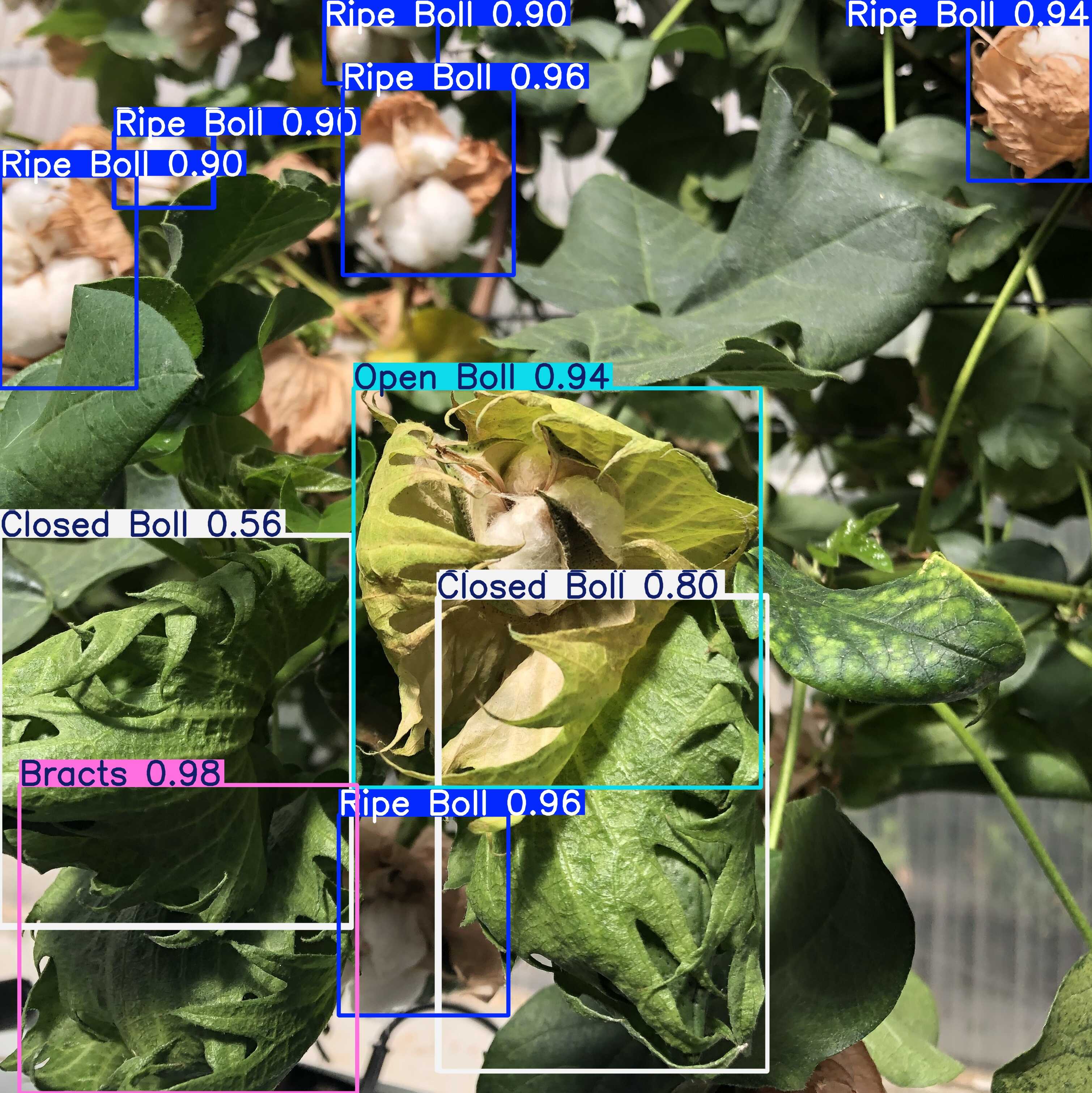}\end{subfigure}\interspacea%
\begin{subfigure}{\imgw}\includegraphics[width=\linewidth]{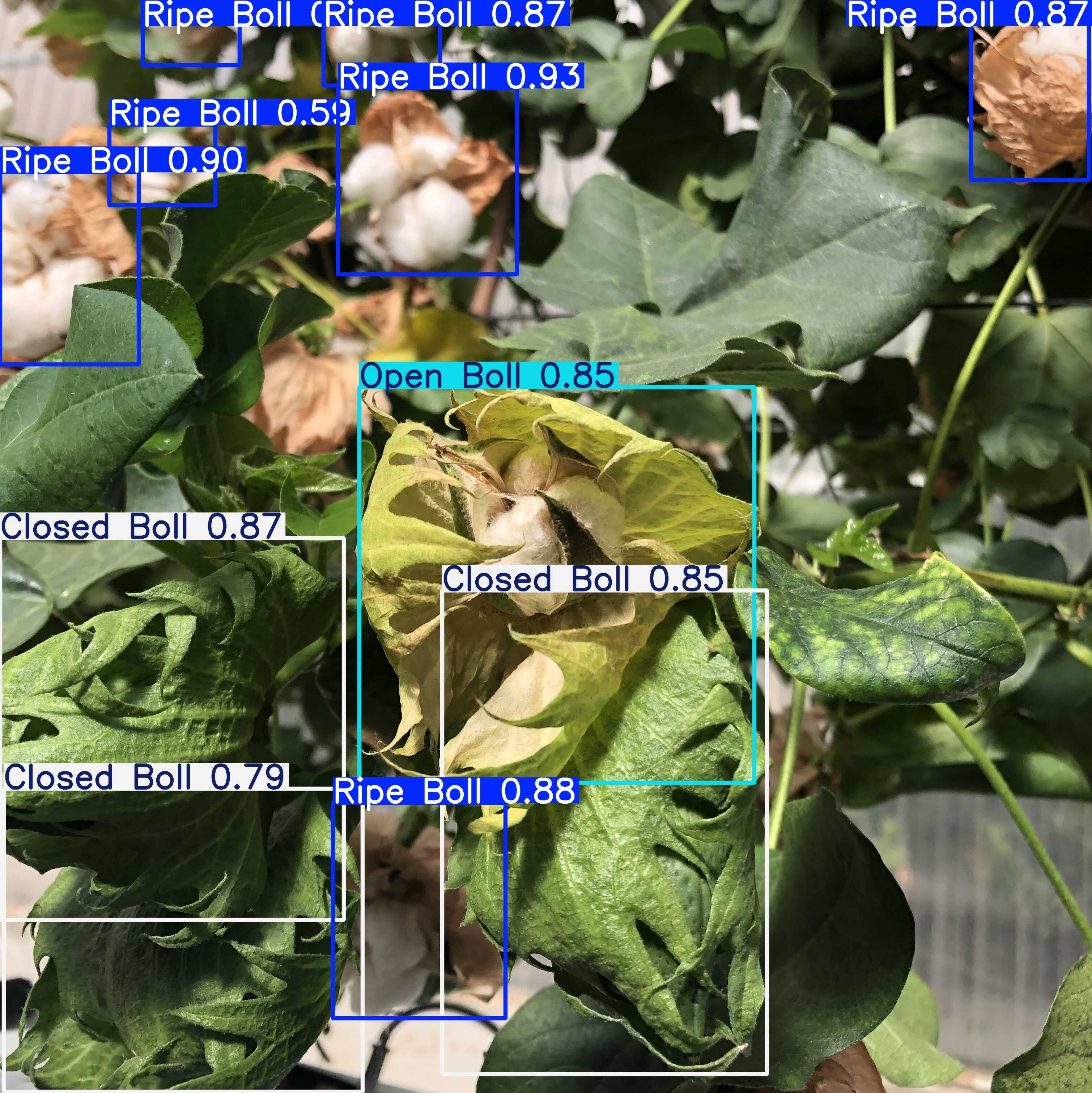}\end{subfigure}\interspacea%
\begin{subfigure}{\imgw}\includegraphics[width=\linewidth]{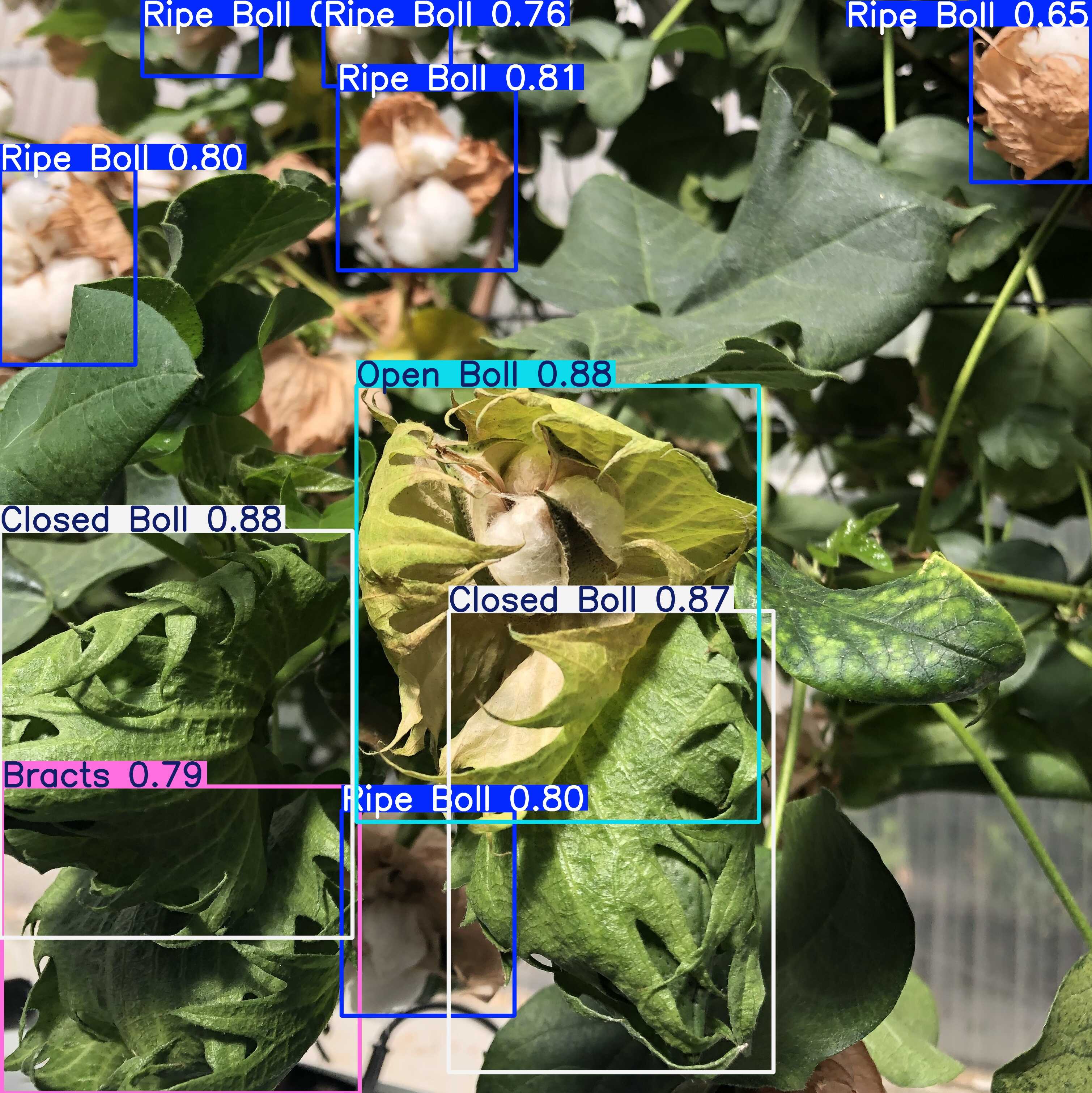}\end{subfigure}\interspacea%
\begin{subfigure}{\imgw}\includegraphics[width=\linewidth]{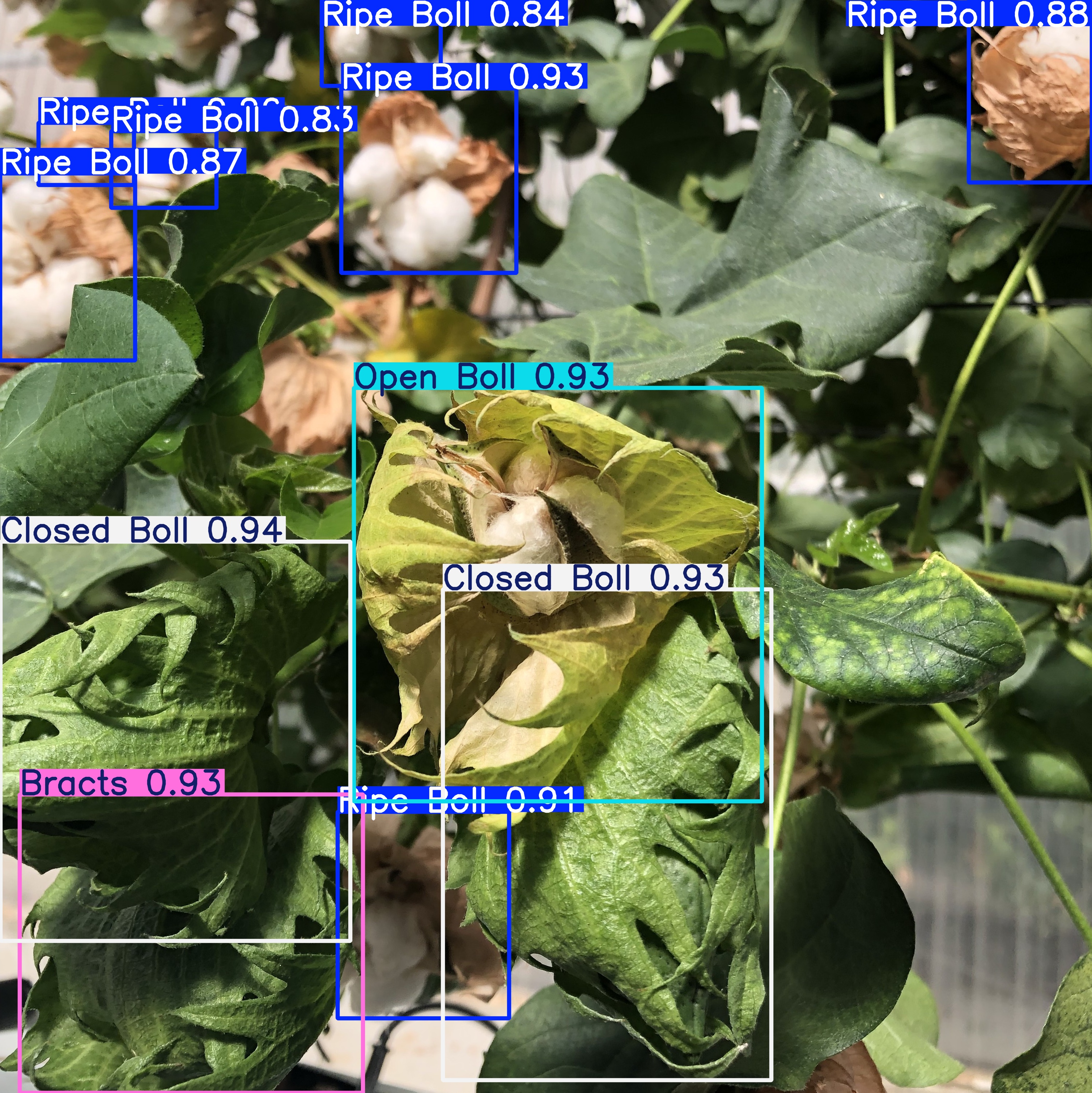}\end{subfigure}

\vspace{1mm}
\begin{subfigure}{\imgw}\includegraphics[width=\linewidth]{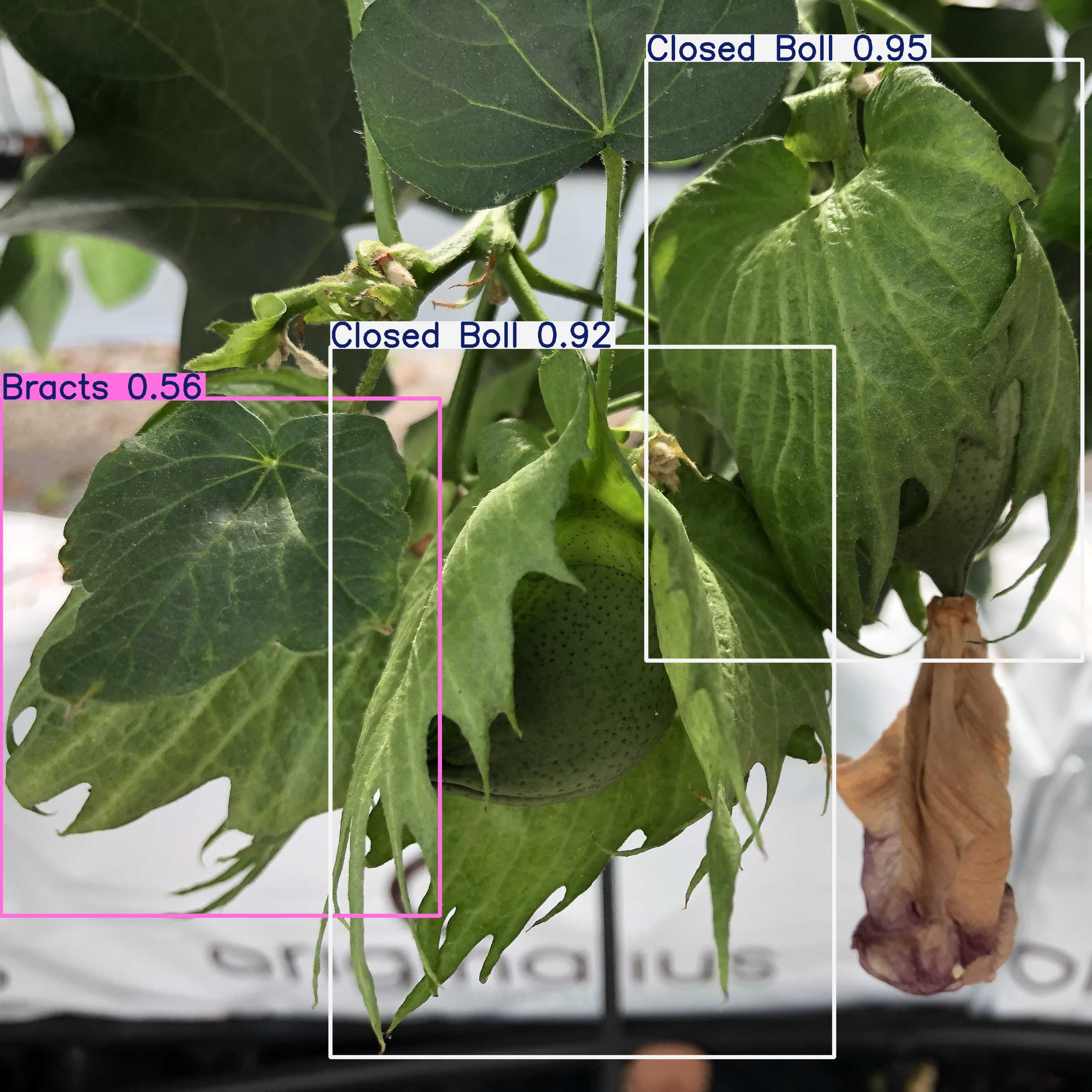}\end{subfigure}\interspacea%
\begin{subfigure}{\imgw}\includegraphics[width=\linewidth]{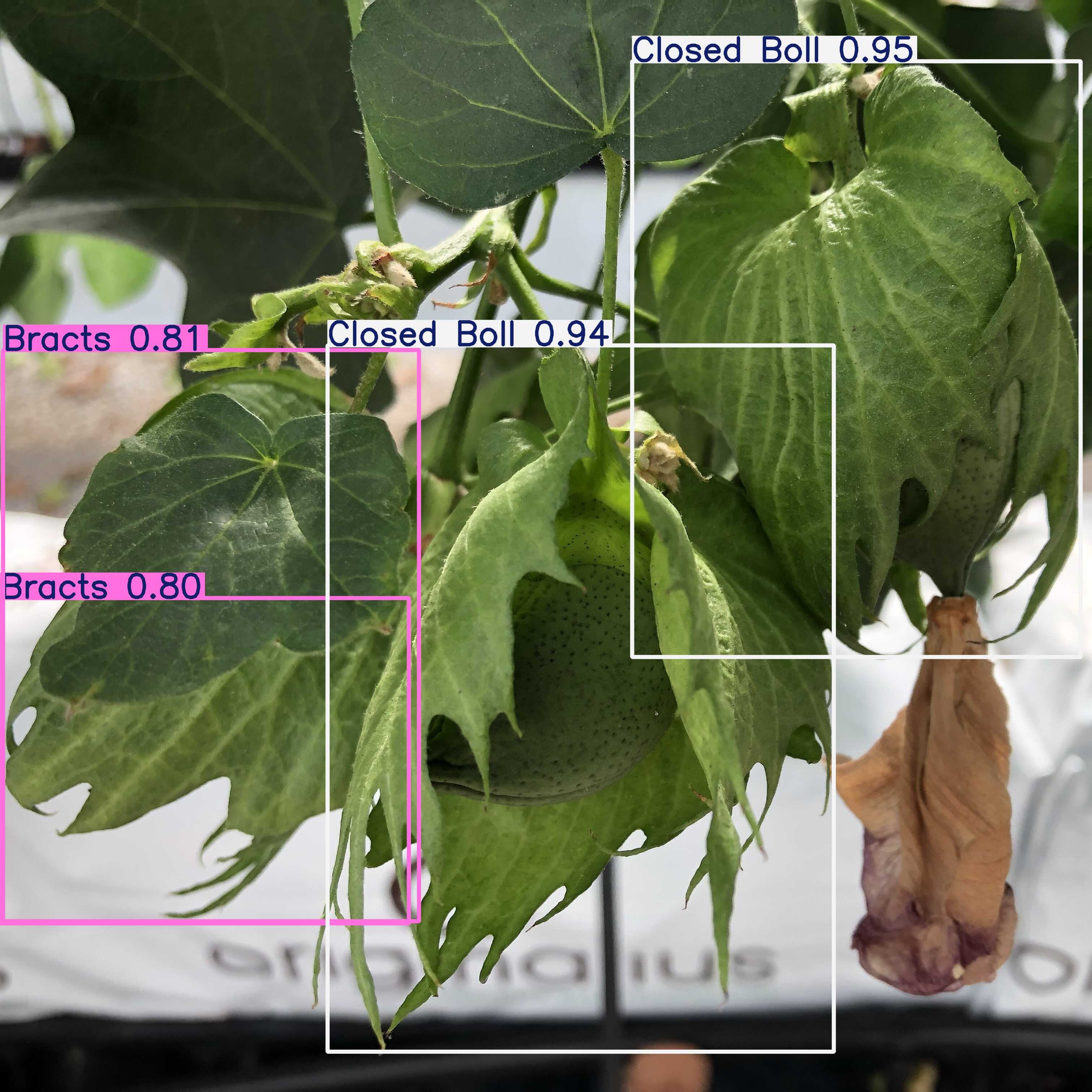}\end{subfigure}\interspacea%
\begin{subfigure}{\imgw}\includegraphics[width=\linewidth]{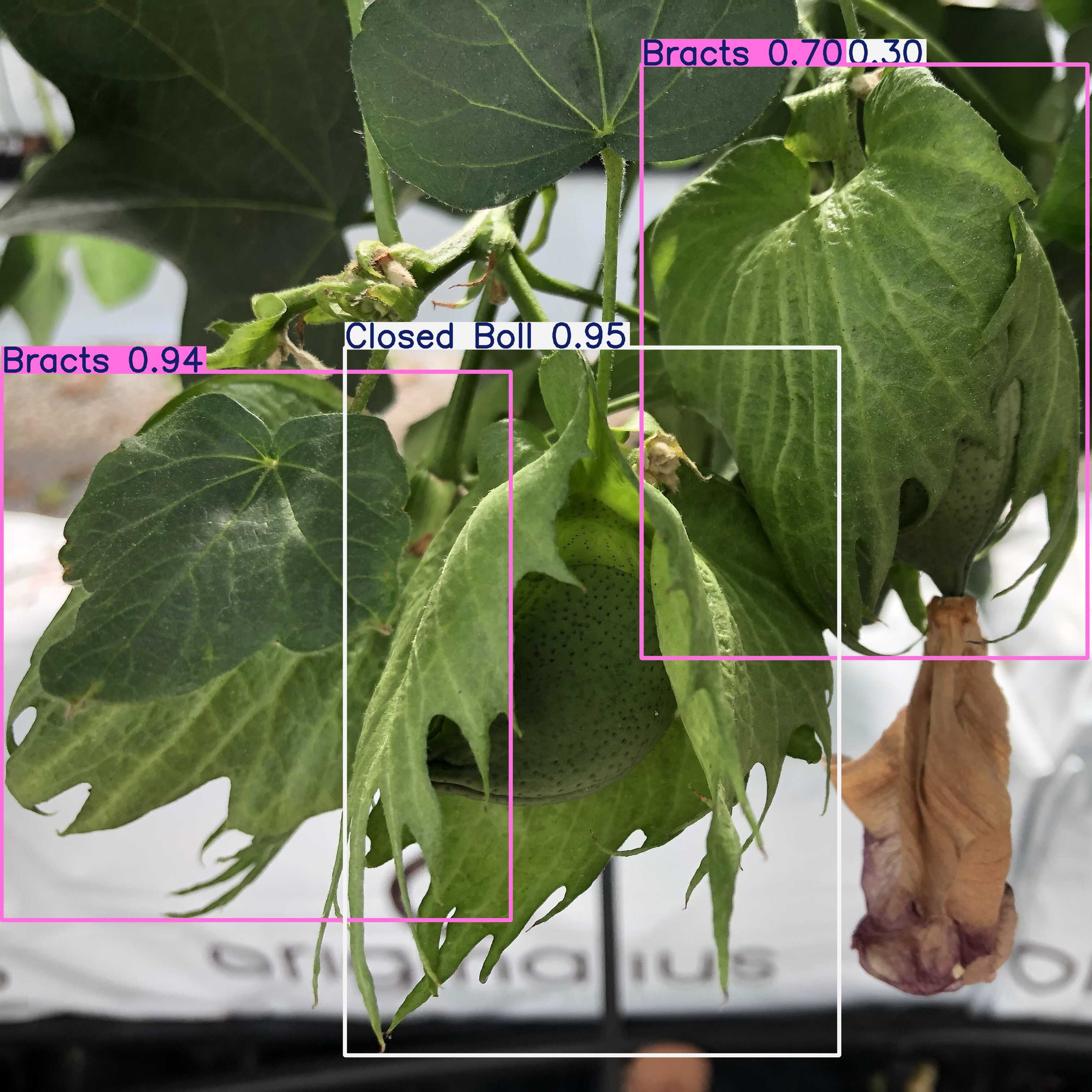}\end{subfigure}\interspacea%
\begin{subfigure}{\imgw}\includegraphics[width=\linewidth]{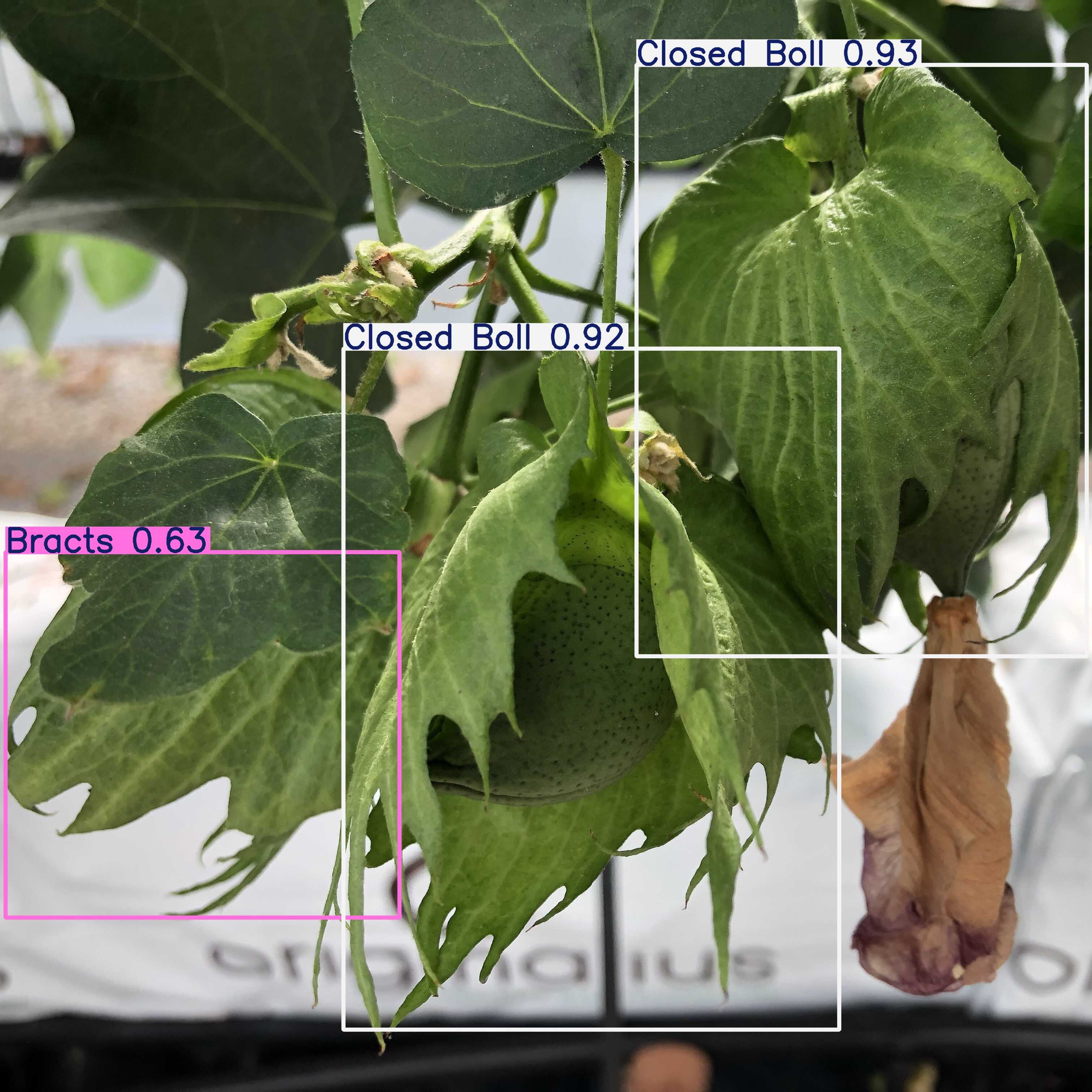}\end{subfigure}\interspacea%
\begin{subfigure}{\imgw}\includegraphics[width=\linewidth]{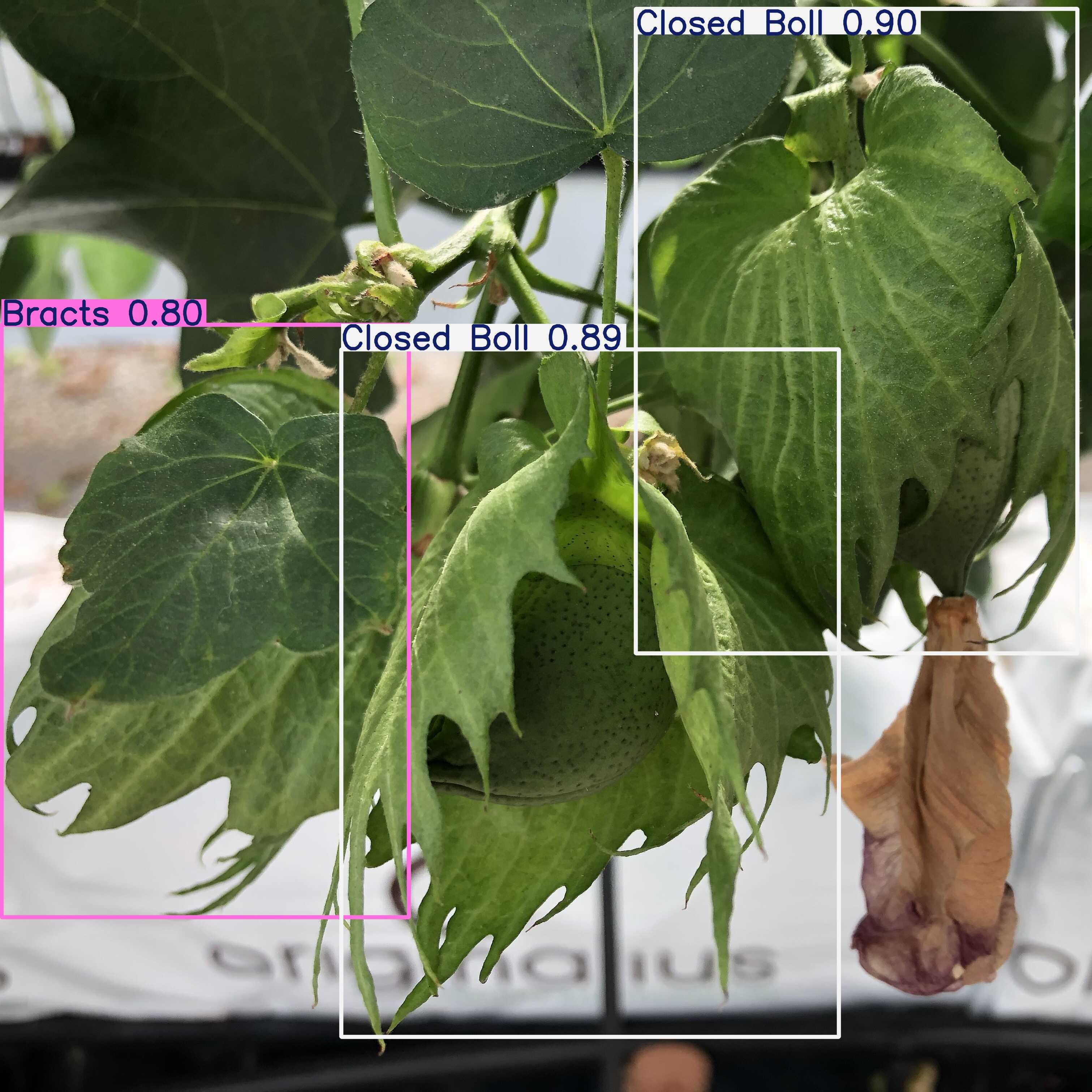}\end{subfigure}\interspacea%
\begin{subfigure}{\imgw}\includegraphics[width=\linewidth]{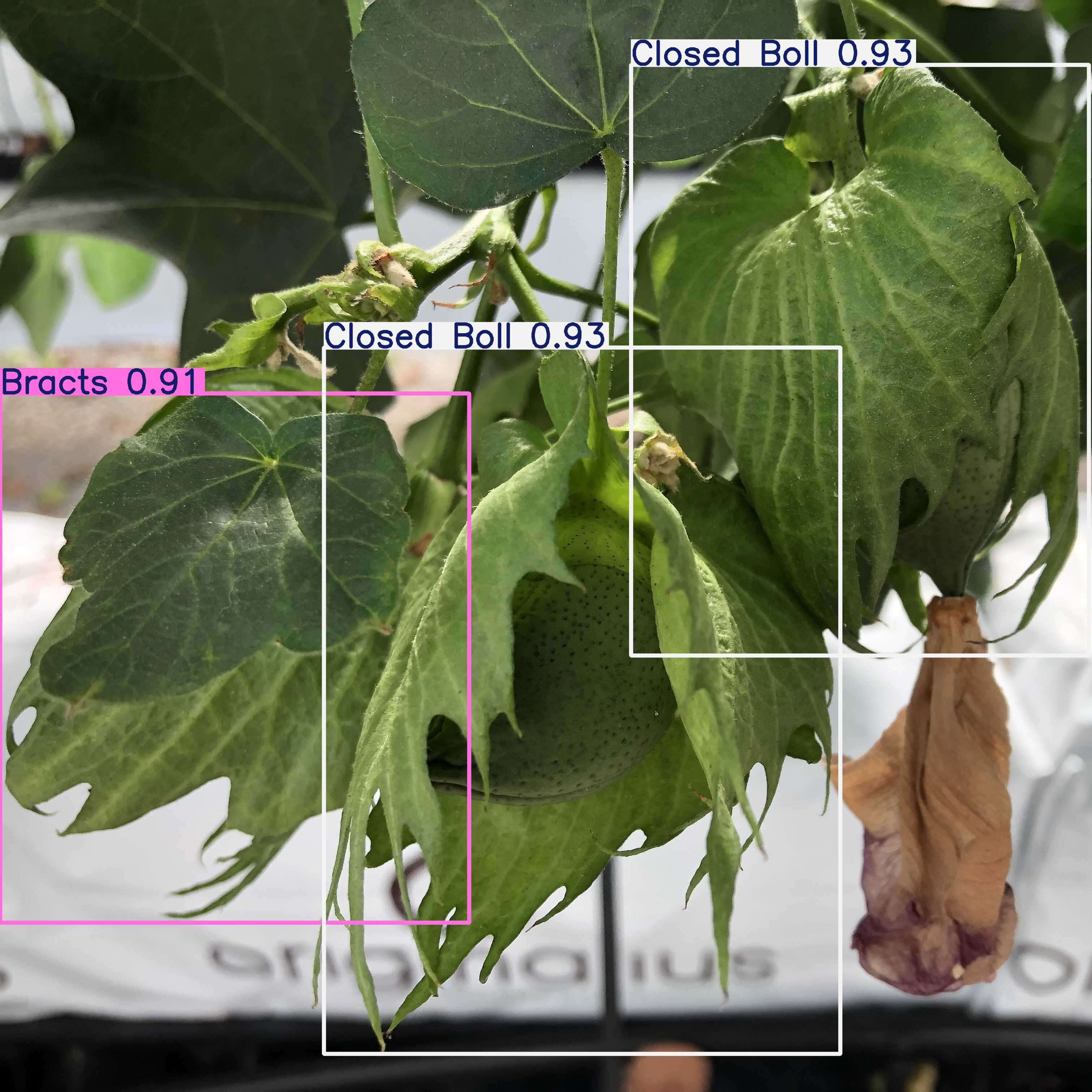}\end{subfigure}
\caption{Detection comparison between standard YOLO models and \modelName. Column-wise, left to right: YOLOv8m, YOLOv9m, YOLOv10m, YOLO11m, YOLOv12m, and \modelName predictions over different \dName classes.}
\label{fig:image_prediction_comparison}
\end{figure*}

The SIoU-based configuration emerges as the most effective, with high values for both precision (0.828) and recall (0.733), and the best results on mAP50 (0.811) and mAP50-95 (0.606). This behaviour suggests a more consistent and accurate localization across various IoU thresholds, demonstrating improved modelling of the geometry and spatial coherence of the boxes. In general, the findings indicate that SIoU offers the optimal balance between detection performance and localization reliability within the evaluated architecture.

\begin{table*}[t]
    \centering
    \begin{tabular}{lcccc}
        \hline
        \textbf{Bounding box loss function} & \textbf{Precision} & \textbf{Recall} & \textbf{mAP\textsubscript{50}} & \textbf{mAP\textsubscript{50:95}} \\
        \hline
            CIoU & 0.828 & 0.686 & 0.809 & 0.561 \\
            DIoU & 0.809 & 0.700 & 0.799 & 0.556 \\
            GIoU & 0.818 & 0.734 & \textbf{0.817} & 0.580 \\
            MDPIoU & 0.776 & \textbf{0.737} & 0.791 & 0.567 \\
            FocalerIoU & 0.783 & 0.733 & 0.809 & 0.586 \\
            % More?
        \hline
            SIoU & \textbf{0.828} & 0.733 & 0.811 & \textbf{0.606} \\
        \hline
    \end{tabular}
    \caption{Comparative analysis of different bounding box loss functions in \modelName.}
    \label{table:bb_loss_function_comparative}
\end{table*}

\subsection{Raw detection analysis}

To visually assist the results of the table~\ref{table:yolo_benchmarks}, we analyse a comparison between the best 5 models of the same table. Fig.~\ref{fig:image_prediction_comparison} shows the predictions for the same images for \modelName and the medium variant of YOLOv8, YOLOv9, YOLOv10, YOLO11 and YOLO12. These detections show better accuracy for all classes than the compared algorithms. 

In the first picture, YOLOv10m shows better accuracy for the predicted boxes, but lacks the ripe boll that \modelName detects. In the fourth picture, YOLO1v0m confuses classes for the right cotton boll. These examples suggest that YOLO10 has problems discerning between classes, specially for very similar classes, and sometimes produce predictions with a lower confidence than \modelName.

We can see some class confusions and hallucinations from the compared baseline YOLO. In the first picture, YOLO11 detects a closed ball when it is not clear that there is a boll. In the second picture, YOLOv8 detects some bracts. In the third picture, YOLOv8 has trouble discerning between bracts and closed cotton boll, YOLOv9 detects more bracts than there are in image (the same happens in picture 4), YOLO11 classifies wrongly the bracts as a boll and YOLO12 leaves ripe cottons in the background without detecting.

We can come up with the conclusion that \modelName outperforms the baseline YOLO models, classifies correctly every prediction and does not hallucinate on data in \dName. 

\section{Conclusions and future work} \label{CONC}

This study proposed several architectural modifications to the YOLO11 framework, integrating two pure attention blocks (SimAM and PHAM), a combined convolutional and lightweight attention module (SEConvBlock), and an advanced CARAFE operator for gradient-aware upsampling. Additionally, the ScyllaIoU (SIoU) loss function was implemented to create a more informed learning process. Experimental results demonstrate that this custom model achieves superior accuracy in both $mAP_{50}$ and $mAP_{50-95}$ compared to the baseline. Furthermore, the model exhibits improved feature localization in relevant image regions, thereby significantly enhancing cotton detection performance.

If we compare the results obtained with ~\citep{GONG2025101209}, we see that, based on an approach, \dName is well specified and phenotypically contextualised, distinguishing between different growth phases, as well as critical states of change in capsule development, such as flower pollination. We also cover important environmental variations, which adds robustness to the model. \modelName improves detection capability in subtle growth phases, and in conjunction with the metrics obtained, we see that it can generalise better in more complex tasks. Furthermore, ~\citep{GONG2025101209} does not provide information on accuracy for different IoU values (mAP50-95), so we do not know its accuracy in difficult detections. ~\citep{GONG2025101209} achieved better metrics for their custom model and dataset, but they demonstrate a lack of knowledge in their class conceptualisation, establishing classes that do not give insightful information (for example, the ``defected'' class). We think that our categorisation of cotton growth-stage classes provides a more comprehensive representation for later deployment in the field.

Finally, achieving a flexible model capable of adapting to the inherent complexity of cotton plants is essential for precise monitoring and maximizing harvest productivity. To achieve the goal of high-detail plantation analysis, future work will involve integrating the proposed object detection algorithm into a mobile robotic platform. This solution will feature a robotic arm equipped with a specialized gripper designed to harvest cotton fibres without compromising their physical properties. 

The integrated system will be responsible for autonomous plantation maintenance, monitoring the maturation of cotton bolls, and selectively harvesting ripe produce. To further refine performance, the dataset will be expanded to include a wider variety of growth stages, reducing false negatives, and ensuring that the robot can reliably discern between ripe and unripe bolls. Continual updates following the latest advances in object detection will remain a priority to ensure the highest possible precision in automated harvesting.

\section{Acknowledgments} \label{ACK}

The authors thank the Plant Nutrition Group at the Centro de Edafología y Biología Aplicada del Segura (CEBAS-CSIC), led by Professor Vicente Martínez, for providing the images used to construct the \dName dataset and for their expertise regarding the biological life cycles and physiology of cotton plants. In addition, we thank Mediterranean Agro Technologies S. L. (MagTech) for providing access to the crops and facilities used to conduct this research.

Funding: This work has been partially funded by the DEMETER 5.0 project PLEC2022-009289 funded by MCIU/AEI/10.13039/501100011033 and by the ”European Union NextGenerationEU/PRTR”; by project SDC007/25/000183 supported by "Departament d'Empresa i Treball, Generalitat de Catalunya" and "European Union NextGenerationEU/PRTR” and also by the European Union under the project ARISE (HORIZON-CL4-2023-DIGITAL-EMERGING-01-101135959).

\section{CRediT authorship contribution statement}

\textbf{Guillem González}: Conceptualization, Data curation, Formal Analysis, Investigation, Methodology, Software, Validation, Visualization, Writing - original draft, Writing - review \& editing. \textbf{Guillem Alenyà}: Funding acquisition, Writing - review \& editing. \textbf{Sergi Foix}: Funding acquisition, Project administration, Resources, Supervision, Writing - review \& editing.

\section{Declaration of competing interests} \label{DECL}

The authors declare that they have no financial, personal, or professional interests that could have influenced or influence the work reported in this paper.

\section{Supplementary data} \label{SUPL} 

Data will be made available on request.

\bibliographystyle{elsarticle-harv} 
\bibliography{refs}

@INPROCEEDINGS{gonzalez2024,
  author={González, Guillem and Martínez, Antonio and Martínez, Vicente and Foix, Sergi and Alenyà, Guillem},
  booktitle={2024 7th Iberian Robotics Conference (ROBOT)}, 
  title={Fast Ready-To-Harvest Cotton Detection and Classification with YOLOv8 in Greenhouse Crops}, 
  year={2024},
  volume={},
  number={},
  pages={1-6},
  doi={10.1109/ROBOT61475.2024.10796900}}

@article{gonzalez2025improved,
  title={An Improved YOLO11 Model for Accurate Cotton Growth-Stage Detection},
  author={González, Guillem and Foix, Sergi and Alenyà, Guillem},
  journal={Frontiers in Artificial Intelligence and Applications},
  pages={266},
  year={2024}
}

@article{omaye2024cross,
  title={Cross-comparative review of Machine learning for plant disease detection: Apple, cassava, cotton and potato plants},
  author={Omaye, James Daniel and Ogbuju, Emeka and Ataguba, Grace and Jaiyeoba, Oluwayemisi and Aneke, Joseph and Oladipo, Francisca},
  journal={Artificial intelligence in agriculture},
  volume={12},
  pages={127--151},
  year={2024},
  publisher={Elsevier}
}

@inproceedings{dewangan2023systematic,
  title={A systematic review on cotton plant disease detection \& classification using machine \& deep learning approach},
  author={Dewangan, Umashankar and Talwekar, RH and Bera, Swagota},
  booktitle={2023 1st DMIHER International Conference on Artificial Intelligence in Education and Industry 4.0 (IDICAIEI)},
  volume={1},
  pages={1--6},
  year={2023},
  organization={IEEE}
}

@INPROCEEDINGS{verma2024,
  title      = "Cotton growth stages detection using fine-tuned {YOLOv8} deep
                learning model",
  booktitle  = "2024 8th International Conference on Intelligent Systems,
                Metaheuristics \& Swarm Intelligence ({ISMSI})",
  author     = "Verma, Pooja and Paul, Ayan and Machavaram, Rajendra and
                Bhattacharya, Mahua",
  publisher  = "ACM",
  month      =  apr,
  year       =  2024,
  address    = "New York, NY, USA",
  conference = "ISMSI 2024: 2024 8th International Conference on Intelligent
                Systems, Metaheuristics \& Swarm Intelligence",
  location   = "Singapore Singapore"
}

@article{Li2016cottonsegmentation,
title = {In-field cotton detection via region-based semantic image segmentation},
journal = {Computers and Electronics in Agriculture},
volume = {127},
pages = {475-486},
year = {2016},
issn = {0168-1699},
doi = {https://doi.org/10.1016/j.compag.2016.07.006},
url = {https://www.sciencedirect.com/science/article/pii/S016816991630480X},
author = {Yanan Li and Zhiguo Cao and Hao Lu and Yang Xiao and Yanjun Zhu and Armin B. Cremers}
}

@article{Li2017cottonsegmentation,
author = {Yanan Li and Zhiguo Cao and Yang Xiao and Armin B. Cremers},
title = {{DeepCotton: in-field cotton segmentation using deep fully convolutional network}},
volume = {26},
journal = {Journal of Electronic Imaging},
number = {5},
publisher = {SPIE},
pages = {053028},
year = {2017},
doi = {10.1117/1.JEI.26.5.053028},
url = {https://doi.org/10.1117/1.JEI.26.5.053028}
}

@article{Li2025pineappleYOLO,
title = {A lightweight pineapple detection network based on YOLOv7-tiny for agricultural robot system},
journal = {Computers and Electronics in Agriculture},
volume = {231},
pages = {109944},
year = {2025},
issn = {0168-1699},
doi = {https://doi.org/10.1016/j.compag.2025.109944},
url = {https://www.sciencedirect.com/science/article/pii/S016816992500050X},
author = {Jiehao Li and Chenglin Li and Shan Zeng and Xiwen Luo and C.L. Philip Chen and Chenguang Yang}
}

@article{liu2024fasteryoloapapples,
	title = {Faster-{YOLO}-{AP}: {A} lightweight apple detection algorithm based on improved {YOLOv8} with a new efficient {PDWConv} in orchard},
	volume = {223},
	issn = {0168-1699},
	url = {https://www.sciencedirect.com/science/article/pii/S016816992400509X},
	doi = {https://doi.org/10.1016/j.compag.2024.109118},
	journal = {Computers and Electronics in Agriculture},
	author = {Liu, Zifu and Abeyrathna, R. M. Rasika D. and Sampurno, Rizky Mulya and Nakaguchi, Victor Massaki and Ahamed, Tofael},
	year = {2024},
	pages = {109118},
}

@article{wang2025insectYOLO,
title = {Insect-YOLO: A new method of crop insect detection},
journal = {Computers and Electronics in Agriculture},
volume = {232},
pages = {110085},
year = {2025},
issn = {0168-1699},
doi = {https://doi.org/10.1016/j.compag.2025.110085},
url = {https://www.sciencedirect.com/science/article/pii/S0168169925001917},
author = {Nan Wang and Shaowen Fu and Qiong Rao and Guiyou Zhang and Mingquan Ding}
}

@article{jin2025cameliaolifera,
title = {CO-YOLO: A lightweight and efficient model for Camellia oleifera fruit object detection and posture determination},
journal = {Computers and Electronics in Agriculture},
volume = {235},
pages = {110394},
year = {2025},
issn = {0168-1699},
doi = {https://doi.org/10.1016/j.compag.2025.110394},
url = {https://www.sciencedirect.com/science/article/pii/S0168169925005009},
author = {Shouxiang Jin and Lei Zhou and Hongping Zhou}
}

@article{GONG2025101209,
title = {Improved YOLOv10n model for enhanced cotton recognition in complex environments},
journal = {Smart Agricultural Technology},
volume = {12},
pages = {101209},
year = {2025},
issn = {2772-3755},
doi = {https://doi.org/10.1016/j.atech.2025.101209},
url = {https://www.sciencedirect.com/science/article/pii/S277237552500440X},
author = {Yutao Gong and Wenwen Ding and Nenghui Huang and Tao Li and Juntao Zhou}
}

@Article{wang2024grapedetection,
AUTHOR = {Wang, Wenhao and Shi, Yun and Liu, Wanfu and Che, Zijin},
TITLE = {An Unstructured Orchard Grape Detection Method Utilizing YOLOv5s},
JOURNAL = {Agriculture},
VOLUME = {14},
YEAR = {2024},
NUMBER = {2},
ARTICLE-NUMBER = {262},
url = {https://www.mdpi.com/2077-0472/14/2/262},
ISSN = {2077-0472},
doi = {10.3390/agriculture14020262}
}

@article{deng2025tomato,
title = {SE-YOLO: A sobel-enhanced framework for high-accuracy, lightweight real-time tomato detection with edge deployment capability},
journal = {Computers and Electronics in Agriculture},
volume = {239},
pages = {110973},
year = {2025},
issn = {0168-1699},
doi = {https://doi.org/10.1016/j.compag.2025.110973},
url = {https://www.sciencedirect.com/science/article/pii/S0168169925010798},
author = {Xiao Deng and Tianlun Huang and Weijun Wang and Wei Feng},
}

@article{zhai2024greentomato,
title = {Green fruit detection methods: Innovative application of camouflage object detection and multilevel feature mining},
journal = {Computers and Electronics in Agriculture},
volume = {225},
pages = {109356},
year = {2024},
issn = {0168-1699},
doi = {https://doi.org/10.1016/j.compag.2024.109356},
url = {https://www.sciencedirect.com/science/article/pii/S0168169924007476},
author = {Yuting Zhai and Zongmei Gao and Yang Zhou and Jian Li and Yuqi Zhang and Yanlei Xu}
}

@inproceedings{wang2019carafecontentawarereassemblyfeatures,
  title={Carafe: Content-aware reassembly of features},
  author={Wang, Jiaqi and Chen, Kai and Xu, Rui and Liu, Ziwei and Loy, Chen Change and Lin, Dahua},
  booktitle={Proceedings of the IEEE/CVF international conference on computer vision},
  pages={3007--3016},
  year={2019}
}

@inproceedings{ouyang2023EMAblock,
   title={Efficient Multi-Scale Attention Module with Cross-Spatial Learning},
   url={http://dx.doi.org/10.1109/ICASSP49357.2023.10096516},
   doi={10.1109/icassp49357.2023.10096516},
   booktitle={ICASSP 2023 - 2023 IEEE International Conference on Acoustics, Speech and Signal Processing (ICASSP)},
   publisher={IEEE},
   author={Ouyang, Daliang and He, Su and Zhang, Guozhong and Luo, Mingzhu and Guo, Huaiyong and Zhan, Jian and Huang, Zhijie},
   year={2023},
   month=jun, pages={1–5} }

@inproceedings{woo2018CBAMblock,
  title={Cbam: Convolutional block attention module},
  author={Woo, Sanghyun and Park, Jongchan and Lee, Joon-Young and Kweon, In So},
  booktitle={Proceedings of the European conference on computer vision (ECCV)},
  pages={3--19},
  year={2018}
}

@inproceedings{hu2019squeezeandexcitationnetworks,
  title={Squeeze-and-excitation networks},
  author={Hu, Jie and Shen, Li and Sun, Gang},
  booktitle={Proceedings of the IEEE conference on computer vision and pattern recognition},
  pages={7132--7141},
  year={2018}
}

@InProceedings{pmlr-v139-yang21o,
    title = 	 {SimAM: A Simple, Parameter-Free Attention Module for Convolutional Neural Networks},
    author =       {Yang, Lingxiao and Zhang, Ru-Yuan and Li, Lida and Xie, Xiaohua},
    booktitle = 	 {Proceedings of the 38th International Conference on Machine Learning},
    pages = 	 {11863--11874},
    year = 	 {2021},
    editor = 	 {Meila, Marina and Zhang, Tong},
    volume = 	 {139},
    series = 	 {Proceedings of Machine Learning Research},
    month = 	 {18--24 Jul},
    publisher =    {PMLR},
    url = 	 {http://proceedings.mlr.press/v139/yang21o.html}
}

@article{pham_block,
author = {Dong, Hao and Yuan, Mu and Wang, Shu and Zhang, Long and Bao, Wenxia and Liu, Yong and Hu, Qingyuan},
year = {2023},
month = {06},
pages = {},
title = {PHAM-YOLO: A Parallel Hybrid Attention Mechanism Network for Defect Detection of Meter in Substation},
volume = {23},
journal = {Sensors},
doi = {10.3390/s23136052}
}

@article{mca,
title = {MCA: Multidimensional collaborative attention in deep convolutional neural networks for image recognition},
journal = {Engineering Applications of Artificial Intelligence},
volume = {126},
pages = {107079},
year = {2023},
issn = {0952-1976},
doi = {https://doi.org/10.1016/j.engappai.2023.107079},
url = {https://www.sciencedirect.com/science/article/pii/S0952197623012630},
author = {Yang Yu and Yi Zhang and Zeyu Cheng and Zhe Song and Chengkai Tang}
}

@INPROCEEDINGS{tripletattention,
  author={Misra, Diganta and Nalamada, Trikay and Arasanipalai, Ajay Uppili and Hou, Qibin},
  booktitle={2021 IEEE Winter Conference on Applications of Computer Vision (WACV)}, 
  title={Rotate to Attend: Convolutional Triplet Attention Module}, 
  year={2021},
  volume={},
  number={},
  pages={3138-3147},
  doi={10.1109/WACV48630.2021.00318}
}

@article{gevorgyan2022scyllaiou,
  title={SIoU loss: More powerful learning for bounding box regression},
  author={Gevorgyan, Zhora},
  journal={arXiv preprint arXiv:2205.12740},
  year={2022}
}

@software{yolo11,
author = {Jocher, Glenn and Qiu, Jing and Chaurasia, Ayush},
license = {AGPL-3.0},
month = jan,
title = {{Ultralytics YOLO}},
url = {https://github.com/ultralytics/ultralytics},
version = {8.0.0},
year = {2023}
}

@article{yolov10,
  title={Yolov10: Real-time end-to-end object detection},
  author={Wang, Ao and Chen, Hui and Liu, Lihao and Chen, Kai and Lin, Zijia and Han, Jungong and others},
  journal={Advances in Neural Information Processing Systems},
  volume={37},
  pages={107984--108011},
  year={2024}
}

@inproceedings{yolov9,
  title={Yolov9: Learning what you want to learn using programmable gradient information},
  author={Wang, Chien-Yao and Yeh, I-Hau and Mark Liao, Hong-Yuan},
  booktitle={European conference on computer vision},
  pages={1--21},
  year={2024},
  organization={Springer}
}

@article{zhang2022perennial,
  title={Perennial cotton ratoon cultivation: A sustainable method for cotton production and breeding},
  author={Zhang, Xin and Yang, Qian and Zhou, Ruiyang and Zheng, Jie and Feng, Yan and Zhang, Baohong and Jia, Yinhua and Du, Xiongming and Khan, Aziz and Zhang, Zhiyong},
  journal={Frontiers in Plant Science},
  volume={13},
  pages={882610},
  year={2022},
  publisher={Frontiers Media SA}
}

@article{wang2020cottonproperties,
  title={Physical structure, properties and quality of cotton},
  author={Wang, Hua and Siddiqui, Muhammad Qasim and Memon, Hafeezullah},
  journal={Cotton science and processing technology: Gene, ginning, garment and green recycling},
  pages={79--97},
  year={2020},
  publisher={Springer}
}

@incollection{negm2020cottonpicking,
title = {1 - Cotton fibres, picking, ginning, spinning and weaving},
editor = {Ryszard M. Kozłowski and Maria Mackiewicz-Talarczyk},
booktitle = {Handbook of Natural Fibres (Second Edition)},
publisher = {Woodhead Publishing},
edition = {Second Edition},
pages = {3-48},
year = {2020},
series = {The Textile Institute Book Series},
isbn = {978-0-12-818782-1},
doi = {https://doi.org/10.1016/B978-0-12-818782-1.00001-8},
url = {https://www.sciencedirect.com/science/article/pii/B9780128187821000018},
author = {Mohamed Negm and Suzan Sanad}
}

@article{cathey1979cottondissecation,
  title={Acceleration of Boll Dehiscence with Desiccant Chemicals 1},
  author={Cathey, George W},
  journal={Agronomy Journal},
  volume={71},
  number={3},
  pages={505--508},
  year={1979},
  publisher={Wiley Online Library}
}

@misc{labelImg,
  author = {Tzutalin},
  title = {LabelImg},
  howpublished = {Git code},
  year = {2015},
  url = {https://github.com/tzutalin/labelImg}
}

@article{albumentations,
  title={Albumentations: fast and flexible image augmentations},
  author={Buslaev, Alexander and Iglovikov, Vladimir I and Khvedchenya, Eugene and Parinov, Alex and Druzhinin, Mikhail and Kalinin, Alexandr A},
  journal={Information},
  volume={11},
  number={2},
  pages={125},
  year={2020},
  publisher={Multidisciplinary Digital Publishing Institute}
}

@misc{roboflowComputerVision,
  author       = {{Roboflow}},
  title        = {Computer Vision Model Leaderboard | Object Detection Benchmarks},
  year         = {2026},
  url          = {https://leaderboard.roboflow.com/},
  note         = {Accessed 13 February 2026}
}

\end{document}